\documentclass[twoside,11pt]{article}

\usepackage[preprint]{jmlr2e}
\usepackage[utf8]{inputenc} 
\usepackage[T1]{fontenc}    
\usepackage{hyperref}       
\usepackage{url}   
\usepackage[ruled,longend,linesnumbered]{algorithm2e}
\usepackage{float}
\usepackage{multirow}
\usepackage{booktabs}      
\usepackage{amsfonts}
\usepackage{rotating}
\usepackage{nicefrac}       
\usepackage{microtype}      
\usepackage{graphicx}
\usepackage{tabularx}
\usepackage[dvipsnames, table]{xcolor}
\usepackage{amsmath}
\usepackage{enumitem}
\usepackage{caption}
\usepackage{listings}
\usepackage[toc,header]{appendix}
\usepackage{minitoc}
\usepackage{subfig}
\usepackage{amssymb}
\usepackage{booktabs}
\usepackage{latexsym}
\usepackage{subcaption}
\usepackage{tikz}
\newcommand{\blue}[1]{\textcolor{blue}{#1}}

\newcolumntype{C}{>{\centering\arraybackslash}X}

\hypersetup{
    colorlinks = true,
    allcolors = {Tan},
    linkbordercolor = {white},
}

\ShortHeadings{Complementary Information Mutual Learning}{Chuyun Shen et al.}
\firstpageno{1}

\begin{document}

\doparttoc 
\faketableofcontents 

\title{Complementary Information Mutual Learning for Multimodality Medical Image Segmentation}

\author{\name Chuyun~Shen \email cyshen@stu.ecnu.edu.cn \\
    \addr School of Computer Science and Technology\\
    East China Normal University\\
    Shanghai 200062, China
    \AND
    \name Wenhao~Li \email liwenhao@cuhk.edu.cn\\
    \addr School of Data Science\\
    The Chinese University of Hong Kong, Shenzhen\\
    Shenzhen Institute of Artificial Intelligence and Robotics for Society\\
    Shenzhen 518172, China
    \AND
    \name Haoqing~Chen \email 51215901005@stu.ecnu.edu.cn \\
    \addr School of Computer Science and Technology\\
    East China Normal University\\
    Shanghai 200062, China
    \AND
    \name Xiaoling~Wang \email xlwang@cs.ecnu.edu.cn \\
    \addr School of Computer Science and Technology\\
    East China Normal University\\
    Shanghai 200062, China
    \AND
    \name Fengping~Zhu \email zhufengping@fudan.edu.cn \\
    \addr Huashan Hospital\\
    Fudan University\\
    Shanghai 200040, China
    \AND
    \name Yuxin~Li \email liyuxin@fudan.edu.cn \\
    \addr Huashan Hospital\\
    Fudan University\\
    Shanghai 200040, China
    \AND
    \name Xiangfeng~Wang \email xfwang@cs.ecnu.edu.cn \\
    \addr School of Computer Science and Technology\\
    East China Normal University\\
    Shanghai AI Laboratory\\
    Shanghai 200062, China
    \AND
    \name Bo~Jin \email bjin@tongji.edu.cn\\
    \addr School of Software Engineering\\
    Shanghai Research Institute for Intelligent Autonomous Systems\\
    Tongji University\\
    Shanghai 200092, China
}

\maketitle

\begin{abstract}
%%%
Radiologists must utilize medical images of multiple modalities for tumor segmentation and diagnosis due to the limitations of medical imaging technology and the diversity of tumor signals. 
This has led to the development of multimodal learning in medical image segmentation. 
However, the redundancy among modalities creates challenges for existing \textit{subtraction}-based joint learning methods, such as misjudging the importance of modalities, ignoring specific modal information, and increasing cognitive load.
These thorny issues ultimately decrease segmentation accuracy and increase the risk of overfitting. 
This paper presents the \textbf{complementary information mutual learning (CIML)} framework, which can mathematically model and address the negative impact of inter-modal redundant information.
CIML adopts the idea of \textit{addition} and removes inter-modal redundant information through inductive bias-driven task decomposition and message passing-based redundancy filtering.
CIML first decomposes the multimodal segmentation task into multiple subtasks based on expert prior knowledge, minimizing the information dependence between modalities.
Furthermore, CIML introduces a scheme in which each modality can extract information from other modalities additively through message passing.
To achieve non-redundancy of extracted information, the redundant filtering is transformed into complementary information learning inspired by the variational information bottleneck.
The complementary information learning procedure can be efficiently solved by variational inference and cross-modal spatial attention. 
Numerical results from the verification task and standard benchmarks indicate that CIML efficiently removes redundant information between modalities, outperforming SOTA methods regarding validation accuracy and segmentation effect. 
To emphasize, message-passing-based redundancy filtering allows neural network visualization techniques to visualize the knowledge relationship among different modalities, which reflects interpretability.
%%%%
\end{abstract}
%% main text

\section{Introduction}

\begin{figure}[!ht]
    \centering
    \subfloat[\textit{Subtract} Operation]{\includegraphics[width=0.95\linewidth]{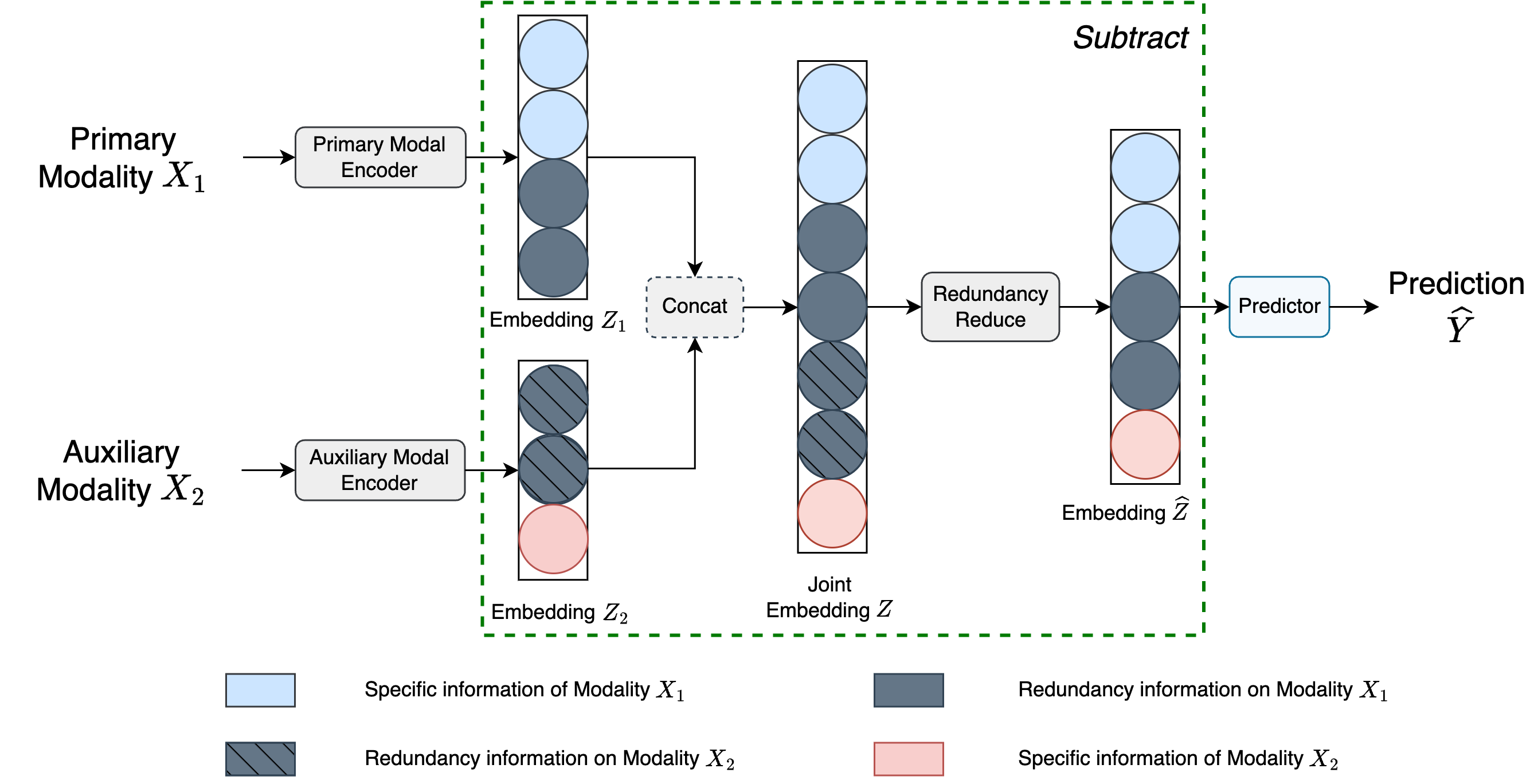}}

    \subfloat[\textit{Addition} Operation]{
        \includegraphics[width=0.95\linewidth]{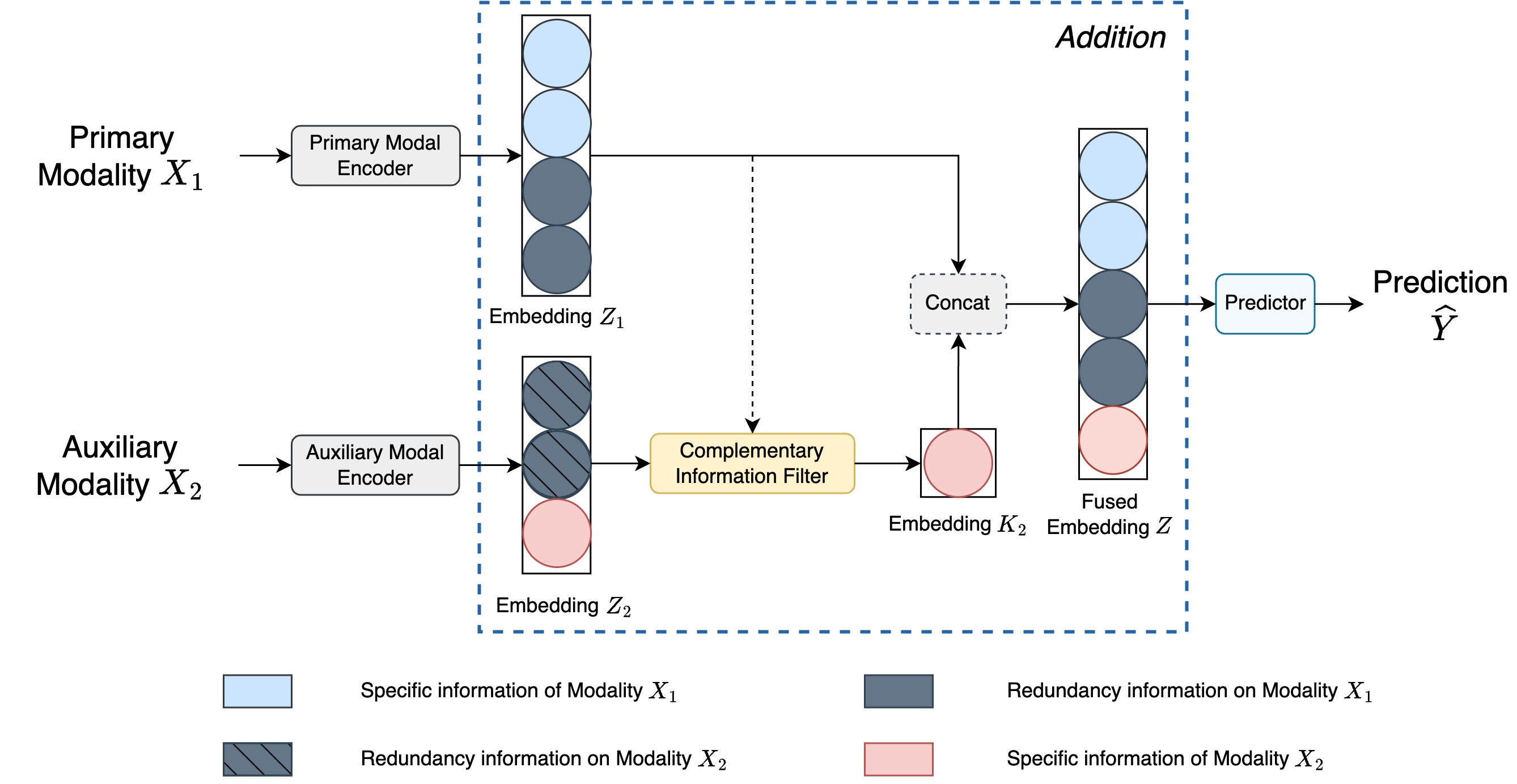}
    }
    
    \caption{\textcolor{blue}{The Diagram of \textit{Addition} and \textit{Subtract} operations. The dark gray circle with black slashes and the dark gray circle without black slashes represent the same information in embeddings. The \textit{Subtract} operation first concatenates the information from different modalities and eliminates redundancy. The \textit{Addition} operation first eliminates cross-modal redundant information and then concatenates the embeddings.}}
    \label{fig:add_sub}
\end{figure}

The capacity of humans to develop a refined comprehension of their external environment can be attributed to the synergistic effects of multiple correlated sensory stimuli. This interaction results in emergent complexity, where the entirety of the system surpasses the simple sum of its individual components~\citep{cohen1997functional,gazzaniga2006cognitive}.
In the field of neuroscience, attention\footnote{Upon encountering external stimuli, humans possess the ability to selectively concentrate on specific elements within these stimuli. This attention mechanism constitutes a fundamental aspect of human cognitive capacity, augmenting the efficacy of information processing~\citep{zhang2012neural}.} focused on modality-rich spatiotemporal events optimizes information acquisition~\citep{li2008visual,li2023multi}. Consequently, this enables individuals to utilize multimodal data for enhanced informational gain.

The significance of multimodal information in influencing human cognition and its implications for the development of artificial intelligence (AI) systems that emulate human intelligence has been widely recognized~\citep{mccarthy2006proposal}. This recognition has led to the advancement of multimodal learning (MML) as a key research area in AI~\citep{baltruvsaitis2018multimodal}.
MML represents a general framework for constructing AI models capable of extracting and relating information from multimodal data~\citep{xu2022multimodal}. 
In recent years, significant advances have been made in MML, particularly in computer vision and natural language processing~\citep{bayoudh2021survey}. 
Large-scale MML models have achieved near-human performance on specific tasks~\citep{ramesh2021zero,reed2022a,rombach2022high}.

This paper focuses on medical image segmentation, an essential task in medical image analysis that involves assigning labels to each pixel or voxel to identify distinct organs, tissues, or lesions.
The intricate pathological or physiological features of human tissues and lesions, combined with the variable sensitivity of imaging technologies across different human body components, necessitate the use of multimodal medical images for patient diagnosis and treatment.
For instance, clinical guidelines for spontaneous intracerebral hemorrhage indicate that determining appropriate strategies relies on the mismatch between two magnetic resonance imaging modalities: perfusion and diffusion imaging~\citep{greenberg20222022}.
Consequently, MML has become increasingly prevalent in medical image segmentation~\citep{zhou2019review}.

Different from the modal-intensive events experienced by humans, machine learning problems with raw multimodal often present unaligned data~\citep{wei2023multi}, highlighting \textit{modality alignment} and \textit{modality synergy} as two thorny issues in MML. 
Fortunately, given the advanced understanding of the human body in medicine, image registration techniques~\citep{hill2001medical} have matured, enabling the alignment of different medical modalities. 
Multimodal medical image segmentation emphasizes the \textit{modality synergy}  problem, namely constructing knowledge relationships among modalities~\citep{han2022radiogenomic} to enable better information complementation and fusion.

\begin{figure}[!ht] 
  \centering
  \begin{minipage}[b]{\linewidth} 
  \captionsetup{justification=centering}
  \subfloat[FLAIR $\rightarrow$ WT]{
    \centering
    \begin{minipage}[b]{0.19\linewidth}
      \includegraphics[width=\linewidth]{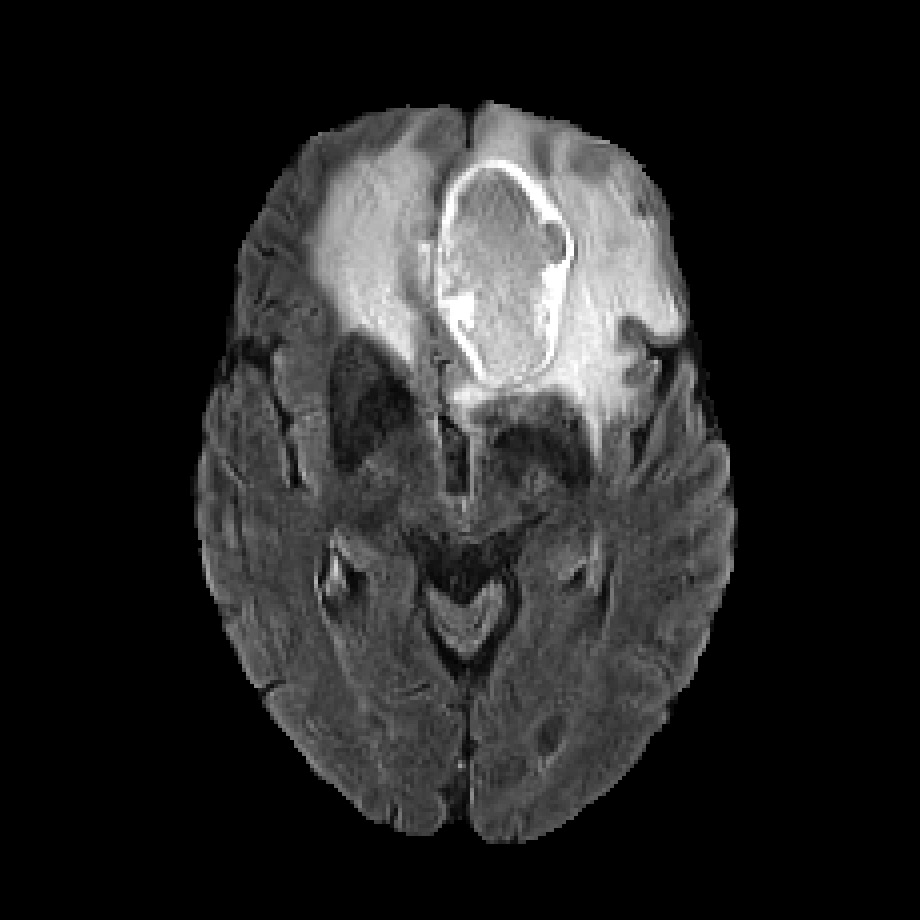}\\
      \vspace{-0.4cm}
      \includegraphics[width=\linewidth]{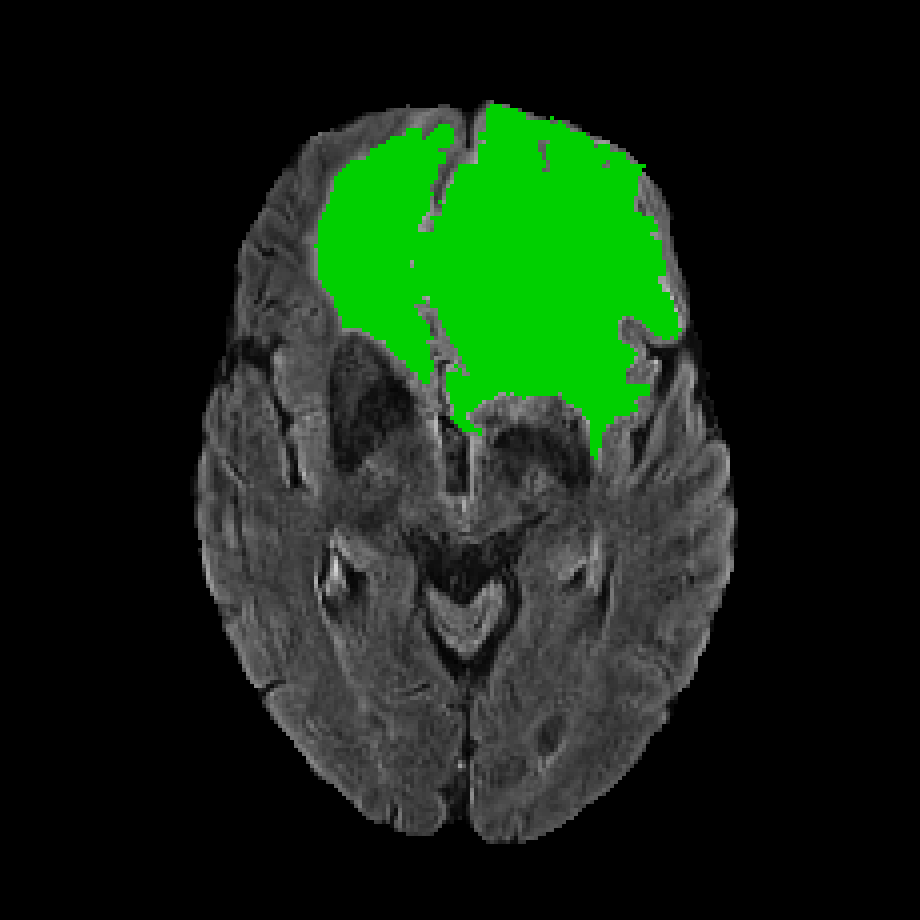}
    \end{minipage}
  } 
  \hspace{-0.2cm}
    \subfloat[\blue{T2} $\rightarrow$ TC]{
    \centering
    \begin{minipage}[b]{0.19\linewidth}
      \includegraphics[width=\linewidth]{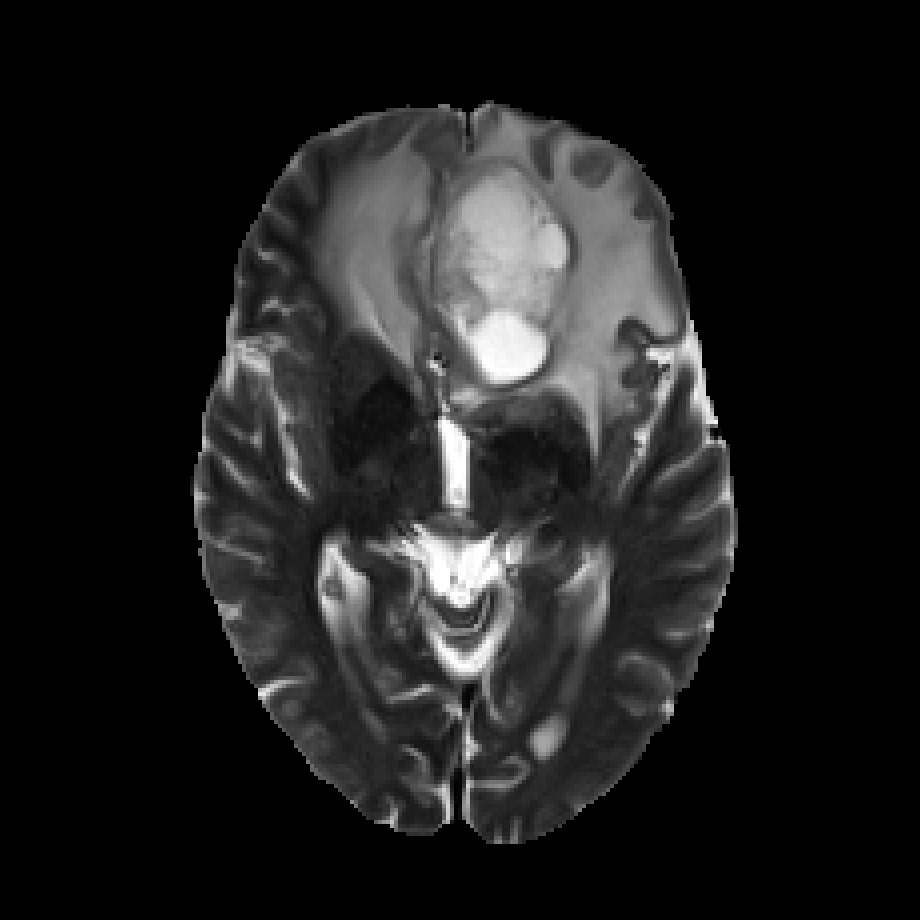}\\
      \vspace{-0.4cm}
      \includegraphics[width=\linewidth]{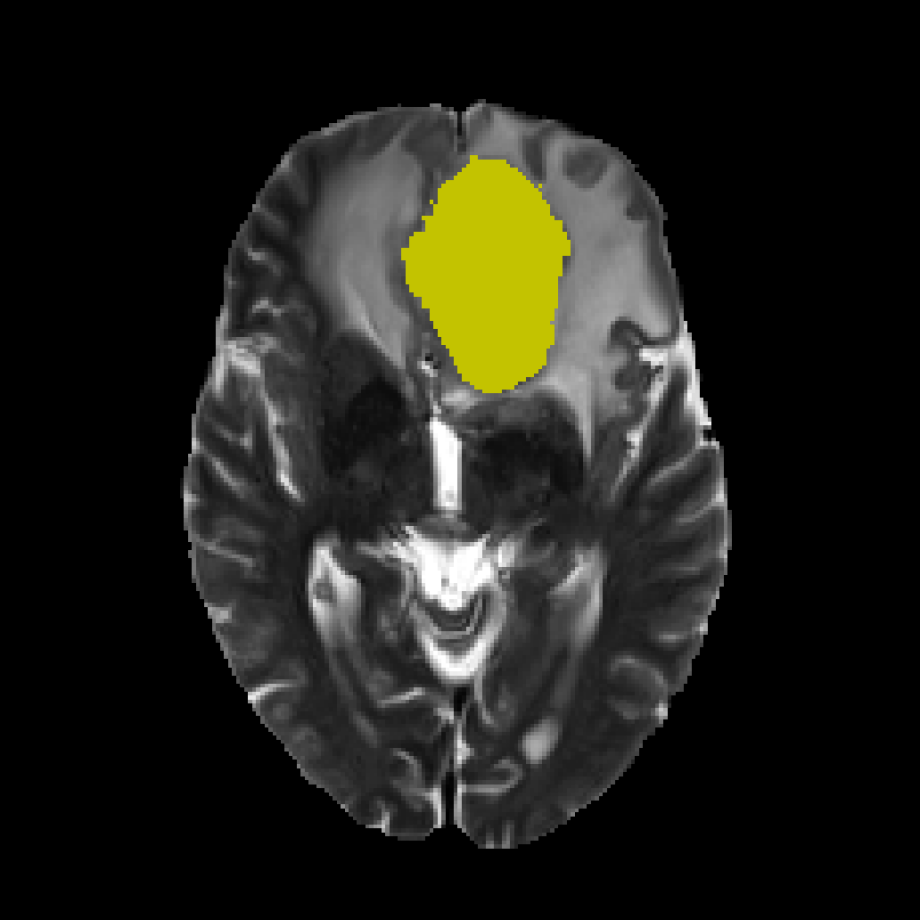}
    \end{minipage}
  }
  \hspace{-0.2cm}
    \subfloat[T1CE $\rightarrow$ TC+ET]{
    \centering
    \begin{minipage}[b]{0.19\linewidth}
      \includegraphics[width=\linewidth]{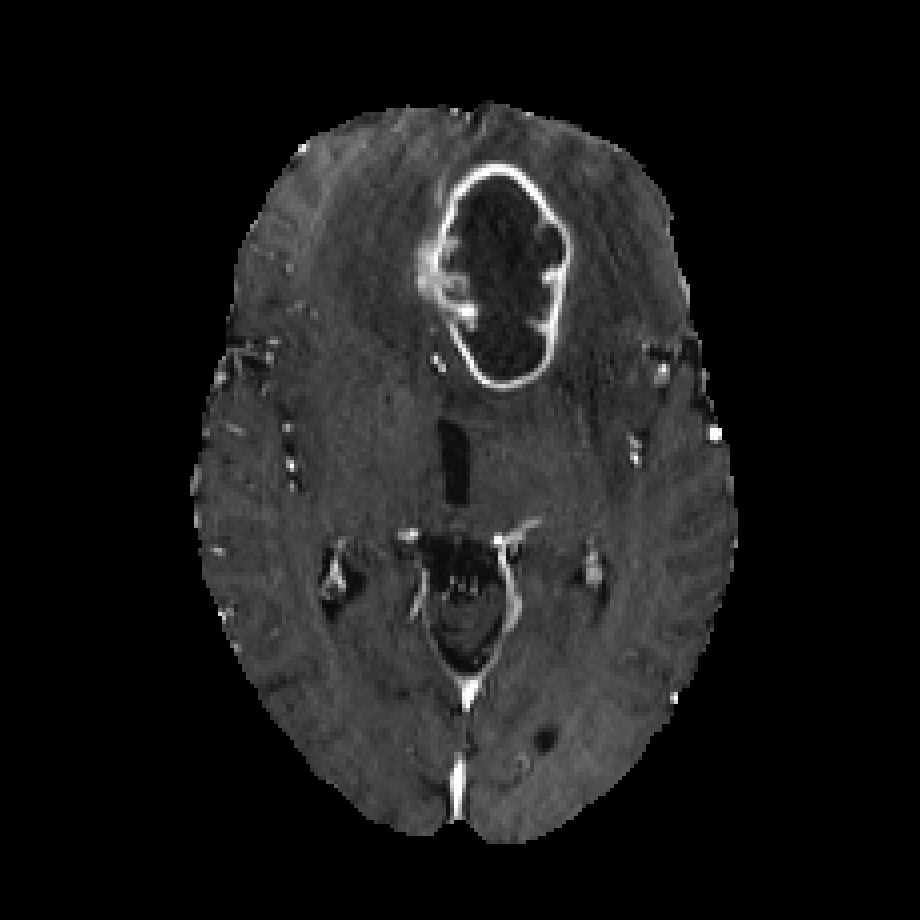}\\
      \vspace{-0.4cm}
      \includegraphics[width=\linewidth]{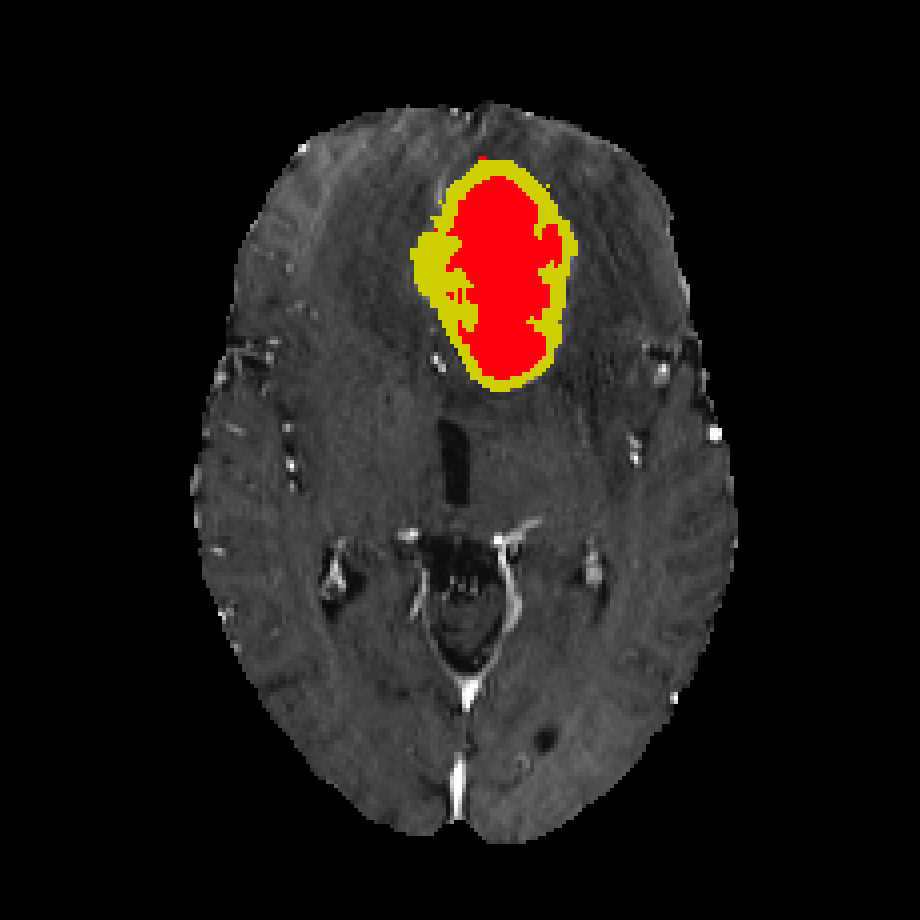}
    \end{minipage}
  }
  \hspace{-0.2cm}
    \subfloat[FLAIR $\rightarrow$ WT+TC+ET]{
    \centering
    \begin{minipage}[b]{0.39\linewidth}
      \includegraphics[width=\linewidth]{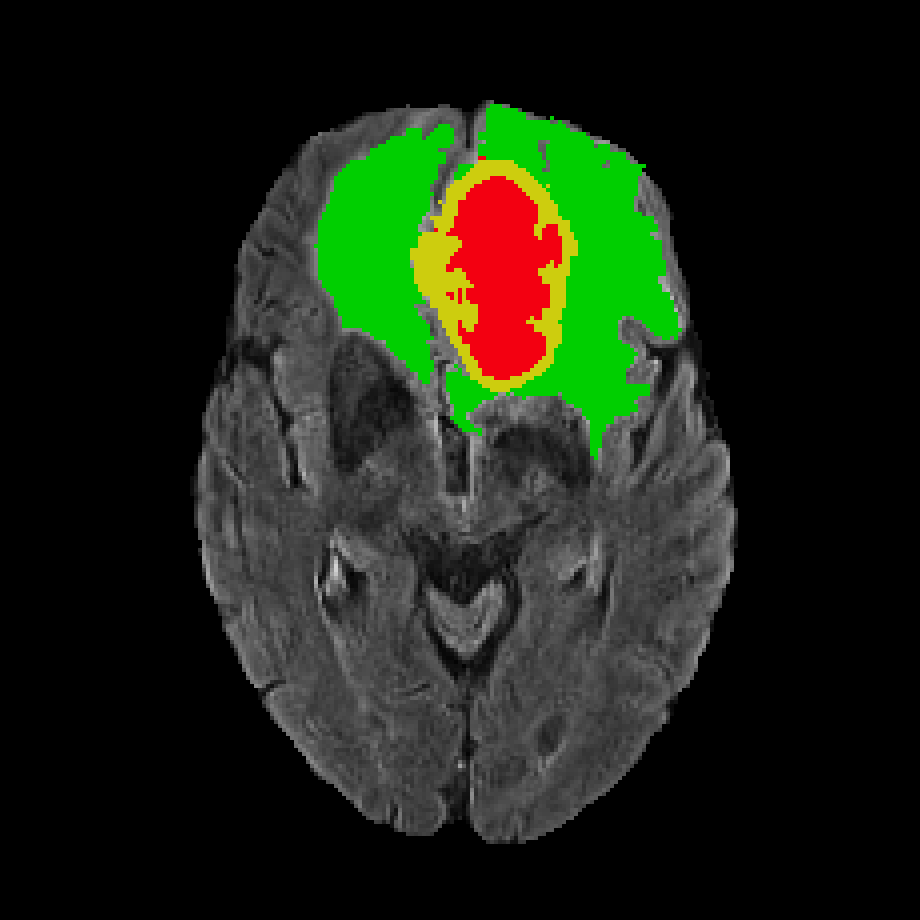}
    \end{minipage}
  }
  \end{minipage}
  \caption{Dataset annotation for \texttt{BraTS2020}. Displayed are image patches with tumor structures annotated in various modalities (bottom left) and the final labels for the entire dataset (right).
  Image patches show from left to right: (a) the FLAIR image and the whole tumor (WT) visible in FLAIR; (b) the T2 image and the tumor core (TC) in T2; (c) the T1CE image, and the enhancing tumor (ET) visible in T1CE (yellow), surrounding the necrotic and non-enhancing tumor core(red); (d) Final labels of the tumor structures: edema (green), ET (yellow), necrotic and non-enhancing tumor core (red).
  In the \texttt{BraTS2020}challenge, images are requested to segment into WT, TC, and ET regions.
  }
  \label{fig:brats2020}
\end{figure}

Multimodal medical image segmentation approaches are commonly designed with an end-to-end scheme to learn intermodal associations~\citep{isensee2018nnu, zhang2021modality, zhang2021brain,zhou2020one, zhou2022tri, ding2021rfnet, dolz2018hyperdense}.
Certain medical conditions~(Figure~\ref{fig:brats2020}) require segmentor to simultaneously identify multiple regions, such as the tumors, edemas, and necrotic tumor cores.
We then refer to the end-to-end learning scheme above as \textit{joint learning}, as it jointly maps multimodal inputs to single or multiple regions.
% \textcolor{green}{yun: as it maps multimodal inputs to a joint representation through multiple encoders or single encoder.}
The joint learning method typically involves fusing multimodal image encoding into a deep encoder-decoder architecture, which outputs the segmentation(s).
These methods can be broadly classified into early fusion and mid-term fusion. 
The former directly concatenates multimodal images as input to the network~\citep{oktay2018attention, isensee2018nnu, zhang2021brain, hatamizadeh2022unetr, mai2022multimodal}. 
While the latter uses modality-specific encoders to extract individual features that are later combined in the middle layers of the network and share the same decoder~\citep{xing2022nestedformer, zhang2021modality, ding2021rfnet, zhou2020one, zhou2022tri, dolz2018hyperdense}.

However, redundant information is present among medical images of different modalities, as evidenced by many works on the cross-modal generation that aim to increase the number of training samples or reduce medical costs by generating high-cost modalities based on low-cost modalities~\citep{van2015does,ben2019cross,zhou2020hi,bouman2023multicenter}. 
In MML, taking into account that fixed representations can only encapsulate a limited amount of information, redundant information can cause joint learning algorithms to misjudge the importance of different modalities~\citep{li2020adversarial}, disregard specific modal information~\citep{wang2022amsa}, generate additional cognitive load~\citep{mayer2003nine,cao2009modality,knoop2019modality}, and ultimately reduce prediction accuracy and result in overfitting~\citep{lin20213d,mai2022multimodal}. 

Redundant information in MML can be categorized into \textit{intra-modality} and \textit{inter-modality} redundancy. 
The former refers to redundancy within a single modality, and the latter refers to consistent information across different modalities.
Existing MML techniques predominantly focus on addressing \textit{intra-modality} redundancy~\citep{lin20213d,wang2022amsa}.
In the case of \textit{inter-modality} redundancy, two separate operations, namely \textit{addition} and \textit{subtraction}, can be utilized to minimize redundant information in joint representations, as depicted in Figure~\ref{fig:add_sub}. 
A considerable number of mid-term fusion strategies~\citep{xing2022nestedformer, zhang2021modality, ding2021rfnet, zhou2020one, zhou2022tri, dolz2018hyperdense} implicitly reduce \textit{inter-modality} redundancy during joint learning by integrating modalities and applying end-to-end learning principles. 
An alternative method~\citep{mai2022multimodal} employs the information bottleneck to decrease redundancy in joint representations associated with the \textit{subtraction} operation. 
Conversely, the \textit{addition} operation, which amalgamates complementary information from multiple modalities, is intuitively superior efficacy in eradicating \textit{inter-modality} redundant information in comparison to the \textit{subtraction} operation.

Furthermore, experienced radiologists often analyze multimodal data in clinical practice by designating a primary modality and several auxiliary modalities for pathological diagnosis. This approach is exemplified in the BraTS challenge~\citep{menze2014multimodal} (Figure~\ref{fig:brats2020}). 
In this challenge, the segmentation target was the different areas of glioblastoma, which is the most common type of brain malignancy. 
Glioblastoma is characterized by its resistance to treatment and poor prognosis, making it a critical focus for advancements in medical imaging and treatment strategies\citep{li2019decoding}.
Human annotators primarily employ the T2 modality\footnote{T1, T1CE, T2, and FLAIR represent four modalities generated by MRI imaging technology.} for segmenting the edema region while using the FLAIR modality to verify the presence of edema and other fluid-filled structures. Subsequently, the tumor core (TC) is identified through the combined use of T1CE and T1 modalities.
The expertise of these radiologists suggests that specific mapping relationships exist between modalities and target areas. Certain modalities facilitate the identification of particular area boundaries, while others serve as supplementary aids. Gleaning insights from this expert knowledge and incorporating it as an inductive bias can potentially reduce the complexity of learning relationships between modalities and corresponding regions. 
A similar example of adding priors to reduce learning difficulty is DetexNet~\citep{liu2020detexnet}, which simplifies low-level representation patterns by embedding expert knowledge.

Inspired by these observations, we propose the \textbf{Complementary Information Mutual Assistance Learning (CIML)} framework. The primary objective of the CIML framework is to effectively eliminate \textit{inter-modal} redundant information in multi-modal segmentation tasks. To leverage doctors' prior knowledge regarding the correspondence between modalities and regions, the learning multimodal segmentation task is decomposed into multiple single-modal segmentation subtasks, as illustrated in Figure~\ref{fig:CIMLoverview}.
Within the CIML framework, each modality serves a dual purpose: as the main modality, the corresponding segmentor encodes its information, integrates messages transmitted by other modalities, and extracts non-redundant information to complete the single-modal subtask; as an auxiliary modality, the corresponding segmentor transmits auxiliary information to other modalities. This approach minimizes the mutual influence of modal information and ensures sufficient feature extraction for the target area in each subtask.
When multiple regions require segmentation, such as in \texttt{BraTS2020}, expert prior knowledge\footnote{To study the robustness of CIML to different task decompositions, we conducted experiments on a medical image segmentation task where expert prior knowledge is not available, using random matching of modalities and regions to be segmented. 
Although this scenario is rare, experimental results show that CIML is moderately sensitive to different task decompositions.} can be employed to match modalities to regions. 
In tasks involving only one region to segment, such as \texttt{autoPET}, each modal segment is matched to the corresponding region individually. 
The primary mode is the matching mode of the target (sub)region, while the remaining modes serve as auxiliary modes. 
The final segmentation is obtained by combining or averaging all unimodal segmentations.

After task decomposition, since the auxiliary mode typically contains less additionally useful information (non-redundant information), we filter redundant information to extract non-redundant information representation from messages transmitted in the auxiliary mode. We first adopt an information theory perspective to transform the problem into an equivalent complementary information learning problem. Inspired by the variational information bottleneck~\citep{alemi2016deep}, we model this problem as a mutual information bi-objective optimization problem. Variational inference is then employed to make optimization problems more tractable, including cross-entropy and Kullback-Leibler (KL) divergence minimization, which can be efficiently solved through automatic differentiation. Finally, we introduce cross-modal spatial attention as a parameterized backbone to achieve practical implementation.

Overall, CIML adopts two mechanisms, namely \textit{task decomposition} and \textit{redundancy filtering}, to minimize the \textit{inter-modal} redundant information that is relied upon by the algorithm in the segmentation process. 
Task decomposition physically minimizes the interdependence of information between modalities (through inductive bias). 
At the same time, redundancy filtering extracts as little information as possible from other modalities using the \textit{addition} operation at the algorithm level.
To validate the effectiveness of CIML, we first perform a visual examination of a human-designed image segmentation task to confirm its redundancy-free nature.
Subsequently, we evaluate our framework on standard multimodal medical image segmentation benchmarks, including \texttt{BarTS2020}, \texttt{autoPET}, and \texttt{MICCAI HECKTOR 2022}. Experimental results demonstrate that CIML significantly outperforms state-of-the-art algorithms in terms of validation accuracy and segmentation quality.

Moreover, the incorporation of task decomposition and redundancy filtering allows us to utilize neural network visualization techniques, such as Grad-CAM~\citep{selvaraju2017grad}, to gain insights into the contribution of each modality to the segmentation of different regions.
By visualizing the relationships and knowledge sharing among modalities, we enhance the credibility and interpretability of multimodal medical image segmentation algorithms, ultimately improving their effectiveness in clinical diagnosis and treatment.
Main contributions can be summarized as follows:
%to \textbf{multimodal medical image segmentation}:

\noindent 1) We introduce the \textbf{Complementary Information Mutual Learning (CIML)} framework, which aims to enhance the information fusion efficiency in multimodal learning. CIML presents a pioneering approach by algorithmic modeling and mitigating the negative impact of \textit{inter-modal} redundant information that arises in the joint learning used by state-of-the-art techniques;

% \smallskip
\noindent 2) CIML adopts a unique perspective of \textit{addition} to eliminate \textit{inter-modal} redundant information through inductive bias-driven \textbf{task decomposition} and message passing-based \textbf{redundancy filtering}, thus effectively decreasing the difficulty of constructing knowledge relationships among modalities in multimodel learning;

% \smallskip
\noindent 3) We establish an equivalent transformation from the redundant filtering problem to the complementary information learning problem based on the variational information bottleneck and solve it efficiently with variational inference and cross-modal spatial attention;

% \smallskip
\noindent 4) Message passing-based redundancy filtering allows for applying neural network visualization techniques, such as Grad-CAM~\citep{selvaraju2017grad}, for visualizing the knowledge relationship among different modalities, which reflects the interpretability.

\smallskip
The following is the roadmap of this paper.
Section \ref{related} provides the related works and preliminaries.
Section \ref{method} describes the proposed framework. 
% xxxxx are presented in Section \ref{xxxx}.
Experiments and numerical results are presented in Section \ref{experiments}, and we conclude this paper in Section \ref{conclusion}.

\section{Related Work and Preliminaries}
\label{related}

In this section, we present a comprehensive literature review on the state-of-the-art multimodal fusion strategies, mutual information, and information bottleneck techniques. In addition, we apply the class activation map methodology to visualize complementary information and thus provide an overview of this technique.

\subsection{Multimodal Fusion and Redundancy Reducing}
Multimodal machine learning has a broad range of applications, including but not limited to audio-visual speech recognition ~\citep{yuhas1989integration}, image captioning ~\citep{xu2015show}, visual question answering ~\citep{wu2017visual}, besides medical image analysis.

Multimodal learning involves the challenge of combining information from two or more modalities to perform accurate predictions ~\citep{baltruvsaitis2018multimodal,10.1145/3649447}. To effectively extract relevant information from multiple sources, various techniques must be employed to capture and integrate an appropriate set of informative features from multiple modalities. Early fusion and intermediate fusion schemes are the most commonly used methods for this purpose.
Early fusion approaches~\citep{oktay2018attention, isensee2018nnu, zhang2021brain, hatamizadeh2022unetr, li2023expectation} adopt a single stream fusion strategy, where multimodal images fuse before input into a neural network. 
However, these methods can hardly explore the inter-modality connections. 
Intermediate fusion approaches\citep{xing2022nestedformer, zhang2021modality, ding2021rfnet, zhou2020one, zhou2022tri, dolz2018hyperdense, yao2024drfuse} follow a multi-stream fusion strategy, where features are fused in the middle layers of the network and share the same decoder.
Among these multi-stream methods, attention mechanisms are often utilized to emphasize contributions from different modalities. 
Methods such as NestedFormer ~\citep{xing2022nestedformer}, ModalityNet ~\citep{zhang2021modality}, Tri-attentionNet ~\citep{zhou2022tri}, and One-shotMIL ~\citep{zhou2020one} leverage attention mechanisms to achieve this. 
The Tri-attentionNet algorithm ~\citep{zhou2022tri} additionally models the relationship between modalities' features, which helps to improve segmentation accuracy. However, they do not fully utilize the relationship between tumor regions and modalities.
To address this limitation, the RFNet framework ~\citep{ding2021rfnet} was proposed, which employs a region-aware fusion scheme. This approach considers the different contributions of various modalities to each region, as different modalities have distinct presentations and sensitivities to different tumor regions.
Besides early fusion and intermediate fusion approaches, the PolicyFuser~\citep{huang2023multi} retains one independent decision for each sensor and fusion decision.
DrFuse~\citep{yao2024drfuse} disentangles shared and unique features, then applies a disease-wise attention layer for each modality to make the final prediction.

\blue{Addressing missing modalities in multimodal medical image segmentation has emerged as a critical research focus. 
A common strategy involves synthesizing the missing modalities using generative models. For example, \citet{orbes2018simultaneous} demonstrated the synthesis of FLAIR images from T1 modality for segmentation tasks. 
Similarly, the heteromodal variational encoder-decoder framework proposed by \citet{dorent2019hetero} performs joint modality completion and segmentation, generating the required modality.
Another approach focuses on learning modality-invariant feature spaces. 
The method by \citet{havaei2016hemis} achieves segmentation by learning features that are consistent across different modalities. ShaSpec \citep{wang2023multi} further enhances this by leveraging all available modalities during training, learning both shared and specific features. Knowledge distillation and feature transfer techniques also address missing modalities effectively. 
The Modality-Aware Mutual Learning framework \citep{zhang2021modality} involves modality-specific models learning collaboratively to distill knowledge. 
ProtoKD \citep{wang2023prototype} transfers pixel-wise knowledge from multi-modality to single-modality data, enabling robust feature representation and effective inference with a single modality.
Latent space information imputation provides an alternative by estimating task-related information of missing modalities. 
M$^3$Care \citep{zhang2022m3care} utilizes auxiliary information from similar patient neighbors, guided by a modality-adaptive similarity metric, to estimate missing data in the latent space, thus facilitating clinical tasks. 
Finally, self-attention-based modality fusion offers another solution. SFusion \citep{liu2023sfusion} introduces a self-attention-based fusion block that automatically learns to fuse available modalities without synthesizing missing ones. 
By projecting features into a self-attention module and generating latent multimodal correlations, SFusion constructs a shared representation for downstream models.}

\blue{These varied approaches underscore the innovative strategies developed to handle missing modalities in multimodal medical image segmentation, enhancing the robustness and accuracy of segmentation models.
In our work, we aim to eliminate redundancy to extract more effective information, which contrasts with most methods addressing missing modalities. These methods typically utilize redundant information to complete other modalities, enabling the use of networks designed for full modalities to accomplish segmentation tasks. 
The main strategy of these works is contrary to ours. 
Moreover, the M$^3$Care \citep{zhang2022m3care} method is worth considering for integration with our approach. 
By leveraging similar patient neighbors, M$^3$Care can extract the necessary complementary information to aid in segmentation. 
In future work, we will try to learn from M$^3$Care to enable our method to cope with the situation of missing modalities.
}

Further, the current methods for fusing multimodal features do not consider \textit{inter-modal} redundancy, which may lead to misjudging the importance of modalities, ignoring specific modal information, and increasing cognitive load. To address this limitation, two methods have been proposed in the context of multi-view, which is similar to multimodal data.
CoUFC ~\citep{zhao2020co} couples the correlated feature matrix and the uncorrelated ones together to reconstruct data matrices. Although CoUFC utilizes an implicit way of eliminating redundancy by focusing on correlated features and uncorrelated features, its solution is not applicable to high-dimensional, high-resolution medical images.
Another work ~\citep{tosh2021contrastive} introduces a contrastive learning method, which learns transformation functions from one view to the other in an unsupervised fashion and then learns a linear predictor for downstream tasks. However, this work focuses on theoretical analysis, lacks validation on complex high-dimensional data, and does not eliminate intra-modal redundancy due to its unsupervised fashion.

There are two main differences between CIML and existing approaches.
Firstly, our CIML algorithm decomposes the original task, thereby facilitating the establishment of an association between modal and target regions. 
Secondly, we employ redundancy filtering to extract complementary information, thereby eliminating redundancy and maximizing information gain from auxiliary modalities.

\subsection{Mutual Information and Information Bottleneck}
The application of information-theoretic objectives to deep neural networks was first introduced in ~\citet{tishby2015deep}, although it was deemed infeasible at the time. 
However, variational inference provides a natural way to approximate the problem. 
To bridge the gap between traditional information-theoretic principles and deep learning, the variational information bottleneck (VIB) framework was proposed in ~\citet{alemi2016deep}. 
This framework approximates the information bottleneck (IB) constraints, enabling the application of information-theoretic objectives to deep neural networks. 

Several works ~\citep{federici2020learning, wu2018multimodal, zhu2020predicting, lee2021variational, mai2022multimodal, wang2019deep} have been proposed to adopt the IB for multi-view or MML, which is the most relevant to our work.
IB variants such as those proposed in ~\citet{federici2020learning, wu2018multimodal, zhu2020predicting, lee2021variational} extend the VIB framework for multi-view learning. 
These methods obtain a joint representation via Product-of-Expert (PoE) ~\citep{hinton2002training}.
Another work ~\citep{wang2019deep} proposes a deep multi-view IB theory, which aims to maximize the mutual information between the labels and the learned joint representation while simultaneously minimizing the mutual information between the learned representation of each view and the original representation.
In addition, a recent study ~\citep{mai2022multimodal} introduced a multimodal IB approach, which aimed to learn a multimodal representation that is devoid of redundancy and can filter out extraneous information in unimodal representations. 
Instead of applying PoE, this work develops three different IB variants to study multimodal representation.
CIML differs from the above multi-view IB methods in the following ways: 1). We take into account the varying importance of different modalities. Drawing on expert knowledge that specific modalities contain a greater amount of relevant information than others, CIML decomposes the task and designates some modalities as primary and others as auxiliary;
2). We assume that the primary modality contains the majority of information about the target region. To maximize information gain and minimize redundancy for segmentation, our method constrains the representation from the auxiliary modalities to contain only complementary information.

\subsection{Class Activation Map}
A widely-used method for determining the most influential pixels or voxels, specifically those with intensity changes that significantly affect the prediction score, involves the generation of a class activation map (CAM) \citep{zhou2016learning, selvaraju2017grad}. 
These maps highlight the regions in the input data that contribute the most to the model's output, thereby providing insights into the decision-making process of the model.
CAM assigns weights to feature maps in a specific convolutional layer and can be easily integrated into a pre-trained deep model without introducing additional parameters. Several variations have been proposed that build upon CAM to more accurately highlight important regions in the image, such as Gradient-weighted Class Activation Mapping (Grad-CAM) ~\citep{selvaraju2017grad}. Grad-CAM uses the gradient signal of the activations in a convolutional layer and has been successfully applied to image classification. An extension of Grad-CAM ~\citep{vinogradova2020towards} produces heatmaps showing the relevance of individual pixels or voxels for semantic segmentation.

In this work, we apply Grad-CAM to visualize voxels that provide complementary information on auxiliary modalities. 
By doing so, we aim to identify and highlight the most informative and relevant regions in the auxiliary modality for accurate target prediction.

\section{Methodology}
\label{method}

In this section, we introduce the Complementary Information Mutual Learning (CIML) framework for medical image segmentation, which aims to efficiently segment through eliminating \textit{inter-modality} redundancy. The framework incorporates two primary mechanisms: \textit{task decomposition} (Section~\ref{ssec:IBTD}) and \textit{redundancy filtering} (Section~\ref{ssec:MPRF}).
The \textit{task decomposition} mechanism seeks to reduce the interdependence of information between modalities by drawing on expert prior knowledge as an inductive bias.
On the other hand, the redundancy filtering mechanism reduces the amount of redundant information extracted from other modalities through the variational information bottleneck and variational inference. 
We also introduce the \textit{cross-modality information gate module} that utilizes cross-modal spatial attention to implement redundancy filtering practically.

We assume that our dataset comprises independent and identically distributed (i.i.d.) samples $ \{ X^i \in \mathbb{R} ^ {T \times W} \}^N_{i=1} $ drawn from a medical image data distribution.
In this context, $i$ represents the index of the image, while $T$ and $W$ represent the number of modalities and the number of voxels, respectively.
To distinguish between different modalities, we employ subscripts, such as $\{X^i_m\}_{m \in \{1,2,\cdots, T\}}$. 
Our objective is to segment the image into $u$ distinct regions by classifying every voxel into one of $u$ classes. 
We define $Y^i \in \{1, 2, \cdots, u\}^W$ as the segmentation mask for the $i$-th sample. Since the multimodal images are spatially aligned, all modalities within a single sample share the same mask, which can be expressed as $Y^i_m = Y^i$ for all $m \in \{1, 2, \cdots, T\}$.

\subsection{Inductive Bias-driven Task Decomposition}
\label{ssec:IBTD}

\begin{figure}[htbp]
    \centering
    \subfloat[CIML for \texttt{BraTS2020} challenge.]{\includegraphics[width=0.75\linewidth]{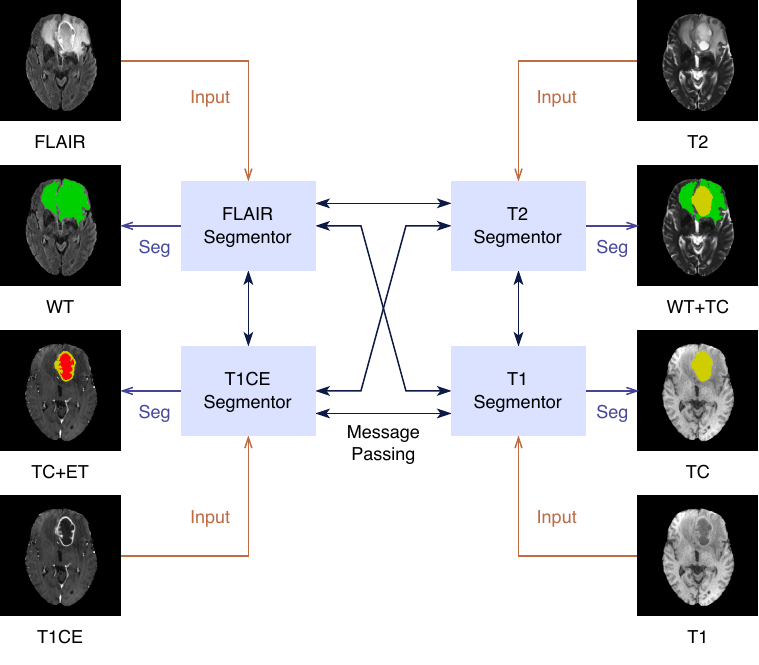}
    }

    \subfloat[CIML for \texttt{autoPET} challenge.]{
        \includegraphics[width=0.75\linewidth]{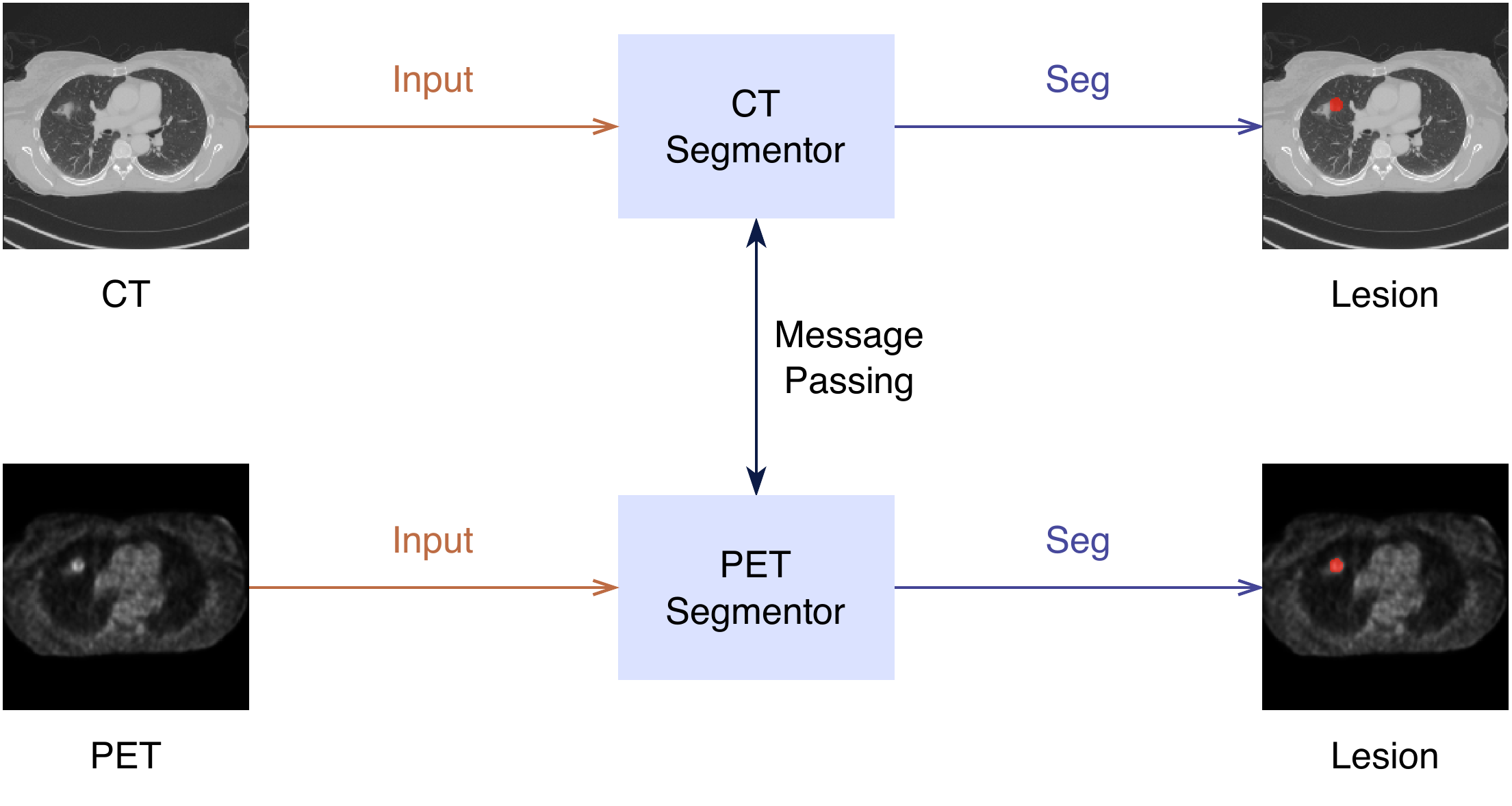}
    }
    
    \caption{\blue{Illustration of complementary information mutual learning (CIML) framework for \texttt{BraTS2020} challenge and \texttt{autoPET} challenge. The input to each segmentor consists of multimodal images that are specific to each modality. After processing, the segmentors send a portion of the embeddings as messages to other segmentors to assist with other sub-tasks and accept messages from other segmentors to extract efficient information. The dark blue lines with bi-directional arrows in the figures represent the message passing. Finally, the segmentors complete their sub-tasks. The \texttt{MICCAI HECKTOR 2022} challenge also applies a similar framework to the \texttt{autoPET} challenge.}}
    \label{fig:CIMLoverview}
\end{figure}

\begin{figure}[ht!]
    \centering
    \includegraphics[width=0.7\textwidth]{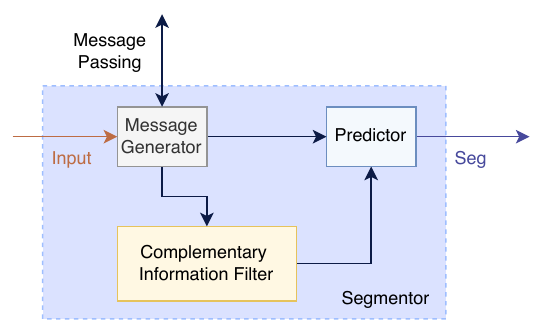}
    \caption{
    Segmentor contains three parts: message generator, complementary information filter, and predictor. 
    }
    \label{fig:segmentoroverview}
\end{figure}

CIML applies a unique perspective of \textit{addition} to eliminate \textit{inter-modal} redundant information.
The first step of \textit{addition} is \textit{task decomposition}, which is driven by the inductive bias extracted from expert prior knowledge. 
As shown in Fig.~\ref{fig:CIMLoverview}, \textit{task decomposition} involves decomposing the task into several subtasks.
For sub-task $\tau_\omega$, the modality $X_{\gamma_\omega}$ is assigned as the primary modality, which contains significant information for the target (sub-)regions segmentation. 
In some cases, multiple modalities are combined as primary modalities for a sub-task, depending on the task.
Furthermore, the remaining modalities, which serve as primary modalities for other sub-tasks, are treated as auxiliary modalities that provide complementary information to assist with the sub-task $\tau_\omega$.
This unique perspective of \textit{addition} allows us to exploit the complementary information from multiple modalities effectively, reducing redundancy and improving the accuracy and efficiency of the segmentation algorithm.

% This task decomposition is so flexible that it allows algorithm designers to develop different segmentation models for each sub-task and to use different approaches to generate messages for mutual learning. 

% Furthermore, in subsection \ref{ssec:Segmentor}, we introduce the module in segmentors in detail.  

\begin{figure}[h]
    \centering
    \includegraphics[width=0.9\textwidth]{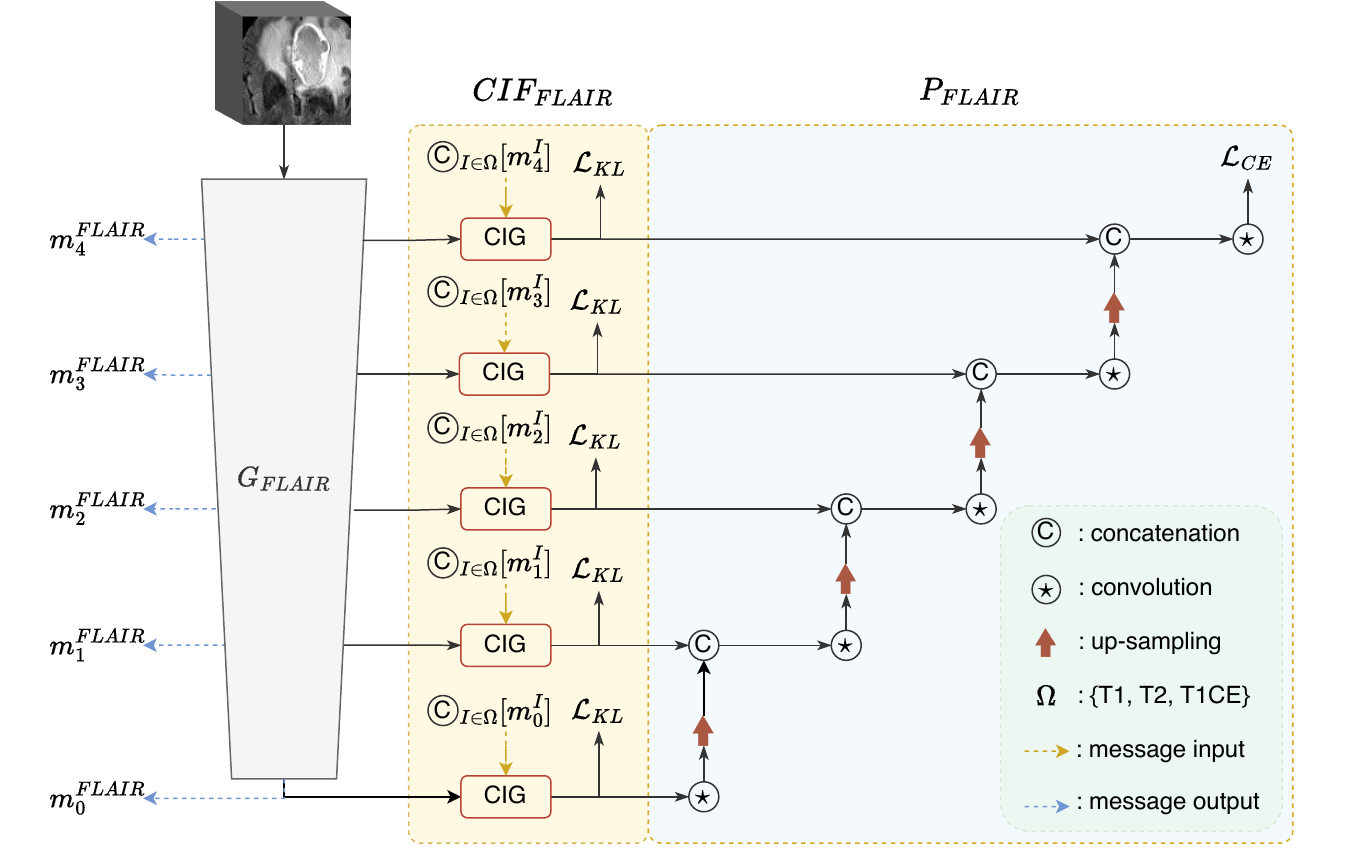}
    \caption{
    \blue{Schematic of the network architecture of the segmentor. 
    Generator $G_{FLAIR}$ is employed to extract features from FLAIR images individually. 
    Complementary Information Filter (CIF) Module is used to extract complementary information from messages, and predictor $P_{FLAIR}$ is utilized to generate the final segmentation.
    }}
    \label{fig:Complementary Information Mutual Learning}
\end{figure}

Additionally, we utilize a message-passing mechanism between sub-tasks to transport efficient information from auxiliary modalities.
CIML utilizes a distinct sub-model for each sub-task, which is responsible for extracting uni-modal features, performing message passing, obtaining complementary information, and predicting the target (sub-)regions. 
These sub-models are referred to as \textit{segmentors} (Figure~\ref{fig:segmentoroverview}), with the segmentor for sub-task $\tau_\omega$ denoted as
$$f_\omega \left( X_{\gamma_\omega}, M_{1 \sim T /\ \gamma_\omega} \mid \theta_\omega \right)$$and parameterized by $\theta_\omega$.
In this notation, $M_{1 \sim T /\ \gamma_\omega}$ denotes messages from other sub-tasks. 

A segmentor comprises three fundamental components: a message generator, a complementary information filter, and a predictor.
The message generator encodes input images to produce embeddings and messages. Subsequently, the complementary information filter utilizes the embeddings from the message generator and messages from other modalities to extract complementary information, thereby enhancing the segmentor's performance. Lastly, the predictor leverages the embeddings generated by the message generator and the complementary information to generate the final predictions.

To describe \textit{task decomposition} more clearly, we take \texttt{BraTS2020} as an example.
As illustrated in Fig.\ref{fig:CIMLoverview}, the original multi-target task is divided into four distinct sub-tasks.
The sub-tasks involve utilizing FLAIR as the primary modality for segmenting the whole tumor (WT) region, employing T1 as the primary modality for segmenting both the tumor core (TC) and the enhanced tumor (ET), leveraging T2 as the primary modality for segmenting the WT and TC regions and adopting T1CE as the primary modality for segmenting the TC and ET regions.
This task decomposition is based on expert prior knowledge, which suggests that the selected primary modalities contain the most informative features for accurately segmenting their respective target regions. 
In addition, to test the performance of different decompositions, we designed ablation experiments to compare several different decomposition methods, as detailed in Section \ref{experiments}.

The segmentor architecture is based on the nnUNet~\citep{isensee2018nnu} and utilizes an intermediate fusion scheme. 
It consists of an encoder and a decoder, which are connected through skip connections. 
A schematic representation of the segmentor's overall structure is provided in Fig.\ref{fig:segmentoroverview}, while the structure corresponding to sub-task $\tau_{FLAIR}$ is depicted in Fig.\ref{fig:Complementary Information Mutual Learning}.
For the sub-task $\tau_{FLAIR}$,
the message generator, $G_{FLAIR}$, acts as the encoder and is composed of four stages. It takes the FLAIR images as input, which are first cropped into 3D patches and generate embedded images. 
At each stage, $G_{FLAIR}$ produces embeddings that are skip-connected to the decoder in the original U-Net architecture. 
These embeddings serve as messages denoted by $ \{M_{FLAIR}^{\omega}\}_{ \omega \in \{1, 2, 3, 4 \}}$ for assisting other sub-tasks.
The decoder of the segmentor for sub-task $\tau_{FLAIR}$ is composed of the complementary information filter ($CIF_{FLAIR}$) and predictor ($P_{FLAIR}$). $CIF_{FLAIR}$ is designed to utilize the embeddings from the message generator ($G_{FLAIR}$) and messages from other segmentors to filter out complementary information. 
It also contains four stages, and in each stage, $CIF_{FLAIR}$ incorporates a Cross-modality Information Gated (CIG) module based on cross-modal spatial attention, which will be explained in detail in the following subsection. 
The output of $CIF_{FLAIR}$ combines the embeddings from the encoder and the complementary representation. Finally, $P_{FLAIR}$ predicts the results of the segmentor based on the output of the complementary information filter.

\subsection{Message Passing-based Redundancy Filtering}
\label{ssec:MPRF}
In the second phase of the \textit{addition} process, message passing-based redundancy filtering is employed to eradicate \textit{inter-modality} redundancy, thereby extracting supplementary information. 
Drawing inspiration from the variational information bottleneck, we reformulate the problem as a bi-objective mutual information optimization problem. 
Subsequently, we leverage variational inference to make the optimization problem more tractable by minimizing cross-entropy and KL divergence, and we efficiently solve it using automatic differentiation. Finally, we employ cross-modal spatial attention as a parameterization backbone to obtain a practical implementation.

To facilitate discussion, we concern a scenario where two sub-tasks, specifically $\tau_1: X_1 \rightarrow Y_1$ and $\tau_2: X_2 \rightarrow Y_2$, are present. 
Our primary focus is on sub-task $\tau_1$, and the same principles can be extended to more than two sub-tasks. 
In this context, $X_1$ and $X_2$ signify the first and second modalities, while $Y_1$ and $Y_2$ denote the corresponding ground truth for each sub-task.

$X_1$ represents the primary modality encompassing the majority of information pertaining to the target region $Y_1$.
In accordance with standard supervised learning literature, we predict $Y_1$ directly by minimizing the supervised learning loss:

\begin{equation}
\begin{split}
\mathcal{L}_{SL} = -\mathbb{E}_{x_1^i, y_1^i \sim \{ X_1, Y_1 \}} \log f_{X_1}(y_1^i \mid x_1^i),
\end{split}
\end{equation}
where $f_{X_1}$ denotes the parameterized segmentation function. 
Besides, we aim to generate a representation $K_2$ derived from primary and auxiliary modalities that encapsulate complementary information to aid in predicting the target $Y_1$.

\begin{figure}[ht!]
    \centering
    \includegraphics[width=0.4\textwidth]{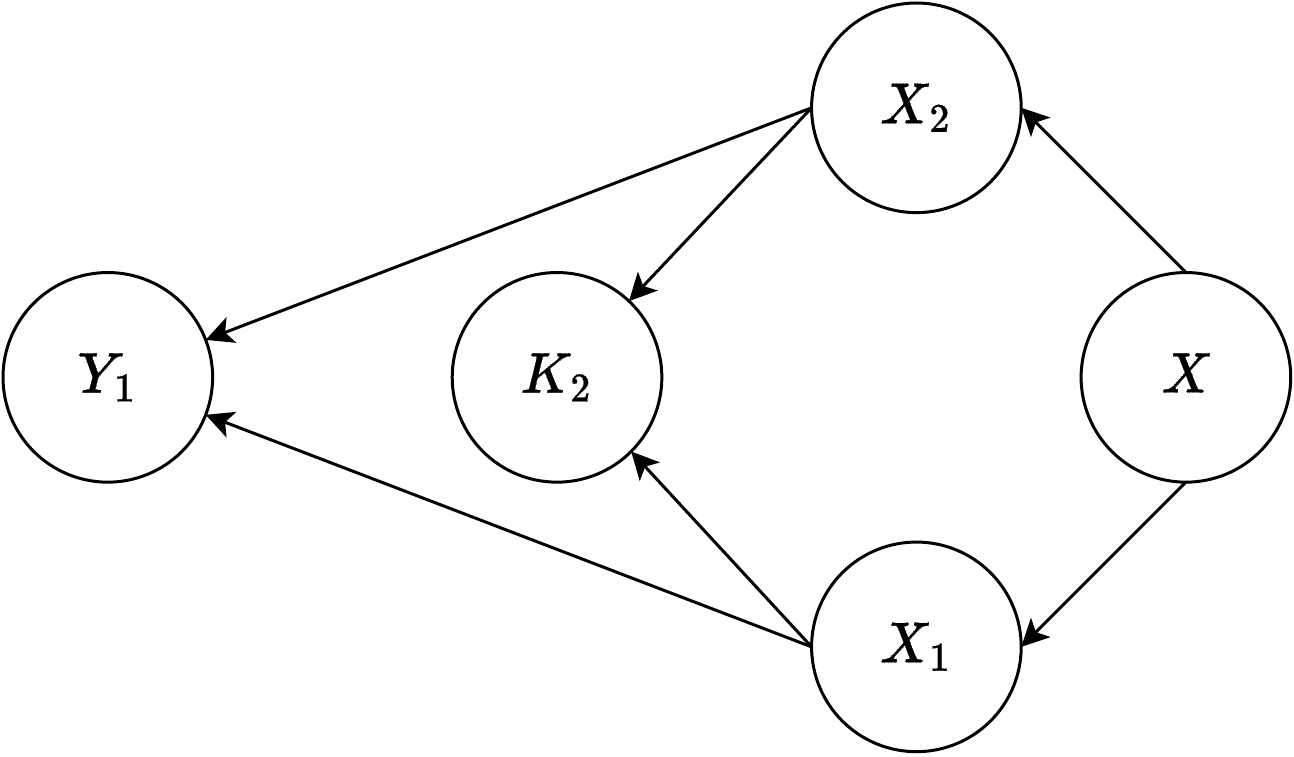}
    \caption{
    A Bayes graph representing the relationship of modalities $X_1$, $X_2$, target $Y_1$, and representation $K_2$. 
    }
    \label{fig:bayes}
\end{figure}

This problem can be modeled within a Bayesian graph (refer to Figure~\ref{fig:bayes}) following three Markov chains: $X_1 \leftarrow X \rightarrow X_2$, $ X_1 \rightarrow Y_1 \leftarrow X_2$, and $ X_1 \rightarrow K_2 \leftarrow X_2 $.
Here, $X$ represents the patient as a hidden variable, and $K_2$ is the representation derived from $X_1$ and $X_2$.

\begin{figure}[ht!]
    \centering
    \includegraphics[width=0.5\textwidth]{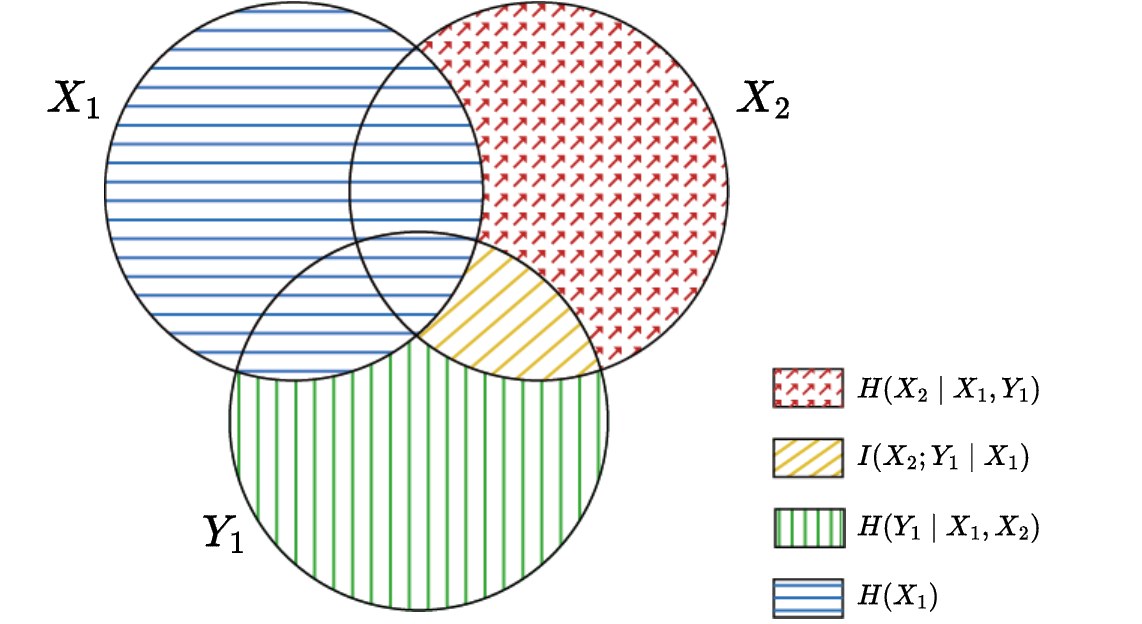}
    \caption{
    Venn diagram of information-theoretic measures for three variables $X_1, X_2$, and $Y_1$, represented by the upper left, upper right, and lower circles, respectively.}
    \label{fig:entropy}
\end{figure}

Figure~\ref{fig:entropy} illustrates the interdependence among variables $X_1$, $X_2$, and the target variable $Y_1$ using a Venn diagram to demonstrate their overlapping relationships \blue{(see \ref{sec:bayesian_graph_rep} for some basic knowledge about Mutual Information and Venn Diagrams)}. 
Entropy ($H$) and mutual information ($I$) are essential concepts in information theory, as they measure the uncertainty in a set of outcomes and the reduction in uncertainty about one variable due to the knowledge of another variable, respectively~\citep{shannon2001mathematical}.
Further, representation $K_2$ is generated following the Markov chain $ X_1 \rightarrow K_2 \leftarrow X_2 $, so the Venn diagram of $K_2$ is within the union of Venn diagram of $X_1$ and $X_2$.
The area shown in Figure~\ref{fig:entropy} with yellow stripes represents the complementary information not contained in $X_1$ and contributes to the identification of $Y_1$.

Employing Markov chains as a foundation, the mutual information ${\cal{I}}_1(X_1, X_2; K_2)$ can be partitioned into three distinct components through the application of the chain rule of mutual information \blue{(more details see Appendix \ref{sec:der_eq2})}:

\begin{equation}
\begin{split}
&{\cal{I}}_1(X_1,X_2;K_2) \\ 
&=\underbrace{{\cal{I}}_2(K_2;Y_1 \mid X_1)}_{\text {complementary predictive information}}
+
\underbrace{{\cal{I}}_3(K_2; X_1)}_{\text {duplicated information}} \\
&+
\underbrace{{\cal{I}}_4(K_2; X_2 \mid X_1, Y_1)}_{\text {unique but irrelevant information}},
\end{split}
\end{equation}
where ${\cal{I}}_2(K_2;Y_1 \mid X_1)$ represents the information in $K_2$ that is not involved by modality $X_1$ and is predictive of $Y_1$. 
While ${\cal{I}}_3(K_2; X_1)$ indicates the duplicated information already involved in modality $X_1$, and ${\cal{I}}_4(K_2; X_2 \mid X_1, Y_1)$ indicates unique information but irrelevant information. 
We regard ${\cal{I}}_3$ and ${\cal{I}}_4$ as \textit{inter-modality} redundancy.

Based on these observations, we formulate two objectives to generate $K_2$, i.e.,
\begin{equation}
    \left\{ \begin{array}{l}
        \min_{K_2} {\cal{I}}_1 (X_1,X_2; K_2),\\  
        \max_{K_2} {\cal{I}}_2 (K_2;Y_1\mid X_1).\\
    \end{array} \right.
\end{equation}
The first objective seeks to maximize the mutual information between $K_2$ and the target $Y_1$, given the modality information of modality $X_1$. This constraint ensures $K_2$ contains the information depicted in the Venn diagram with yellow stripes. 
To guarantee that $K_2$ encompasses solely essential information and minimizes redundancy, the mutual information between $K_2$ and modalities $X_1$, $X_2$ is minimized. 
This dual objective optimization constrains $K_2$ to be a representation of complementary information containing only indispensable information.
Considering the above Bayesian network, the joint probability can be expressed as \blue{(see detailed derivation in Appendix \ref{sec:der_eq4})}:
\begin{equation}
    \begin{aligned}
        p( X_1,X_2, Y_1,K_2 )=p( K_2\mid X_1,X_2 )\cdot p(X_1,X_2, Y_1).
    \end{aligned}
\end{equation}

Furthermore, variational inference can be employed to render the optimization problem unconstrained.
{\color{blue}{The first objective has an upper bound (see detailed derivation in Appendix \ref{sec:der_eq5}):
\begin{equation}
    \begin{split}
        & {\cal{I}}_1(X_1, X_2; K_2) \\
        &= \int p \left( x_1,x_2,\kappa_2 \right) \cdot \log \left( p \left( \kappa_2 \mid x_1,x_2 \right) \big/ p(\kappa_2)\right)  dx_1d\kappa_2dx_2 \\
        &\le \int p(x_1,x_2,\kappa_2)\cdot \log \left( p \left( \kappa_2\mid x_1,x_2\right) \big/ r(\kappa_2)\right) dx_1d\kappa_2dx_2\\
        &\approx \frac{1}{N}\sum_i^N \int 
        p \left( \kappa_2\mid {x_1^i},{x_2^i} \right) \cdot 
        \log \left( p \left( \kappa_2\mid {x_1^i},{x_2^i} \right) \big/ r(\kappa_2) \right) d\kappa_2,
    \end{split}
\end{equation} 
where $r(\kappa_2)$ is a standard normalization distribution.
In practice, we can use neural  networks $\mu_{\theta}(x_1, x_2), \sigma_{\theta}(x_1,x_2)$ to approximate$p \left( \kappa_2 \mid x_1, x_2 \right)$ by 
 $\mathcal{N} \left( \mu_{\theta}(x_1, x_2), \sigma_{\theta}(x_1,x_2) \right)$.
 The second mutual information maximization objective possesses a lower bound (see detailed derivation in Appendix \ref{sec:der_eq6}):
\begin{equation}
    \begin{split}
       & {\cal{I}}_2(K_2;Y_1\mid X_1)\\ &= \int dx_1d\kappa_2dy_1dx_2
       {p(x_1,x_2,\kappa_2,y_1)\log{\frac{p(y_1\mid \kappa_2,x_1)}{p(y_1\mid x_1)}}}\\
        &\ge \int dx_1d\kappa_2dy_1dx_2 
        {p(x_1,x_2,\kappa_2,y_1)\log{\frac{q(y_1\mid \kappa_2,x_1)}{p(y_1\mid x_1)}}}\\
        &\approx \frac{1}{N}  \sum_i^N \int d\kappa_2 p(\kappa_2\mid {x_1^i},{x_2^i})\log{q({y_1^i}\mid {\kappa_2},{x_1^i})}+H,
    \end{split}
\end{equation} 
where $q(y_1\mid \kappa_2,x_1)$ serves as a variational approximation to $p(y_1\mid \kappa_2,x_1)$.
Notice that $H$ is independent of the optimization procedure and can be ignored.}}
Combining both of these bounds, we have that,
\begin{equation}
    \begin{aligned}
    \mathcal{L}_{com} &\approx \frac{1}{N}\sum_{i=1}^{N} \int \left[- p \left( \kappa_2 \mid {x_1^i},{x_2^i} \right) \cdot \log{ q \left( {y_1^i} \mid {\kappa_2},{x_1^i} \right)}\right.
    \\ &
   \left.\, + \beta\, p \left( \kappa_2 \mid {x_1^i},{x_2^i} \right) \cdot \log \left( p(\kappa_2\mid {x_1^i},{x_2^i}) \big/ r(\kappa_2)\right)
     \right] d\kappa_2.
    \end{aligned}
\end{equation}
Furthermore, we can formulate the loss in this way,
\begin{equation}
    \begin{aligned}
    \mathcal{L}_{com} = 
    \mathcal{L}_{CE} + \beta \mathcal{L}_{KL}, \label{equ:loss}
    \end{aligned}
\end{equation}
where $\beta \geq 0$ controls the tradeoff between two objectives.
The total loss contains cross entropy and Kullback–Leibler divergence, and the former is
\begin{equation}
    \begin{aligned}
    \mathcal{L}_{CE}&= 
    \frac{1}{N}\sum_{i=1}^{N} \int \left[ - p(\kappa_2\mid {x_1^i},{x_2^i})\log{q({y_1^i} \mid {\kappa_2},{x_1^i})} \right] d\kappa_2
    \\
    &=\mathbb{E}_{ 
    \begin{subarray}{l}
    \epsilon \sim\mathcal{N}(0,1) \\
    x_1^i, x_2^i, y_1^i \sim \{X_1, X_2, Y_1\}
    \end{subarray}
    }   -\log{q \left( {y_1^i} \mid{x_1^i}, f_{\theta}({x_1^i} ,{x_2^i},\epsilon) \right)},
    \end{aligned}
\end{equation} where $\kappa_2$ is sampled with $f_{\theta}({x_1^i},{x_2^i},\epsilon)$ which is a deterministic function of $x_1^i, x_2^i$ and the gaussian random variable $\epsilon$ which is sampled from a normal gaussian distribution.
On the other hand, $\mathcal{L}_{KL}$ is shown as follow
\begin{equation}
\begin{aligned}
\mathcal{L}_{KL} 
&=\frac{1}{N}\sum_{i=1}^{N}\int \left[ p \left( \kappa_2\mid {x_1^i},{x_2^i} \right) \cdot \log \left( p\left( \kappa_2\mid {x_1^i},{x_2^i}\right) \big/ r(\kappa_2)\right)\right] d\kappa_2
\\ &= \mathrm{KL} \left[ p \left( K_{2} \mid {X_1},{X_2} \right) \mid \mid r(K_{2} ) \right].
\end{aligned}
\end{equation} % where $\mathrm{KL}$ is Kullback–Leibler divergence.

Since both $q({y_1^i} \mid {\kappa_2},{x_1^i})$ and $f_{X_1}(x_1^i)$ involve $X_1$ as input and share the same prediction target $Y_1$, the cross-entropy loss $\mathcal{L}_{CE}$ and the supervised loss $\mathcal{L}_{SL}$ can be simplified by combining them and retaining only the cross-entropy loss $\mathcal{L}_{CE}$.

The resulting loss function consists of two components: the cross-entropy term measures the discrepancy between our predictions and the targets, while the $\mathrm{KL}$ term constrains the representations from CIG modules to represent complementary information. This approach enables the efficient extraction of complementary information representations from messages through automatic differentiation. 
The function
$q \left( {y_1^i} \mid {x_1^i}, f({x_1^i} ,{x_2^i},\epsilon) \right)$ can be utilized to predict the target $Y_1$.

\begin{figure}[htb]
    \centering
    \includegraphics[width=0.7\textwidth]{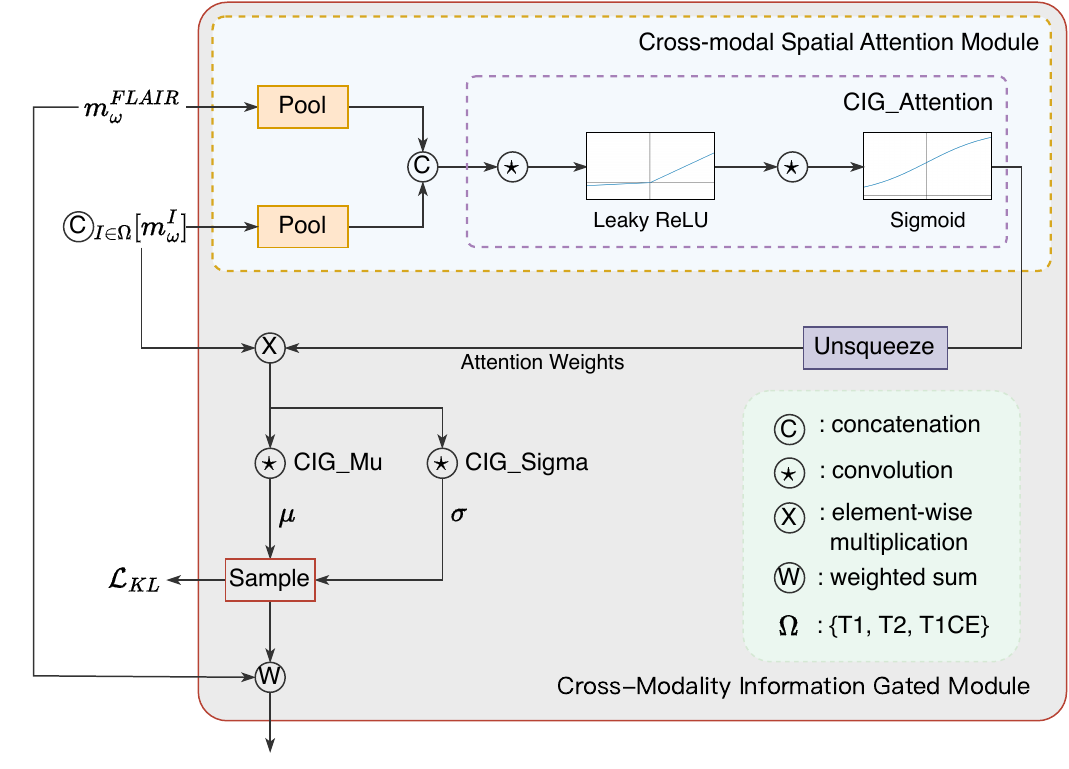}
    \caption{\blue{Visualization of the Cross-Modality Information Gated Module: utilizing cross-modal spatial attention to identify key voxels, filtering complementary information via attention weights, and incorporating residual mechanism for combining local information.}}
    \label{fig:CIGMpic}
\end{figure}

Moreover, in order to improve the extraction of complementary information in auxiliary modalities, we propose a \textit{cross-modality information gate} (CIG) that merges our formulated loss function. 
The CIG module utilizes cross-modal spatial attention as a parameterization backbone to achieve a practical implementation.

Figure~\ref{fig:Complementary Information Mutual Learning} shows that each stage of the complementary information filter in the segmentor contains a CIG module. 
This CIG module is based on a cross-modal spatial attention mechanism and our proposed loss in Equation~\ref{equ:loss} to extract complementary information.
Figure~\ref{fig:CIGMpic} shows the architecture of the CIG module. 
It inputs the output features of encoders and the messages from other segmentors. 
The spatial attention mechanism has proven effective in achieving high performance with limited parameters and has been utilized in various computer vision applications~\citep{woo2018cbam, zhou2022tri, hinton2015distilling}. 
We leverage a cross-modal spatial attention module to extract complementary information that is only contained in messages from other segmentors.

The CIG module utilizes average pooling to reduce the number of network parameters while preserving location information. 
The pooled features are then concatenated and squeezed before passing through \blue{CIG\_Attention module, which contains} two convolutional layers with Leaky ReLU and sigmoid activation functions to obtain the cross-modality attention weights. 
These weights highlight the critical voxels and are unsqueezed and multiplied element-wise with the messages from other sub-tasks.

To generate the complementary information features, two convolutional layers, \blue{CIG\_Mu and CIG\_sigma}, are utilized to obtain the mean $\mu$ and standard deviation $\sigma$. 
These parameters are then used in the reparameterization trick, which is constrained by $\mathrm{KL}$ terms. 
\blue{We employ a weighted sum operation to integrate dominant and complementary information features.
Specifically, we first align the channel dimensions of the auxiliary modality features with those of the primary modality by applying convolutional (weighted) adjustments. 
Subsequently, we combine these features through summation to complete the information.}

With this module, each segmentor in the proposed CIML framework can effectively extract complementary information from auxiliary modalities, allowing for the improvement of multimodal medical image segmentation.

\section{Experiments and Results}
\label{experiments}
% We first verify that our proposed variational inference for complementary information learning
% is efficient through a demonstration experiment.
% Further, to verify that eliminating redundancy helps improve segmentation performance, we compare our proposed CIML with previous convnetand and transformer-based architecture on three public datasets and visualize our results.

In this section, we investigate two questions to determine the feasibility and efficiency of our approach:

\noindent Q1): Can the message passing-based redundancy filtering effectively extract non-redundant information from messages transmitted by auxiliary modalities?

\noindent Q2): Whether removing \textit{inter-modality} redundancy can improve the quality of medical image segmentation?

To solve these issues, we evaluate the proposed approach on four tasks, namely the \texttt{ShapeComposition}, \texttt{BraTS2020}, \texttt{autoPET} and the \texttt{MICCAI HECKTOR 2022}.
We address Q1) using the hand-crafted demonstrated task \texttt{ShapeComposition} and address Q2) using three standardized benchmarks \texttt{BraTS2020}, \texttt{autoPET} and \texttt{MICCAI HECKTOR 2022}.
To further evaluate the effectiveness of our proposed CIML, we conduct ablation experiments by removing components from the proposed framework.

Unlike existing SOTA methods, task decomposition and redundant filtering enable us to use neural network visualizers, such as Grad-CAM~\citep{selvaraju2017grad}, to provide insight into the contribution of each modality to the segmentation of different regions.
By visualizing the relationship of knowledge among modalities, the credibility of multimodal medical image segmentation algorithms is improved, enhancing their effectiveness in clinical diagnosis and treatment.
% Ultimately, we visualize the segmentation results of experiments and the complementary information provided by the auxiliary modalities to demonstrate the accuracy of the segmentation and how the auxiliary modality assists in segmentation.

\begin{table}[!htbp]
\caption{\blue{Network configurations of our CIML. 
We report the Operators, Input Size, Output Size and Kernel Size. Stride can be easily inferred according to Input and Output size as we apply padding equals 1. 
Additionally, we use Batch Normalization in the BraTS2020 dataset and Instance Normalization in other experiments. 
In detail, we use $\textrm{P}$ to denote patch size, $\textrm{C}$ to denote the base number of filters, $\textrm{O}$ to denote the number of channels of output, and $\textrm{K}$ to denote the number of messages.}}
\label{tab: architecture}
\centering
\resizebox{\textwidth}{!}{%
\begin{tabular}{l|l|l|l|l|l}
\toprule
\textbf{Architecture}              & \textbf{Modules}                         & \textbf{Operators}                          & \textbf{Input Size}              & \textbf{Output Size}             & \textbf{Kernel Size} \\ \midrule
\multirow{8}{*}{\textbf{Encoder}}  & \multirow{2}{*}{Down1}          & Conv3D + Norm + LeakyReLU          & P$^3\times $1           & P$^3\times $C           & 3$^3$       \\
                          &                                 & Dilated Conv3D + Norm + LeakyReLU  & P$^3\times $C           & (P/2)$^3\times $C       & 3$^3$       \\ \cmidrule{2-6} 
                          & \multirow{2}{*}{Down2}          & Conv3D + Norm + LeakyReLU          & (P/2)$^3\times $C       & (P/2)$^3\times $2C      & 3$^3$       \\
                          &                                 & Dilated Conv3D + Norm + LeakyReLU  & (P/2)$^3\times $2C      & (P/4)$^3\times $2C      & 3$^3$       \\ \cmidrule{2-6} 
                          & \multirow{2}{*}{Down3}          & Conv3D + Norm + LeakyReLU          & (P/4)$^3\times $2C      & (P/4)$^3\times $4C      & 3$^3$       \\
                          &                                 & Dilated Conv3D + Norm + LeakyReLU  & (P/4)$^3\times $4C      & (P/8)$^3\times $4C      & 3$^3$       \\ \cmidrule{2-6} 
                          & \multirow{2}{*}{Down4}          & Conv3D + Norm + LeakyReLU          & (P/8)$^3\times $4C      & (P/8)$^3\times $8C      & 3$^3$       \\
                          &                                 & Dilated Conv3D + Norm + LeakyReLU  & (P/8)$^3\times $8C      & (P/16)$^3\times $8C     & 3$^3$       \\ \midrule
\multirow{25}{*}{\textbf{Decoder}} & \multirow{2}{*}{CIG\_Attention} & Conv3D + Norm + LeakyReLU          & (P/16)$^3\times $(K+1)  & (P/16)$^3\times $4(K+1) & 3$^3$       \\
                          &                                 & Conv3D + Norm + Sigmoid            & (P/16)$^3\times $4(K+1) & (P/16)$^3\times $K      & 1$^3$       \\ \cmidrule{3-6} 
                          & {[}CIG\_Mu{]}*k                 & Conv3D                             & (P/16)$^3\times $8C     & (P/16)$^3\times $8C     & 1$^3$       \\ \cmidrule{3-6} 
                          & {[}CIG\_Sigma{]}*k              & Conv3D                             & (P/16)$^3\times $8C     & (P/16)$^3\times $8C     & 1$^3$       \\ \cmidrule{3-6} 
                          & \multirow{2}{*}{Up1}            & ConvTranspose3d + Norm + LeakyReLU & (P/16)$^3\times $16C    & (P/8)$^3\times $8C      & 3$^3$       \\ \cmidrule{3-6} 
                          &                                 & Conv3D + Norm + LeakyReLU          & (P/8)$^3\times $16C     & (P/8)$^3\times $8C      & 3$^3$       \\ \cmidrule{3-6} 
                          & \multirow{2}{*}{CIG\_Attention} & Conv3D + Norm + LeakyReLU          & (P/8)$^3\times $(K+1)   & (P/8)$^3\times $4(K+1)  & 3$^3$       \\
                          &                                 & Conv3D + Norm + Sigmoid            & (P/8)$^3\times $4(K+1)  & (P/8)$^3\times $K       & 1$^3$       \\ \cmidrule{3-6} 
                          & {[}CIG\_Mu{]}*k                 & Conv3D                             & (P/8)$^3\times $4C      & (P/8)$^3\times $4C      & 1$^3$       \\ \cmidrule{3-6} 
                          & {[}CIG\_Sigma{]}*k              & Conv3D                             & (P/8)$^3\times $4C      & (P/8)$^3\times $4C      & 1$^3$       \\ \cmidrule{3-6} 
                          & \multirow{2}{*}{Up2}            & ConvTranspose3d + Norm + LeakyReLU & (P/8)$^3\times $8C      & (P/4)$^3\times $4C      & 3$^3$       \\
                          &                                 & Conv3D + Norm + LeakyReLU          & (P/4)$^3\times $8C      & (P/4)$^3\times $4C      & 3$^3$       \\ \cmidrule{2-6} 
                          & \multirow{2}{*}{CIG\_Attention} & Conv3D + Norm + LeakyReLU          & (P/4)$^3\times $(K+1)   & (P/4)$^3\times $4(K+1)  & 3$^3$       \\
                          &                                 & Conv3D + Norm + Sigmoid            & (P/4)$^3\times $4(K+1)  & (P/4)$^3\times $K       & 1$^3$       \\ \cmidrule{3-6} 
                          & {[}CIG\_Mu{]}*k                 & Conv3D                             & (P/4)$^3\times $2C      & (P/4)$^3\times $2C      & 1$^3$       \\ \cmidrule{3-6} 
                          & {[}CIG\_Sigma{]}*k              & Conv3D                             & (P/4)$^3\times $2C      & (P/4)$^3\times $2C      & 1$^3$       \\ \cmidrule{3-6} 
                          & \multirow{2}{*}{Up3}            & ConvTranspose3d + Norm + LeakyReLU & (P/4)$^3\times $4C      & (P/2)$^3\times $2C      & 3$^3$       \\
                          &                                 & Conv3D + Norm + LeakyReLU          & (P/2)$^3\times $4C      & (P/2)$^3\times $2C      & 3$^3$       \\ \cmidrule{2-6} 
                          & \multirow{2}{*}{CIG\_Attention} & Conv3D + Norm + LeakyReLU          & (P/2)$^3\times $(K+1)   & (P/2)$^3\times $4(K+1)  & 3$^3$       \\
                          &                                 & Conv3D + Norm + Sigmoid            & (P/2)$^3\times $4(K+1)  & (P/2)$^3\times $K       & 1$^3$       \\ \cmidrule{3-6} 
                          & {[}CIG\_Mu{]}*k                 & Conv3D                             & (P/2)$^3\times $C       & (P/2)$^3\times $C       & 1$^3$       \\ \cmidrule{3-6} 
                          & {[}CIG\_Sigma{]}*k              & Conv3D                             & (P/2)$^3\times $C       & (P/2)$^3\times $C       & 1$^3$       \\ \cmidrule{3-6} 
                          & \multirow{2}{*}{Up4}            & ConvTranspose3d + Norm + LeakyReLU & (P/2)$^3\times $2C      & P$^3\times $C           & 3$^3$       \\
                          &                                 & Conv3D + Norm + LeakyReLU          & P$^3\times $2C          & P$^3\times $C           & 3$^3$       \\ \cmidrule{2-6} 
                          & Output                          & Conv3D                             & P$^3\times $C           & P$^3\times $O           & 3$^3$       \\ \bottomrule
\end{tabular}}
\end{table}

\subsection{Public Dataset and Evaluation Metrics}

\subsubsection{Datasets}
We evaluate the performance of the proposed CIML on both a demonstration task, named as \texttt{ShapeComposition}, and three publicly available datasets, namely \texttt{BraTS2020}~\citep{menze2014multimodal}, \texttt{autoPET}~\citep{gatidis2022whole}, and \texttt{MICCAI HECKTOR 2022}~\citep{oreiller2022head}. 
\texttt{BraTS2020} is a brain tumor segmentation dataset consisting of four different modalities: Flair, T1CE, T1, and T2, while the \texttt{autoPET} and \texttt{MICCAI HECKTOR 2022} datasets contain positron emission tomography (PET) and computed tomography (CT) images, respectively.

The \texttt{BraTS2020}  includes $369$ subjects for training, with three distinct regions targeted for segmentation: the whole tumor (WT), the tumor core (TC), and the enhancing tumor (ET), in addition to the background. 
The \texttt{autoPET} challenge is composed of $1,014$ studies obtained from the University Hospital Tübingen and is publicly accessible on TCIA. The challenge aims to segment the lesion region. 
The \texttt{MICCAI HECKTOR 2022} dataset consists of $524$ training cases collected from seven different centers, with the goal of segmenting images into two regions: background and lymph nodes (GTVn).
For all datasets, we analyze the performance of various methods via five-fold cross-validation. 

In addition, we manually decompose the segmentation task beforehand to investigate the effects of different assignments on the segmentation performance, which will be presented in the ablation study.
% As some modalities may not specifically correspond to some regions (i.e. T1 modality images are often used to provide structural information.), we assign them areas that are easy to segment accurately so that the corresponding sub-models can extract some accurate information.
For \texttt{BraTS2020}, we default set four sub-tasks, where FLAIR images are used to segment WT regions; T1 images are used to segment TC regions; T2 images are used to segment WT and TC regions; T1CE images are used to segment TC and ET regions.
Since \texttt{autoPET} and \texttt{MICCAI HECKTOR 2022} contain only one target region, we set two segmentors that predict the same target region in these challenges. 
In the final results, we ensemble the results of each segmentor by averaging the results if the same target region is present in multiple segmentors. 
As \texttt{BraTS2020} is widely used in the literature, we mainly focus on this dataset in our experiments.

\subsubsection{Evaluation Metrics}
The evaluation metrics in our experiments include the dice coefficient and $95\%$ Hausdorff distance (HD):
\begin{itemize}
    \item \textbf{Dice coefficient}~\citep{dice1945measures}: 
the dice coefficient measures the segmentation performance of CIML.
Concretely, the dice coefficient from set $X$ to set $Y$ is defined as: 

\begin{equation}
\rm{Dice}(X,Y)=\frac{
2 \cdot \Vert X \cap Y \Vert_1}
{\Vert X \Vert_1 + \Vert Y \Vert_1}.
\end{equation}
It is worth highlighting that higher dice coefficients imply that the predictions are closer to the ground truth, which indicates more accurate segmentation.

\item \textbf{HD$95$}~\citep{henrikson1999completeness}: 
The maximum Hausdorff distance is the maximum distance of a set to the nearest point in the other set. 
More formally, The maximum Hausdorff distance from set $X$ to set $Y$ is  defined as:
\begin{equation}
d_{\mathrm{H}}(X, Y)=\max \left\{\sup _{x \in X} d(x, Y), \sup _{y \in Y} d(y, X)\right\},
\end{equation}
where $\sup$ represents the supremum, $d(a, B)$ is the shortest Euclidean distance between point $a$ and set $B$.
\end{itemize}
% \noindent{1).} 

% \noindent{2).} 
% \subsubsection{Baselines}
% \textcolor{red}{xxxxxx.}

\subsection{Implementation Details}

We run all experiments based on Python $3.8$, PyTorch $1.12.1$, and Ubuntu $20.04$.
All training procedures are performed on a single NVIDIA A$100$ GPU with $40$GB memory.
The initial learning rate is set to $1e-4$, and we employ a “poly” decay strategy as below:
\begin{equation}
\label{poly}
\operatorname{lr}=\text { initial\_lr } \times\left(1-\frac{\text { epoch\_id }}{\text { max\_epoch }}\right)^{0.9}.
\end{equation}
We apply Adam as the optimizer with weight decay set to $3e-5$ and betas set to ($0.9$, $0.999$).
We trained our models using a maximum number of epochs set to $500$ (i.e., the max epoch in Equation~\ref{poly}) for the three public datasets and $1000$ for the demonstration experiment.  
Each epoch consists of $100$ iterations.
To enhance the generalization of our models, we applied the default augmentation strategy in nnUNet~\citep{isensee2018nnu} for the three public datasets.

The configurations of the segmentor for the three public datasets are presented in Table~\ref{tab: architecture}. 
The network architecture for the demonstration experiment is similar but has fewer filters, which are described in more detail in Section~\ref{subsec:demo}.
LeakyReLU activation with a negative slope of 0.01 is used in all experiments. Batch Normalization is used with a batch size of 2 in the \texttt{BraTS2020} dataset, while Instance Normalization is used in the other experiments. 
The patch size is set to $64\times 64\times 64$ for the \texttt{BraTS2020} dataset and $128\times 128\times 128$ for other datasets, unless otherwise specified. 
The notation $\textrm{P}$ refers to patch size, $\textrm{C}$ refers to the base number of filters and $\textrm{K}$ refers to the number of messages.

In our practical implementation, we utilize the sum of Dice loss~\citep{milletari2016v, drozdzal2016importance} and cross-entropy loss instead of exclusively employing cross-entropy loss. Our experimental results demonstrate that the former yields better outcomes.

\subsection{Demonstrated Task: \texttt{ShapeComposition}}
\label{subsec:demo}

\begin{figure}[!ht] 
  \centering
  \begin{minipage}[b]{\linewidth} 
  \captionsetup{justification=centering}
  \subfloat[Primary modality.]{
    \centering
    \begin{minipage}[b]{0.19\linewidth}
      \includegraphics[height=\linewidth]{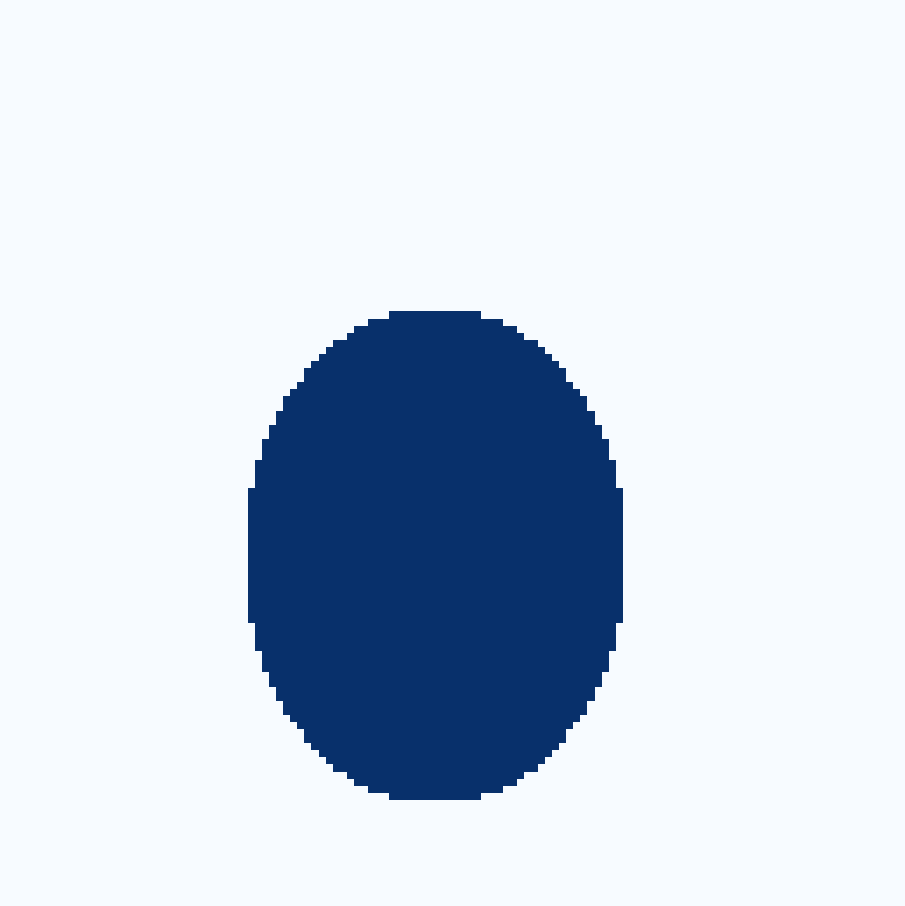}
      \includegraphics[height=\linewidth]{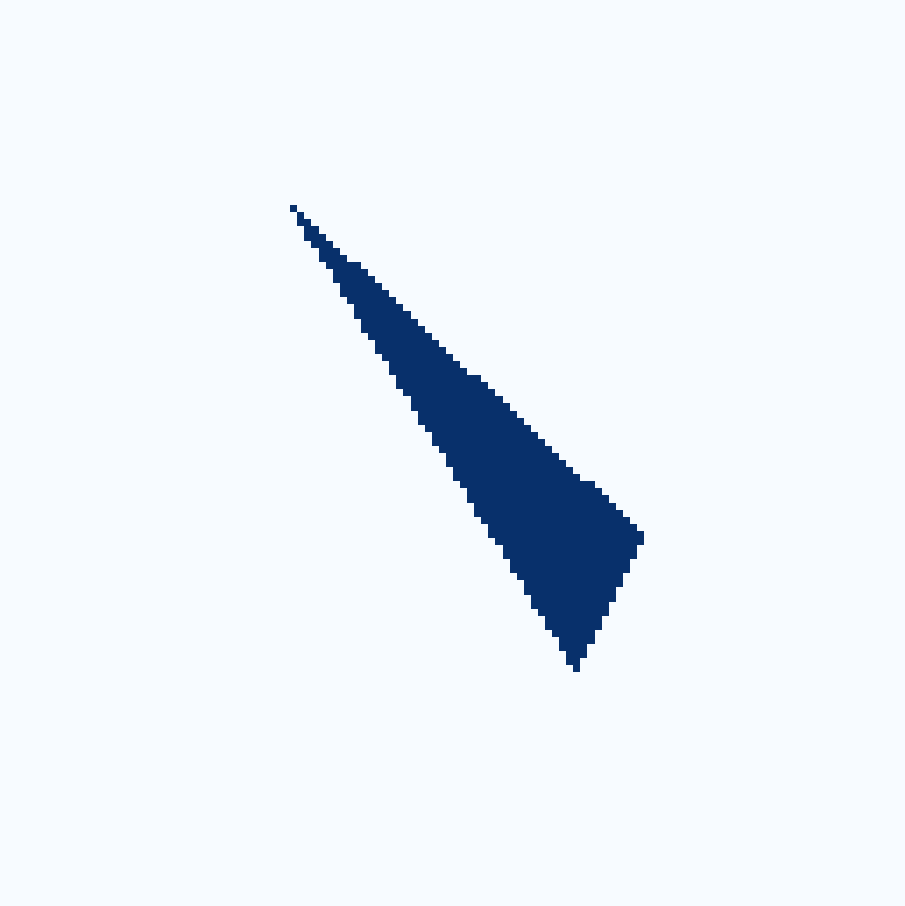}
      \includegraphics[height=\linewidth]{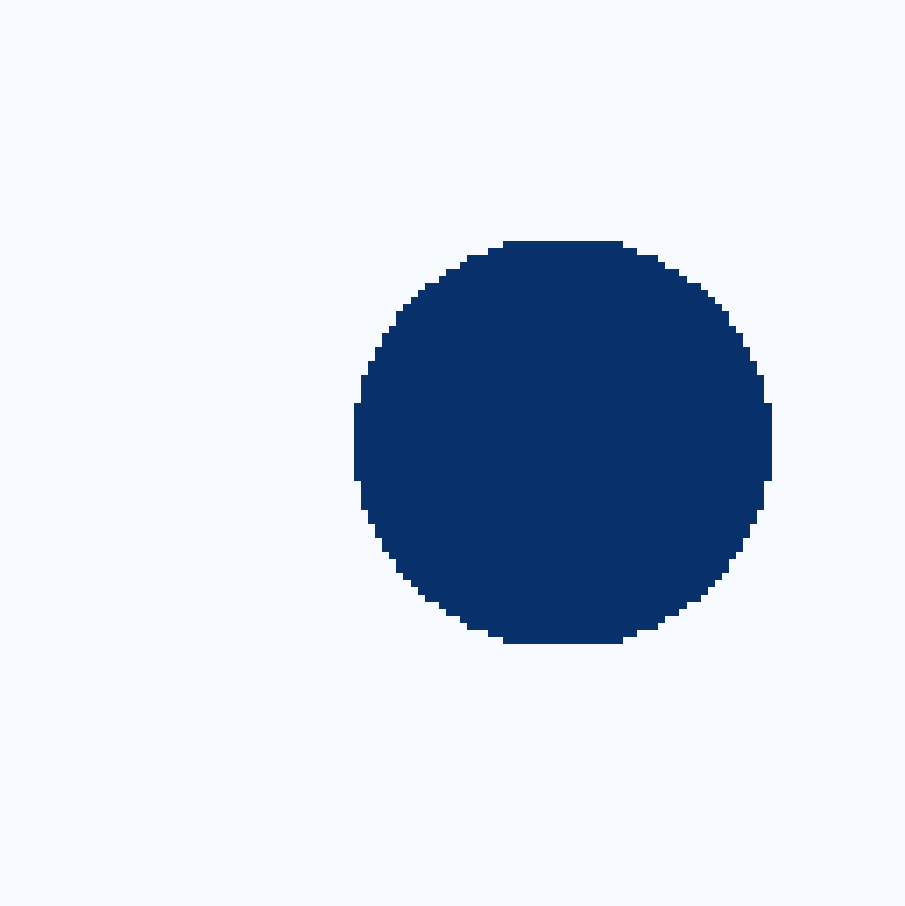}
      \includegraphics[height=\linewidth]{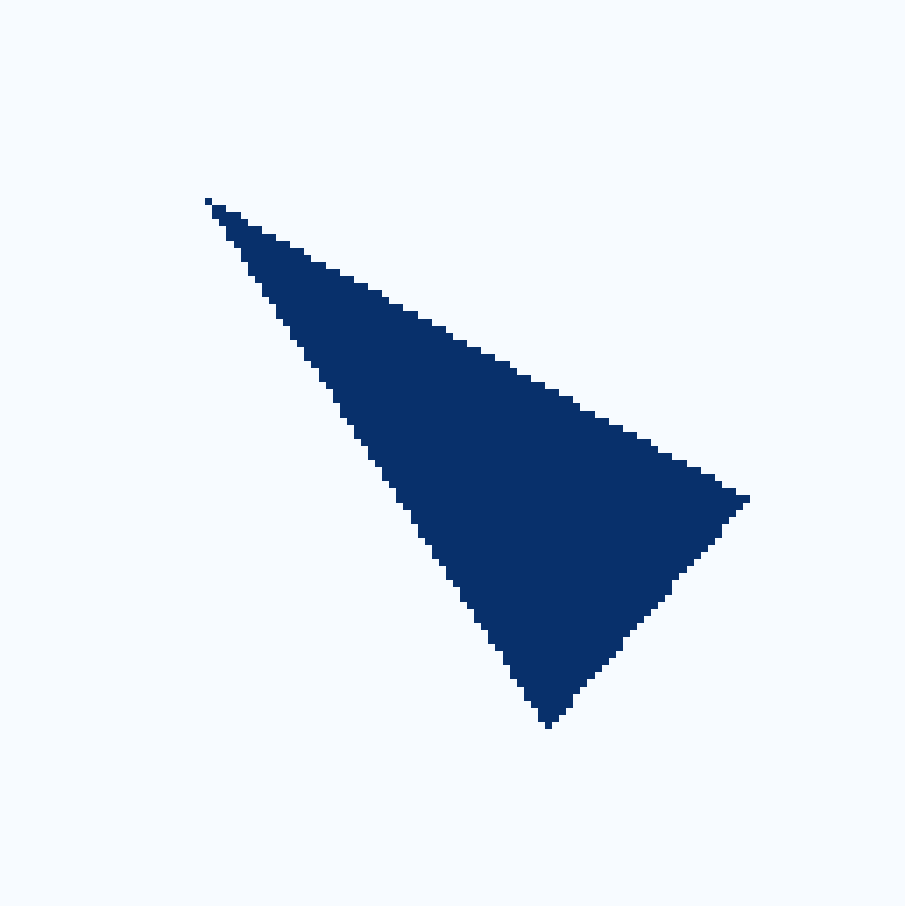}
    \end{minipage}
  } 
  \hspace{-0.25cm}
    \subfloat[Auxiliary modality.]{
    \centering
    \begin{minipage}[b]{0.19\linewidth}
      \includegraphics[height=\linewidth]{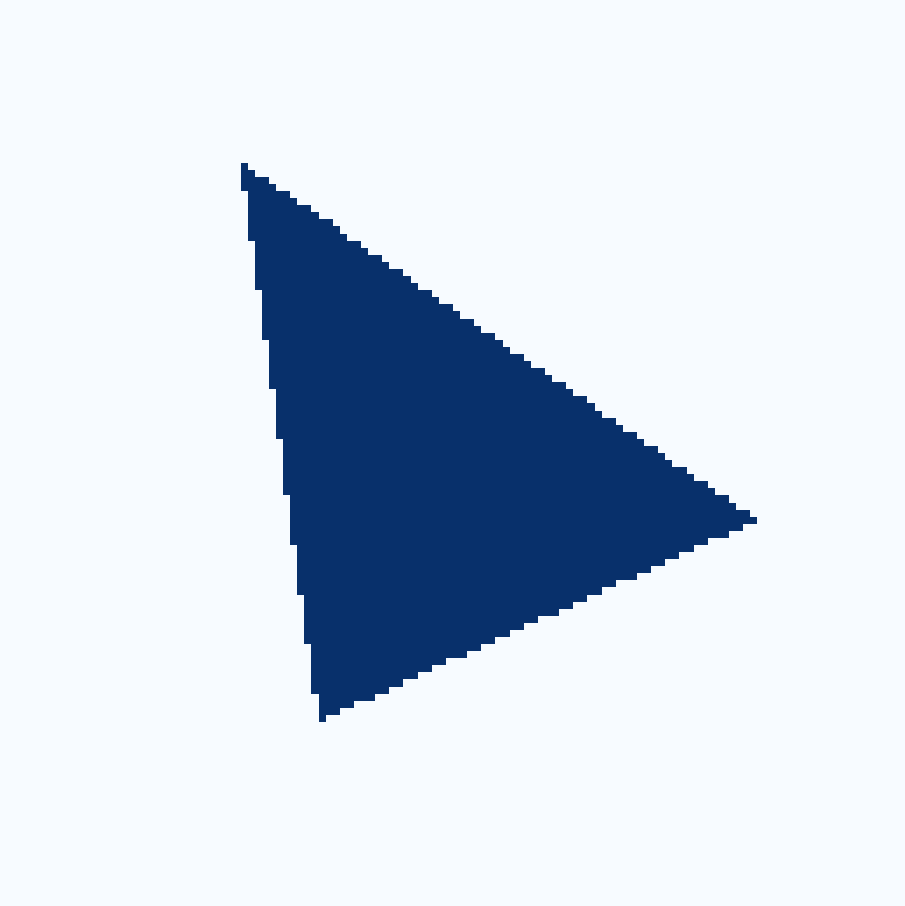}
      \includegraphics[height=\linewidth]{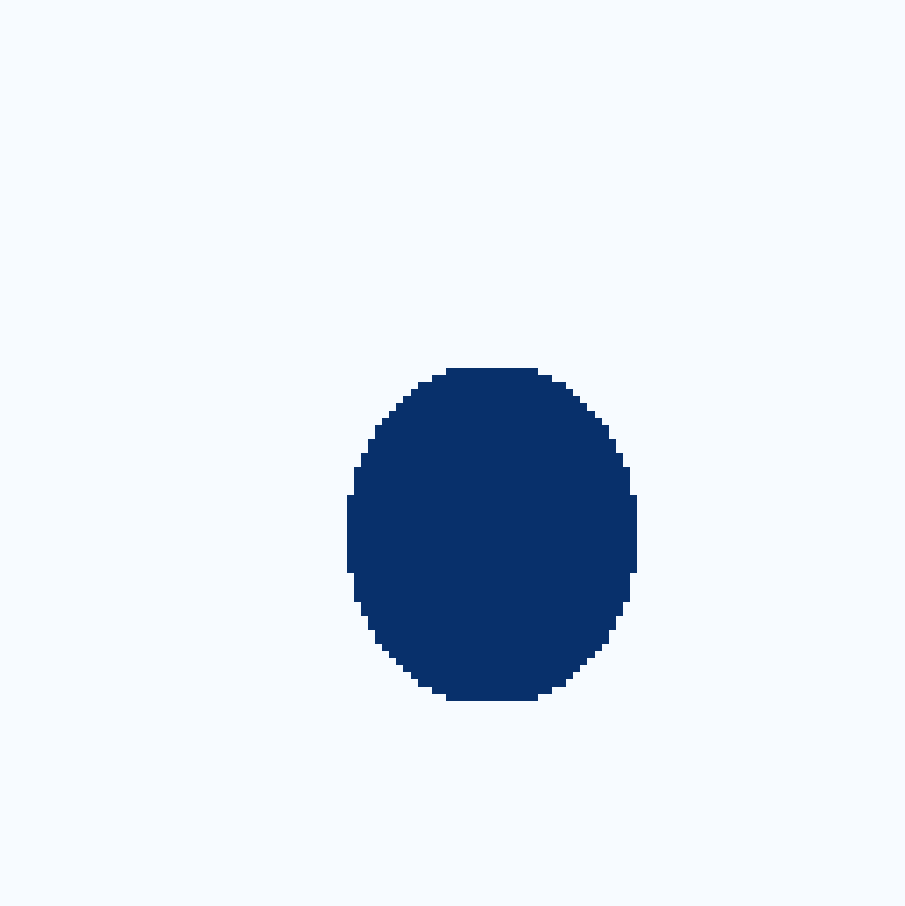}
      \includegraphics[height=\linewidth]{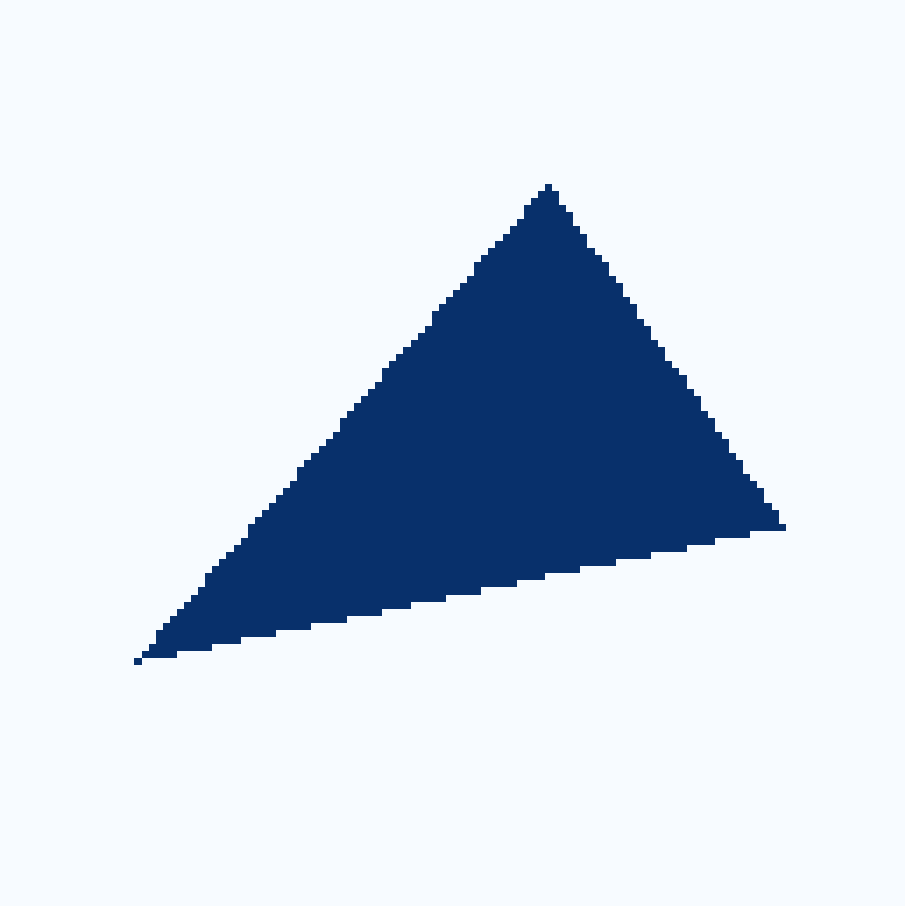}
      \includegraphics[height=\linewidth]{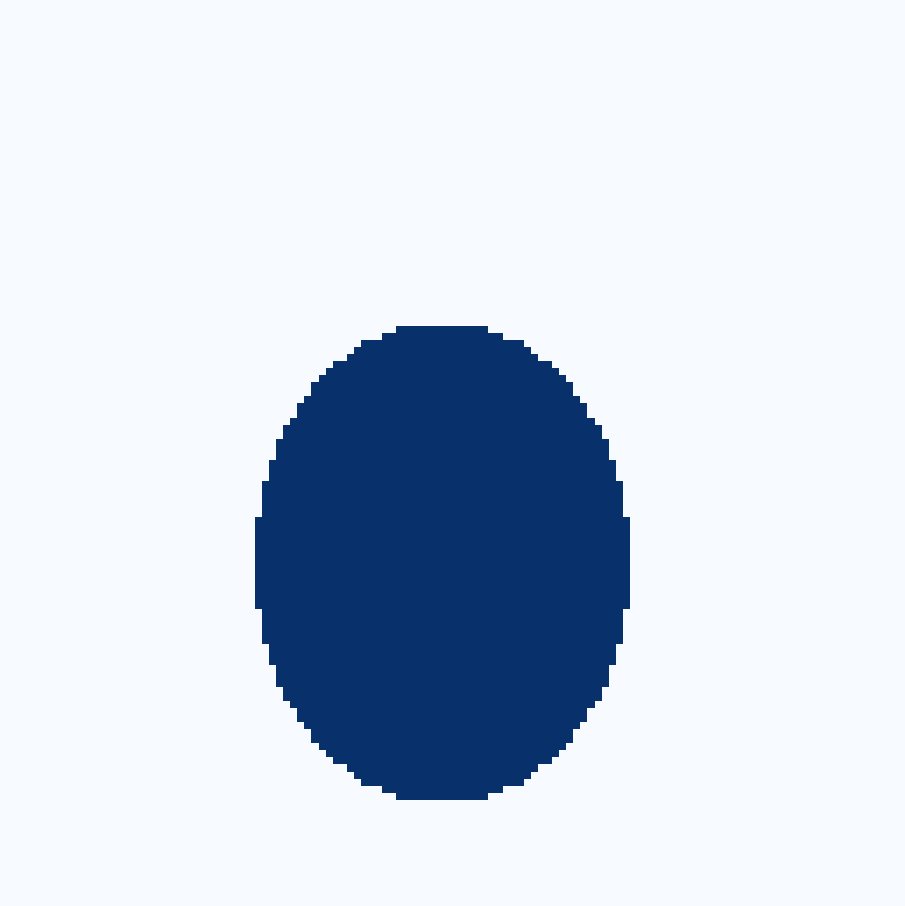}
    \end{minipage}
  }
  \hspace{-0.25cm}
    \subfloat[Predicted Segmentation.]{
    \centering
    \begin{minipage}[b]{0.19\linewidth}
      \includegraphics[height=\linewidth]{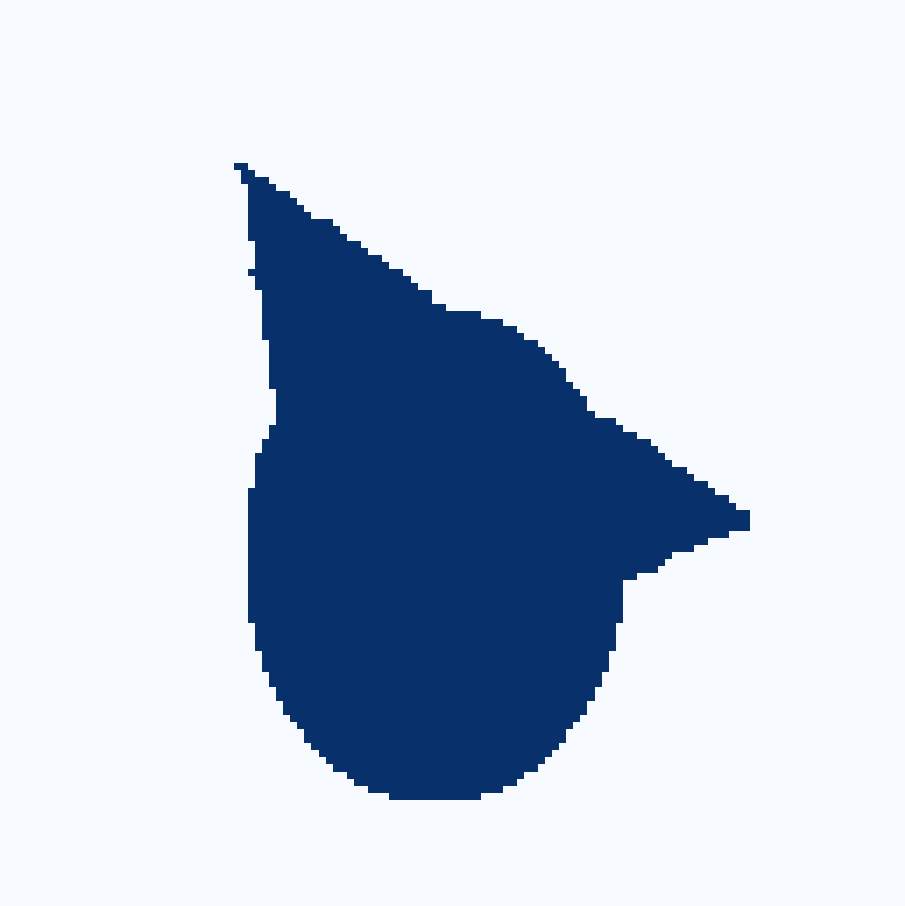}
      \includegraphics[height=\linewidth]{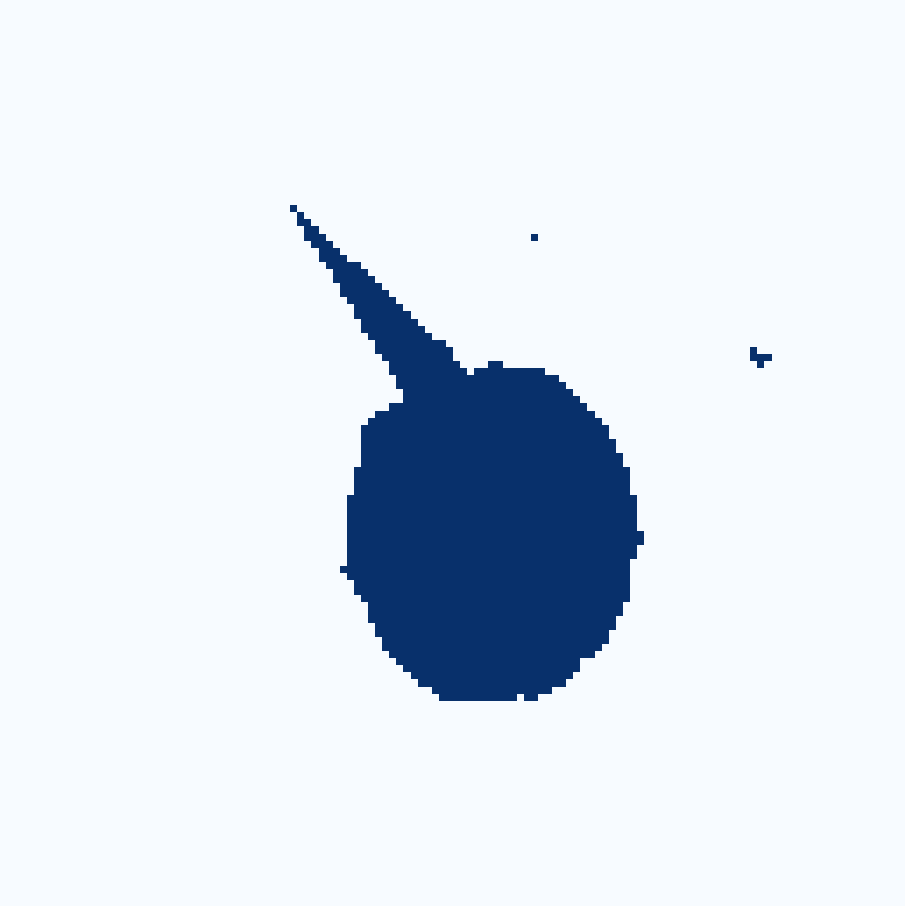}
      \includegraphics[height=\linewidth]{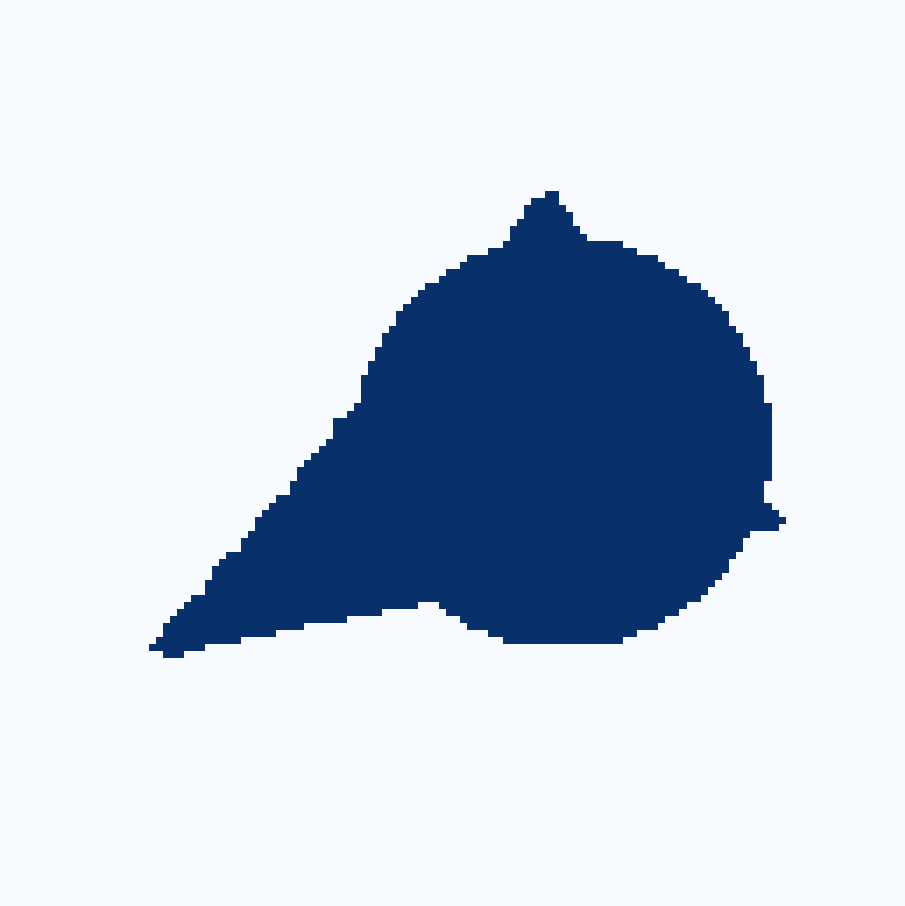}
      \includegraphics[height=\linewidth]{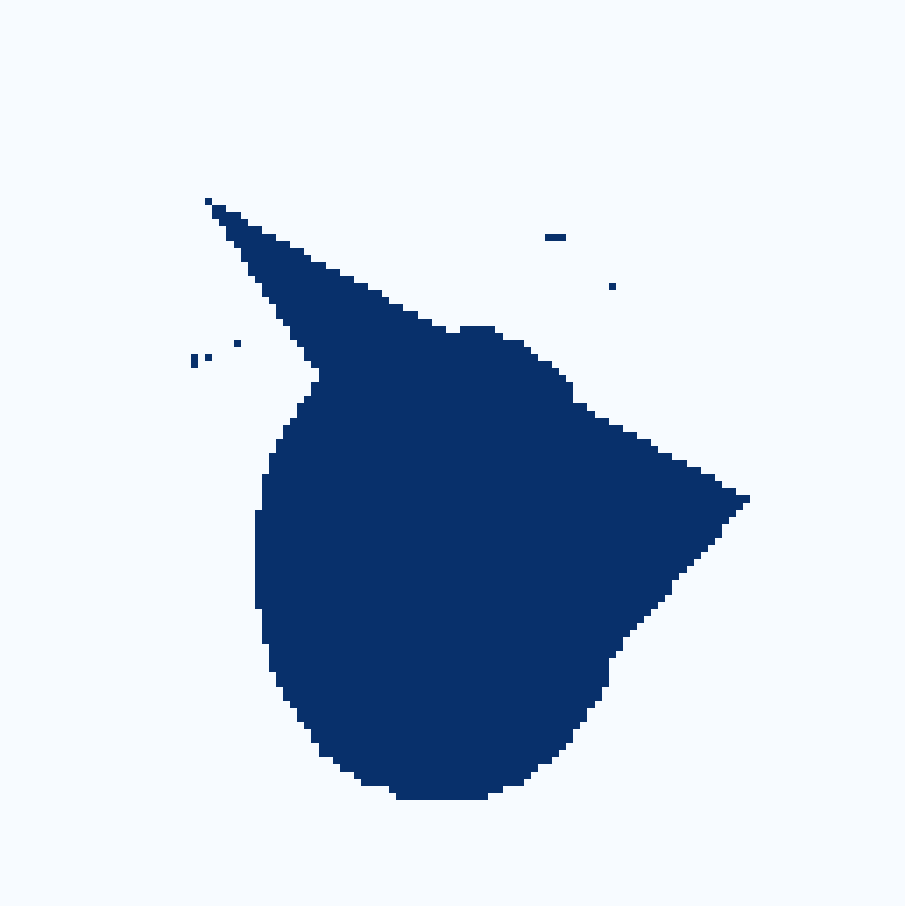}
    \end{minipage}
  }
  \hspace{-0.25cm}
    \subfloat[Ground Truth.]{
    \centering
    \begin{minipage}[b]{0.19\linewidth}
      \includegraphics[height=\linewidth]{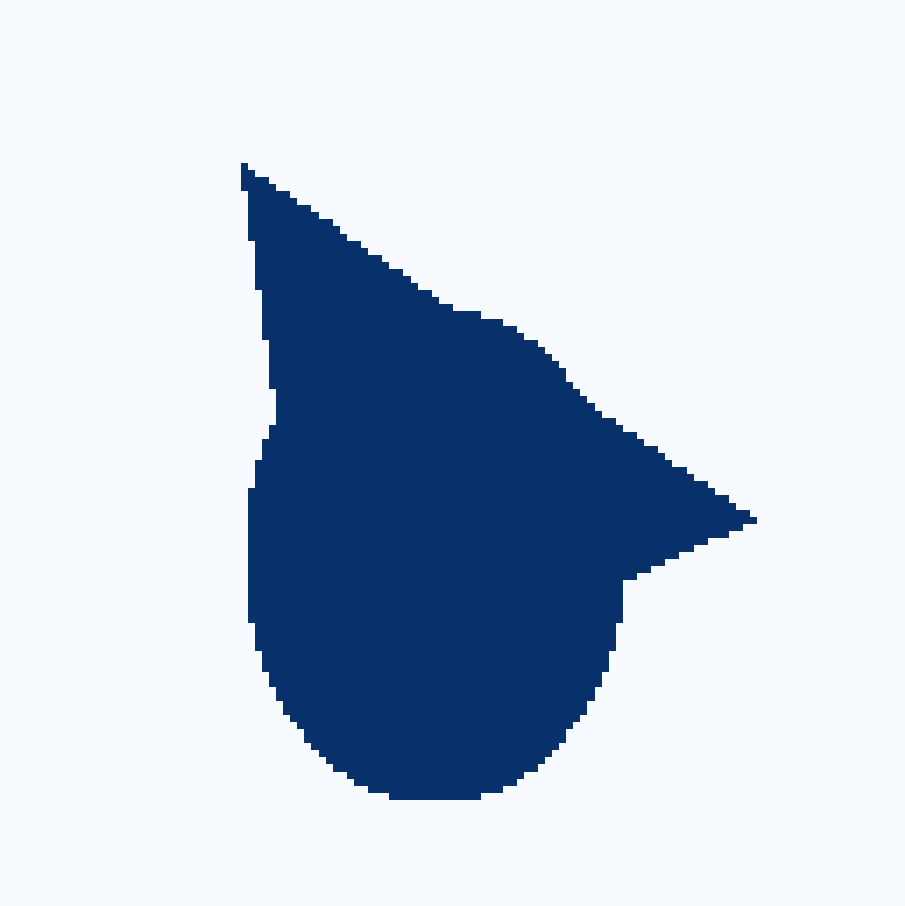}
      \includegraphics[height=\linewidth]{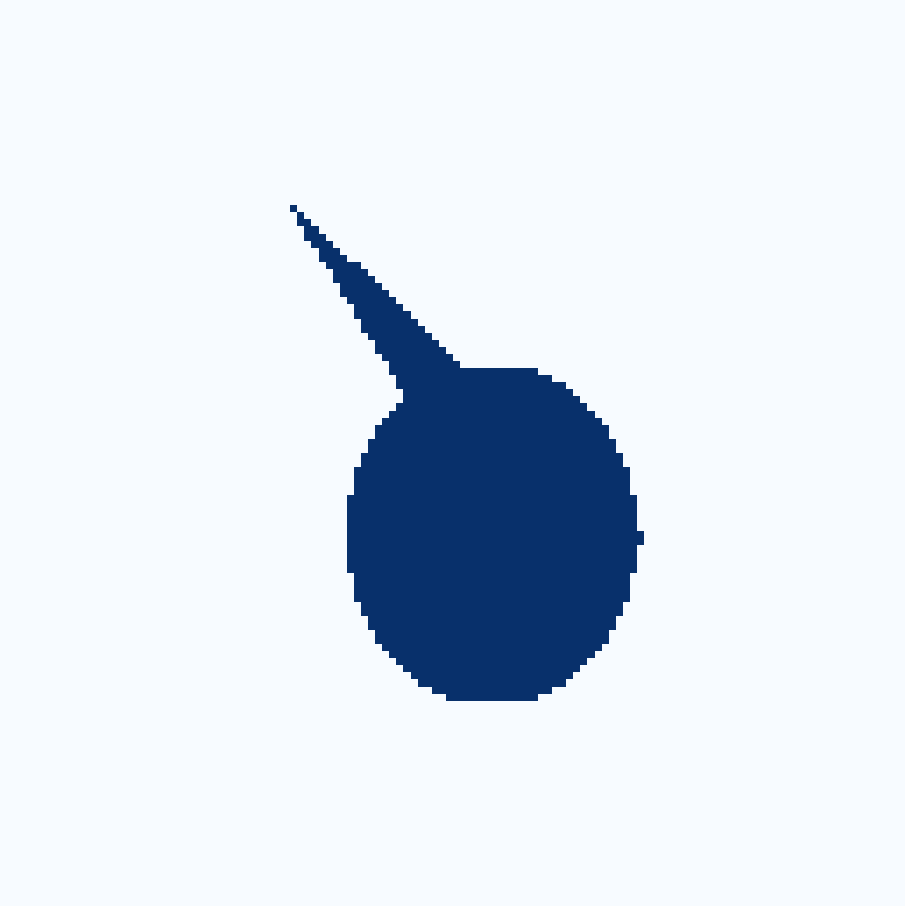}
      \includegraphics[height=\linewidth]{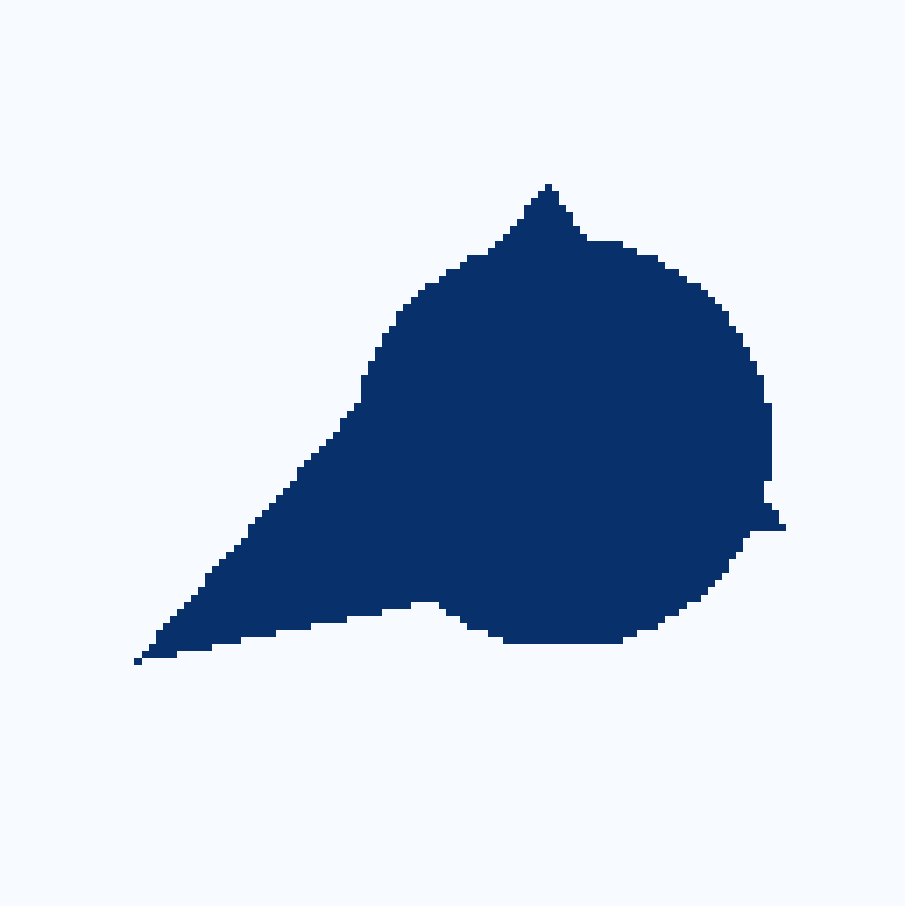}
      \includegraphics[height=\linewidth]{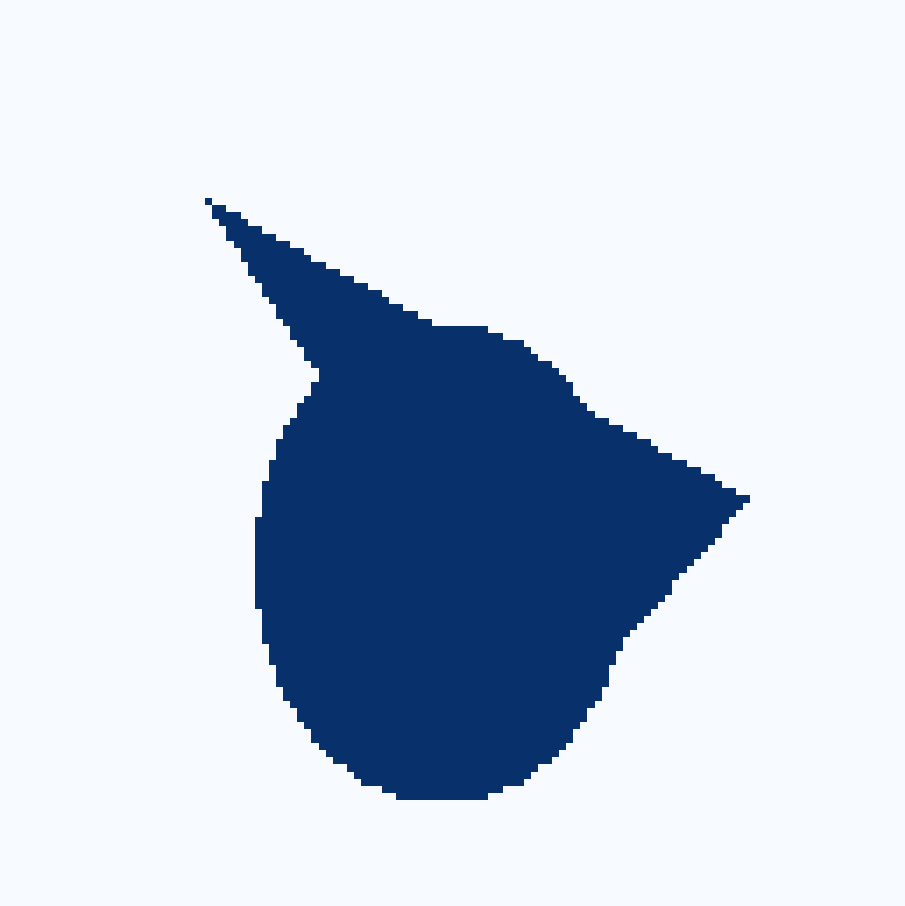}
    \end{minipage}
  }
  \hspace{-0.25 cm}
    \subfloat[Complementary Information.]{
    \centering
    \begin{minipage}[b]{0.19\linewidth}
      \includegraphics[height=\linewidth]{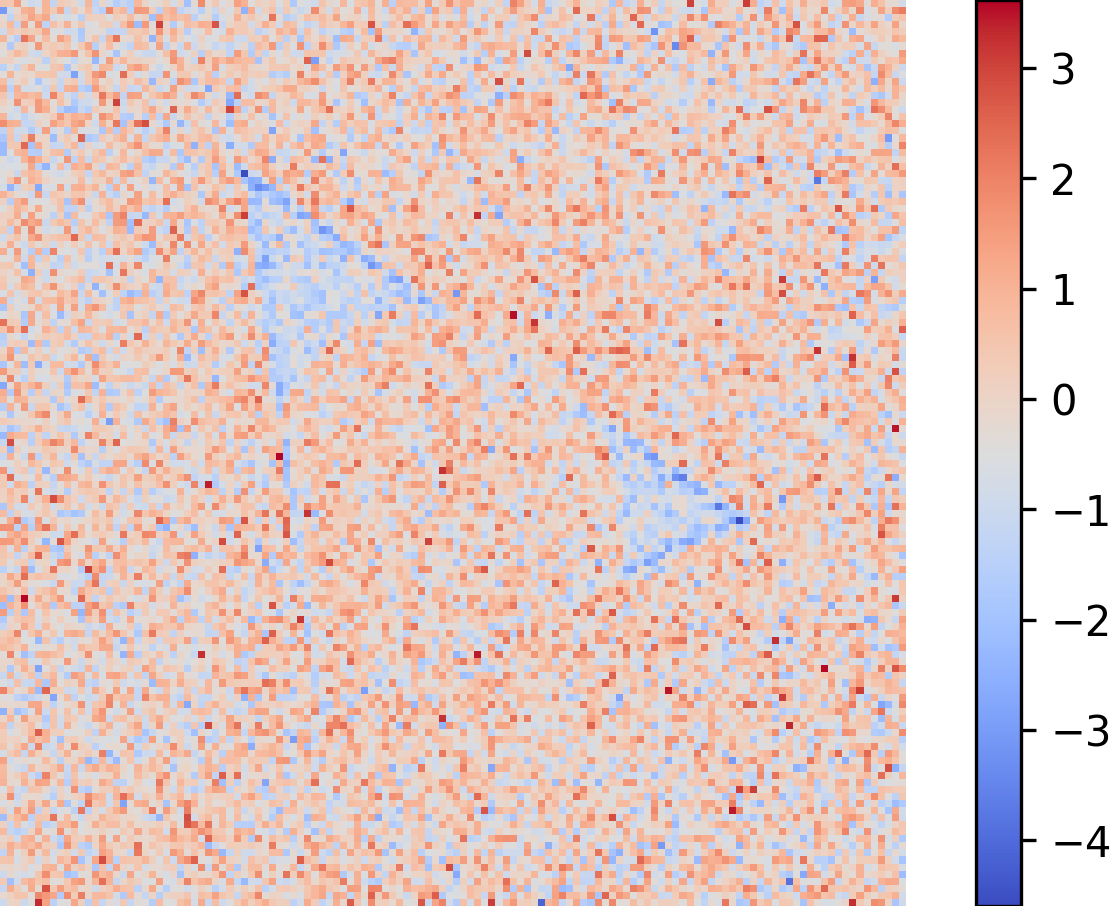}
      \includegraphics[height=\linewidth]{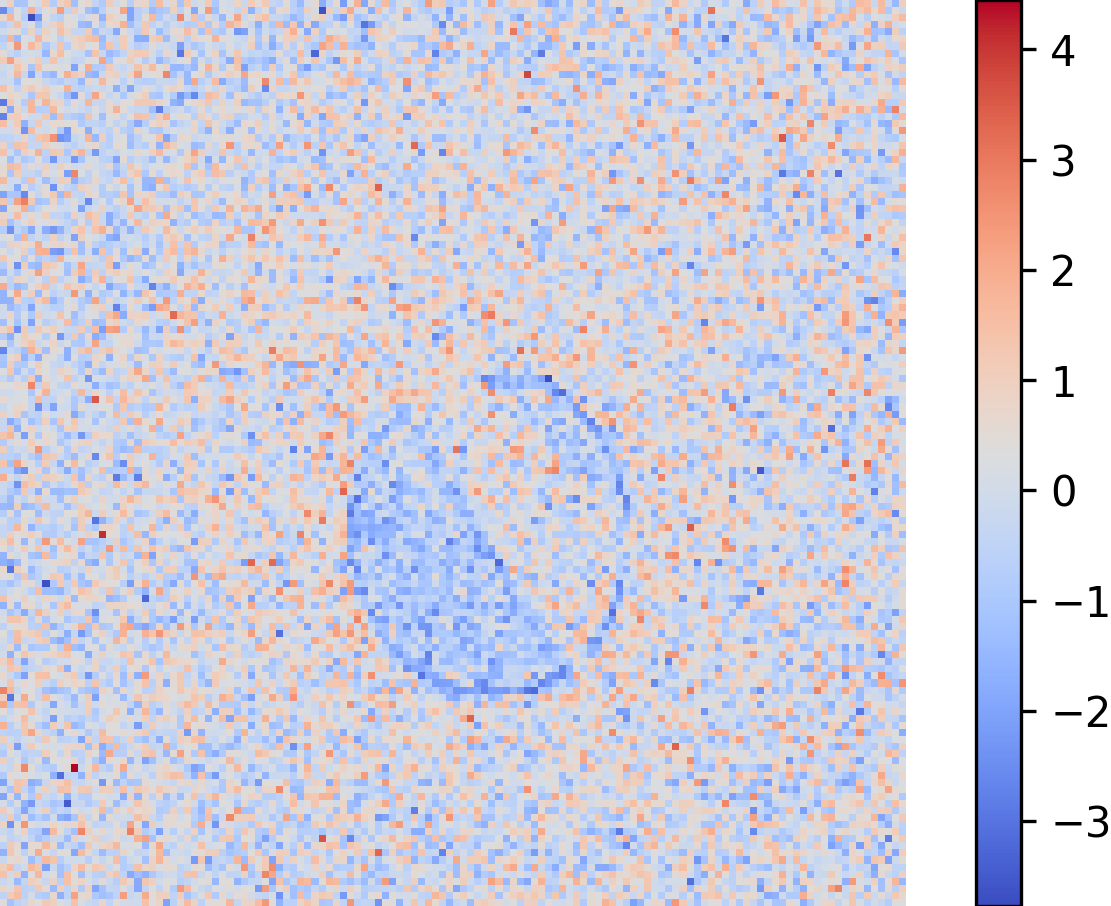}
      \includegraphics[height=\linewidth]{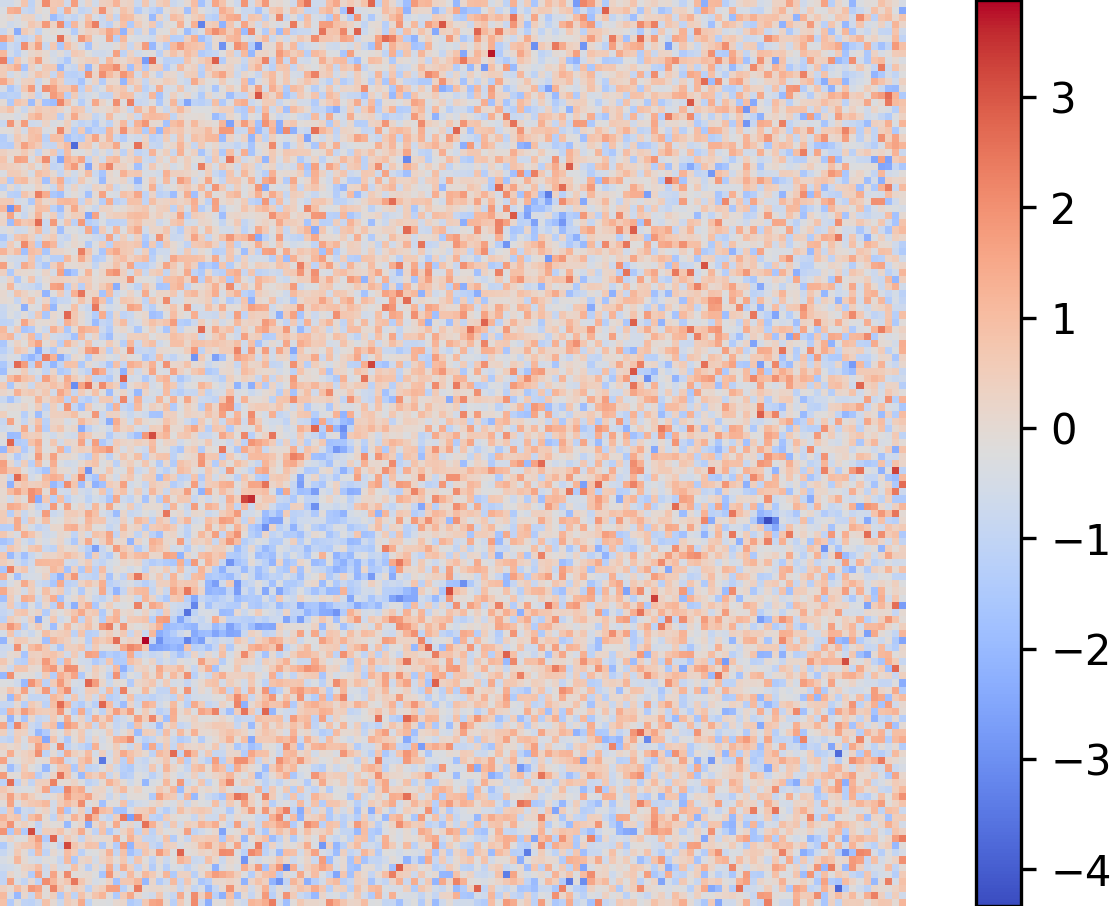}
      \includegraphics[height=\linewidth]{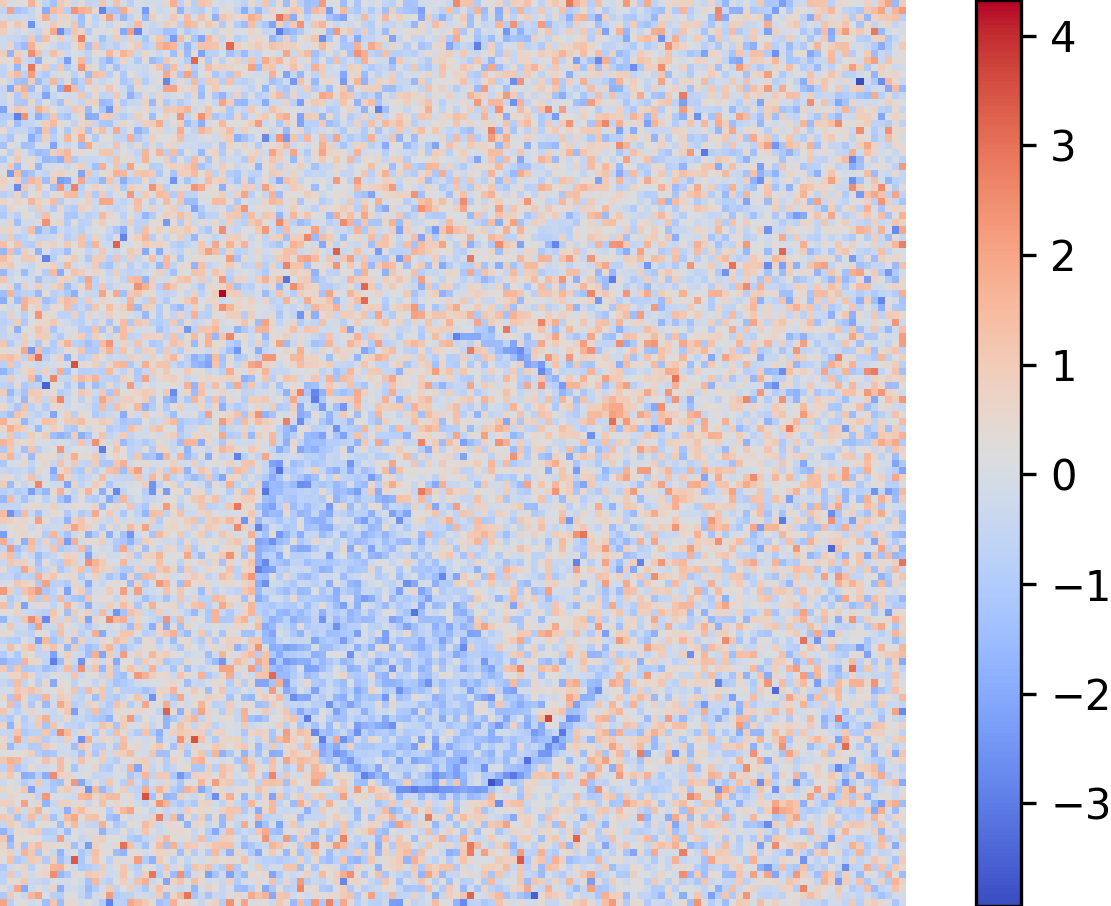}
    \end{minipage}
  }
  \end{minipage}
  \caption{Each row displays the original images and results from a single case. The first and second columns depict the primary and auxiliary modalities, respectively, with the first and second images in the set. The third column shows the predicted segmentation, while the fourth column presents the ground-truth segmentation. The fifth column displays the visualization of the complementary information. \blue{Note: the images are best viewed in color for optimal clarity.}}
  \label{fig:demo}
\end{figure}

To evaluate the effectiveness of our proposed redundancy filtering, we generated an artificial dataset containing 1000 sets of images. Each set consists of a triangle and an ellipse, deliberately overlapping in a specific region. 
The aim of the task is to take the input set of images and generate their union as the output, as illustrated in Figure~\ref{fig:demo}. 
We employed the task decomposition approach in the CIML framework and assigned one of the figures as the primary modality and the other as the auxiliary modality. 
Note that both triangles and ellipses can be considered the primary modality.

In the implementation, we make several simplifications to the network architecture. 
Only one sub-network corresponds to the segmentation, and the other sub-network is utilized to acquire complementary information. 
Specifically, two encoders are employed independently to extract from the primary and auxiliary modalities, respectively. 
Two decoding pathways are then used. 
In the first path, one decoder inputs primary modality features (without gradient backpropagation) and auxiliary modality features, outputting $\mu$ and $\sigma$. 
3D convolution layers without CIG modules are utilized to fuse features, thus eliminating the effect of CIG modules to verify the efficacy of our complementary information learning. 
Then, the reparameterization trick is used to sample complementary information features. 
In the second path, complementary information features are combined with the features directly extracted from the primary modality to predict the final results. 
Furthermore, as described in Section~\ref{ssec:MPRF}, the Kullback–Leibler divergence between the complementary information features and the standard normal distribution is minimized to constrain the complementary information features containing less information from the primary modality.

For qualitative analysis, we maintain the dimension of the complementary information features in line with the original image. 
As depicted in Figure~\ref{fig:demo}, our proposed redundancy filtering-based complementary information extraction is effective, and the network efficiently extracts information from the auxiliary modality that is not present in the primary modality. 
Additionally, it contains little information that is already included in the primary modality.

\subsection{Standardized Benchmarks}%: \texttt{BraTS2020}, \texttt{autoPET} and \texttt{MICCAI HECKTOR 2022}}
\begin{table}[!htbp]
\centering
\caption{Quantitative comparison with SOTA methods on BraTS2020. 
WT, TC and ET signify the Dice coefficients of the whole tumor, the tumor core and the enhancing tumor, respectively.
We use the results of RFNet and DIGEST from their paper. Other results are the result of reproducing them.}
\label{tab: bratssota}
\resizebox{\linewidth}{!}{
\begin{tabular}{@{}l|ll|cccc|cccc@{}}
\toprule
\multirow{2}{*}{Methods Type} & \multirow{2}{*}{Methods} & \multirow{2}{*}{Patch Size} & \multicolumn{4}{c|}{Dice$\uparrow$ \%} & \multicolumn{4}{c}{HD95$\downarrow$} \\
 &  &  & WT & TC & ET & MEAN & WT & TC & ET & MEAN \\ \midrule
\multirow{3}{*}{\begin{tabular}[c]{@{}l@{}}General \\ Methods\end{tabular}} & nnUNet~\citep{isensee2018nnu} & $128^3$ & 91.59 & 87.50 & 83.47 & 87.52 & 4.75 & 4.0 8 & 5.60 & 4.81 \\
 & AttentionUNet~\citep{oktay2018attention} & $128^3$ & 90.62 & 86.33 & 81.32 & 86.09 & 5.42 & 7.88 & 8.98 & 7.43 \\
 & UNETR~\citep{hatamizadeh2022unetr} & $128^3$ & 91.11 & 86.42 & 82.96 & 86.76 & 7.97 & 5.16 & 5.90 & 6.34 \\ \midrule
\multirow{7}{*}{\begin{tabular}[c]{@{}l@{}}Multi-modal  \\ Methods\end{tabular}} 
& MAML~\citep{zhang2021modality} & $128^3$ & 91.40 & 88.05 & 82.40 & 87.28 & 4.84 & 5.95 & 7.90 & 7.82 \\
 & DIGEST~\citep{li2022digest} & $128^3$ & 90.20 & 87.00 & 81.20 & 86.17 & / & / & / & / \\
 & RFNet~\citep{ding2021rfnet} & $128^3$ & 91.11 & 85.21 & 78.00 & 84.77 & / & / & / & / \\
 & ACMINet~\citep{zhuang20223d} & $128^3$ & 91.79 & 87.99 & 82.56 & 87.45 & 5.70 & 5.09 & 6.95 & 5.91 \\
 & NestedFormer~\citep{xing2022nestedformer} & $128^3$ & 91.76 & 88.20 & 83.19 & 87.72 & 5.35 & 5.07 & 7.16 & 5.86 \\
 & CIML(Ours) & $64^3$ & 91.60 & \textbf{89.14} & 83.91 & 88.21 & 5.83 & \textbf{3.70} & \textbf{4.95} & 4.83 \\
 & CIML(Ours) & $128^3$ & \textbf{91.88} & 88.69 & \textbf{84.34} & \textbf{88.30} & \textbf{4.58} & 4.12 & 5.23 & \textbf{4.64} \\ \bottomrule
\end{tabular}}
\end{table}
\begin{table}[!htbp]
\caption{Quantitative comparison with SOTA methods on the autoPET challenge. All outcomes of other algorithms result from reproducing them.}
\label{tab: autopetsota}
\centering
\resizebox{0.7\linewidth}{!}{
\begin{tabular}{l|l|c|c}
\toprule
Methods Type                                                                    & Methods & Dice $\uparrow$ \% & HD95 $\downarrow$ \\ \midrule
\multirow{3}{*}{\begin{tabular}[c]{@{}l@{}}General \\ Methods\end{tabular}}     & nnUNet~\citep{isensee2018nnu}  & 55.23              & 139.84            \\
 & AttentionUNet~\citep{oktay2018attention} & 57.87          & 108.62 \\
 & UNETR~\citep{hatamizadeh2022unetr}         & 41.62          & 177.04         \\ \midrule
\multirow{4}{*}{\begin{tabular}[c]{@{}l@{}}Multi-modal \\ Methods\end{tabular}} 
& MAML~\citep{zhang2021modality}    & 53.9               & 194.36            \\
 & ACMINet~\citep{zhuang20223d}        & 45.67          & 121.81         \\
 & NestedFormer~\citep{xing2022nestedformer}  & 52.51          & 183.14         \\
 & CIML(Ours)    & \textbf{61.37} & \textbf{107.58}         \\ \bottomrule
\end{tabular}}
\end{table}
\begin{table}[!htbp]
\caption{Quantitative comparison with SOTA methods on  MICCAI HECKTOR 2022. All outcomes of other algorithms resulting from reproducing them.}
\label{tab: hectorsota}
\centering
\resizebox{0.7\linewidth}{!}{
\begin{tabular}{l|l|c|c}
\toprule
Methods Type                                                                    & Methods & Dice $\uparrow$ \% & HD95 $\downarrow$ \\ \midrule
\multirow{3}{*}{\begin{tabular}[c]{@{}l@{}}General \\ Methods\end{tabular}}     & nnUNet~\citep{isensee2018nnu}  & 73.61              & 6.09              \\
 & AttentionUNet~\citep{oktay2018attention} & 72.30          & 16.13         \\
 & UNETR~\citep{hatamizadeh2022unetr}         & 74.50          & 5.85          \\ \midrule
\multirow{4}{*}{\begin{tabular}[c]{@{}l@{}}Multi-modal \\ Methods\end{tabular}} & MAML ~\citep{zhang2021modality}   & 70.00              & 29.05             \\
 & ACMINet~\citep{zhuang20223d}        & 74.23          & 8.13          \\
 & NestedFormer~\citep{xing2022nestedformer}  & 75.16          & 6.09          \\
 & CIML(Ours)    & \textbf{76.28} & \textbf{5.17} \\ \bottomrule
\end{tabular}}
\end{table}
% \caption{Analysis results of the proposed method on the BraTS2020 training data via five-fold cross-validation.}

In this section, we compared our proposed CIML algorithms to eight state-of-the-art segmentation methods, including nnUNet~\citep{isensee2018nnu}, AttentionUNet~\citep{oktay2018attention}, UNETR~\citep{hatamizadeh2022unetr}, MAML~\citep{zhang2021modality}, DIGEST~\citep{li2022digest}, RFNet~\citep{ding2021rfnet}, ACMINet~\citep{zhuang20223d} and NestedFormer~\citep{xing2022nestedformer}.
The first three methods are general methods, while the last five methods are designed specifically for multimodal segmentation.
The nnUNet is a widely used benchmark that simplifies the critical decisions in designing an effective segmentation pipeline for any dataset. 
It and its variant, AttentionUNet, both employ early fusion. 
UNETR employs a transformer as the encoder to learn sequence representations of input images. 
MAML utilizes modality-specific encoders and incorporates a cross-modality attention mechanism for information fusion. 
DIGEST is a method applying a deeply supervised knowledge
transfer network learning.
RFNet also uses specific encoders like MAML and includes a region-aware module. 
ACMINet proposes a volumetric feature alignment module to align early features with late features.
NestedFormer is a transformer-based method that designs a nested modality-aware feature aggregation module to model intra- and intermodality features for multimodal fusion.

Since RFNet and DIGEST are not open-source algorithms, we only compare CIML with them in the \texttt{BraTS2020} dataset and offer the results from the original studies.
Precisely, we reproduce other methods using open-sourced codes. 
For a fair comparison, we employ the same training set, i.e., the same batch size, training epochs, and learning rate decay mechanism for all approaches.

As reported in Table~\ref{tab: bratssota}, \ref{tab: autopetsota}, and \ref{tab: hectorsota}, our proposed method demonstrates superior performance compared to other methods, achieving the highest dice score and HD95 score in all regions across the three challenges.
For the \texttt{BraTS2020} dataset, we investigate the optimal performance of all models by using a patch size of $128\times 128 \times 128$. 
Even with a smaller patch size, which means a more limited field of view, our method still outperforms most existing methods.
Our method achieves higher segmentation accuracy compared to the state-of-the-art (SOTA) method, indicating that our algorithm can effectively eliminate the negative impact of \textit{inter-modal} redundant information.

\subsection{Ablation Study} 
\label{abalation}
\begin{table}[!htbp]
\caption{Ablation experiments on various components. The average Dice scores and HD95 scores are reported.
"Message": models transport embeddings as messages; 
"Attention": cross-modal attention mechanism;
"VI": conditional mutual information constrain;}
\label{tab: ablation}
\centering
\resizebox{0.9\linewidth}{!}{
\begin{tabular}{l|cccc|cccc}
\toprule
\multirow{2}{*}{Methods}                          & \multicolumn{4}{c|}{Dice $\uparrow$\%}                                        & \multicolumn{4}{c}{HD95 $\downarrow$}                                     \\
                                                 & WT            & TC             & ET             & MEAN           & WT            & TC           & ET            & MEAN          \\ \midrule
Baseline                                         & 88.08         & 86.54          & 83.78          & 86.13          & 13.13         & 4.70          & 5.31          & 7.71          \\
\midrule
+ Message                                     & 90.30         & 88.37          & 81.38          & 86.68          & 8.71          & 6.22         & 6.63          & 7.18          \\
+ Message + Attention                        & 90.43         & 87.23          & 82.93          & 86.86          & 6.47          & 5.26         & 7.49          & 6.08          \\
+ Message + VI           & 91.54         & 88.70           & 83.72          & 87.99          & \textbf{4.60} & 3.96         & 4.97          & \textbf{4.51} \\
+ Message + Attention + VI & \textbf{91.60} & \textbf{89.14} & \textbf{83.91} & \textbf{88.21} & 5.83          & \textbf{3.70} & \textbf{4.95} & 4.83          \\ \bottomrule
\end{tabular}}
\end{table}
\begin{table}[!htbp]
\caption{Ablation study on various task decomposition. The average Dice scores and HD95 scores are reported.
The results are listed from smallest to largest, according to the MEAN Dice.}
\label{tab: assignment}
\centering
\resizebox{0.9\linewidth}{!}{
\begin{tabular}{llll|cccc|cccc}
\toprule
\multicolumn{4}{c|}{Assignment}  & \multicolumn{4}{c|}{Dice $\uparrow$\%}                              & \multicolumn{4}{c}{HD95 $\downarrow$}                         \\
FLAIR & T1     & T2     & T1CE   & WT             & TC             & ET             & MEAN           & WT            & TC            & ET            & MEAN          \\ 
\midrule
TC    & TC, ET & WT     & TC     & 91.01          & 88.95          & 81.01          & 86.99          & 6.01          & 4.55          & 6.85          & 5.80           \\
WT    & WT     & TC     & TC, ET & 91.13          & 88.96          & 82.38          & 87.49          & 5.16          & 5.34          & 5.13          & 5.21          \\
WT    & WT, TC & TC     & ET     & 91.21          & 89.12          & 82.41          & 87.58          & \textbf{5.08} & 6.18          & 5.24          & 5.50s           \\
WT    & WT, TC & WT, TC & TC, ET & \textbf{91.62} & 88.20          & 83.79          & 87.87          & 5.98          & 3.97          & 5.17          & 5.04          \\
WT    & TC     & WT, TC & TC, ET & 91.60          & \textbf{89.14} & \textbf{83.91} & \textbf{88.21} & 5.83          & \textbf{3.70} & \textbf{4.95} & \textbf{4.83} \\
\bottomrule
\end{tabular}
}
\end{table}
\subsubsection{Importance of Different Components}
The efficacy of the proposed CIML segmentation method relies on several crucial components. 
An ablation study is conducted to evaluate the importance of each component and validate the efficiency of redundancy elimination, as shown in Table~\ref{tab: ablation}. 
The ``Baseline" approach utilizes the nnUNet model for segmentation for each segmentor without incorporating information transport, resulting in each sub-model only being able to extract local information. 
The ``+Message" notation represents the message passing between sub-models, where the local embeddings are used as the message and fed into a one-layer 3D convolutional network equipped with a Batch Normalization layer and a sigmoid activation function, serving as a basic fusion module. 
The ``+Attention" notation signifies the utilization of cross-model spatial attention mechanisms for integrating relevant information from the messages into the local embeddings. 
Finally, the ``+VI" symbol represents the implementation of the variational inference for complementary information learning to extract complementary information from the messages.

Table~\ref{tab: ablation} shows that the proposed message passing, cross-modal spatial attention mechanism, and complementary information learning significantly improved the model's performance.
Specifically, the ``baseline" approach that only utilizes local observations performs the worst in terms of the MEAN dice score. 
Message passing enhances the dice score on the WT and TC regions, as the input to the model contains complete information. 
Conversely, for the ET region, the opposite effect is observed. 
This is because the T1CE modality primarily determines the boundary of the ET region, and the features of other modalities do not contribute to determining the boundary of the ET region. 
This supports the rationale for the task decomposition in our framework. 
Additionally, the utilization of the cross-model spatial attention mechanism and our proposed redundancy filtering can significantly enhance the ability to extract relevant information from messages, resulting in better performance in the dice score and HD95. 
Notably, the implementation of our proposed redundancy filtering results in an even greater improvement. 
Combining the cross-modal spatial attention mechanism with the redundancy filtering yields optimal results.

\subsubsection{Comparison of Different Task Decomposition}
Based on our proposed CIML framework, the task decomposition is flexible. Different modalities have different clinical implications. 
On brain tumour segmentation, ET regions can be clearly discriminated on T1CE, and FLAIR is easier to discriminate WT regions, which have a clearer correspondence with the corresponding regions; 
TC regions have a relatively vague correspondence, so we have conducted some comparison experiments to test which assignment is better.
We set 5 different assignment ways, and present results in Table~\ref{tab: assignment}. 
The results are ranked by MEAN Dice score, and the last assignment has the best MEAN Dice score. 
The first assignment does not assign the ET region to the T1CE modality and the WT region to the FLAIR modality, and the segmentation result is the worst, confirming the correlation between the modality and the target region and that it is reasonable for CIML to introduce human priori knowledge.
Other assignments have similar segmentation results that exceed or are comparable to the results of nnUNet.

\begin{figure}[htb]
    \centering
    \includegraphics[width=0.7\textwidth]{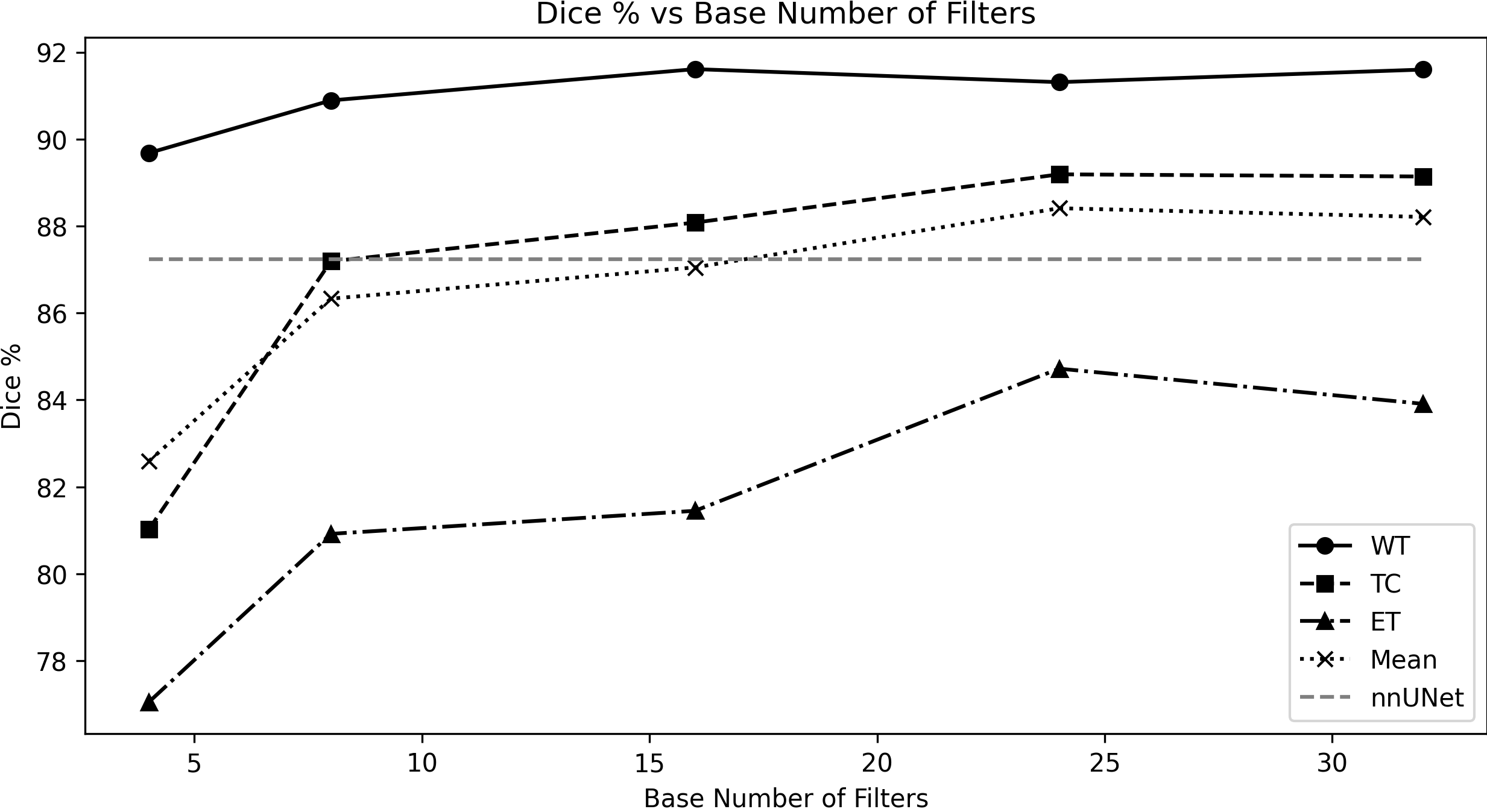}
    \caption{\blue{Visualization of dice score on the \texttt{BraTS2020} dataset for the different base numbers of filters, more filters mean wider network.
    The grey dash horizontal line indicates the average dice score of nnUNet.}}
    \label{fig:nf_dice}
\end{figure}
\begin{figure}[htb]
    \centering
    \includegraphics[width=0.7\textwidth]{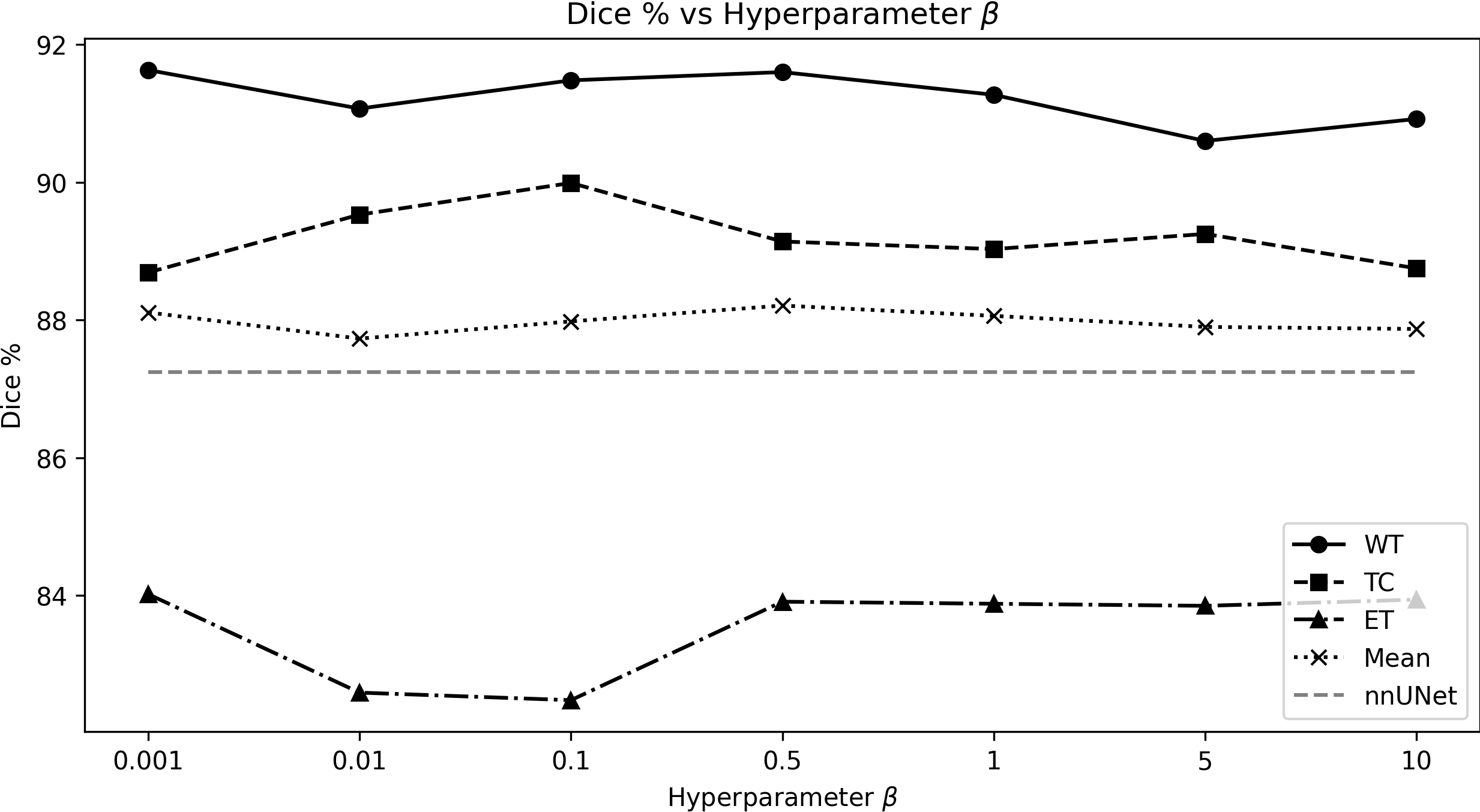}
    \caption{\blue{Visualization of dice score for the different hyperparameter $\beta$ on the \texttt{BraTS2020} dataset.
    The grey dash horizontal line indicates the average dice score of nnUNet.}}
    \label{fig:beta_dice}
\end{figure}
\subsubsection{Hyperparametric Studies}
In our work, there are two essential hyperparameters, the base number of filters and the hyperparameter $\beta$, which controls the tradeoff between the CE loss and KL loss.

As shown in Fig. \ref{fig:nf_dice}, we explore the results of dice scores with filters of $4$, $8$, $16$, $24$ and $32$. As base number filters increase, the mean dice (red dash line) score increases, and $24$ and $32$ base filters approach the best dice score.

Additionally, we experiment with $\beta$ from $1 \times 10^{-3}$ to $10$ and find that the highest average dice were achieved with $\beta$ equals $0.5$. All beta settings exceeded the results of nnUNet, indicating that our algorithm CIML is insensitive to $\beta$ and has good generalization.

\begin{figure}[!htbp] 
  \centering
  \includegraphics[width=\linewidth]{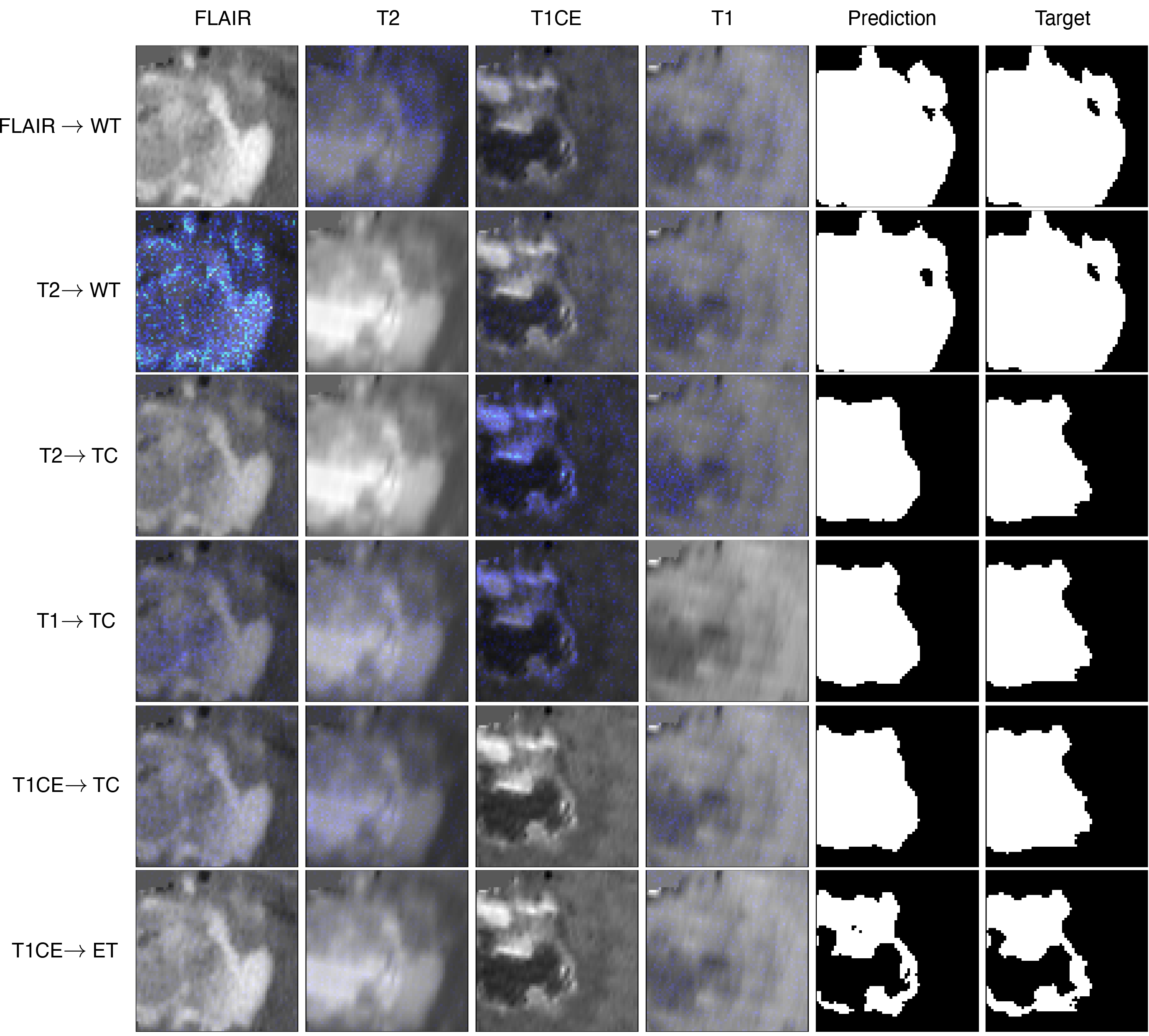}
  \caption{\blue{Information representation visualization of a case from the \texttt{BraTS2020} challenge.  Each row corresponds to a primary modality and a corresponding target region pair.
  The first four columns display the FLAIR, T2, T1CE, and T1 images and the information visualization representation masks (Dark blue indicates the smallest value, light blue means the middle value, and yellow illustrates the largest value).
  Specifically, the primary modal displays only the original image without masks.
  The fifth column, `Prediction,' displays prediction outcomes, while the final column, `Target,' displays the ground truth. 
  Note: the images are best viewed in color for optimal clarity.}}
  \label{fig:visinf}
\end{figure}

\begin{figure}[!htb]
    \centering
    \includegraphics[width=\textwidth]{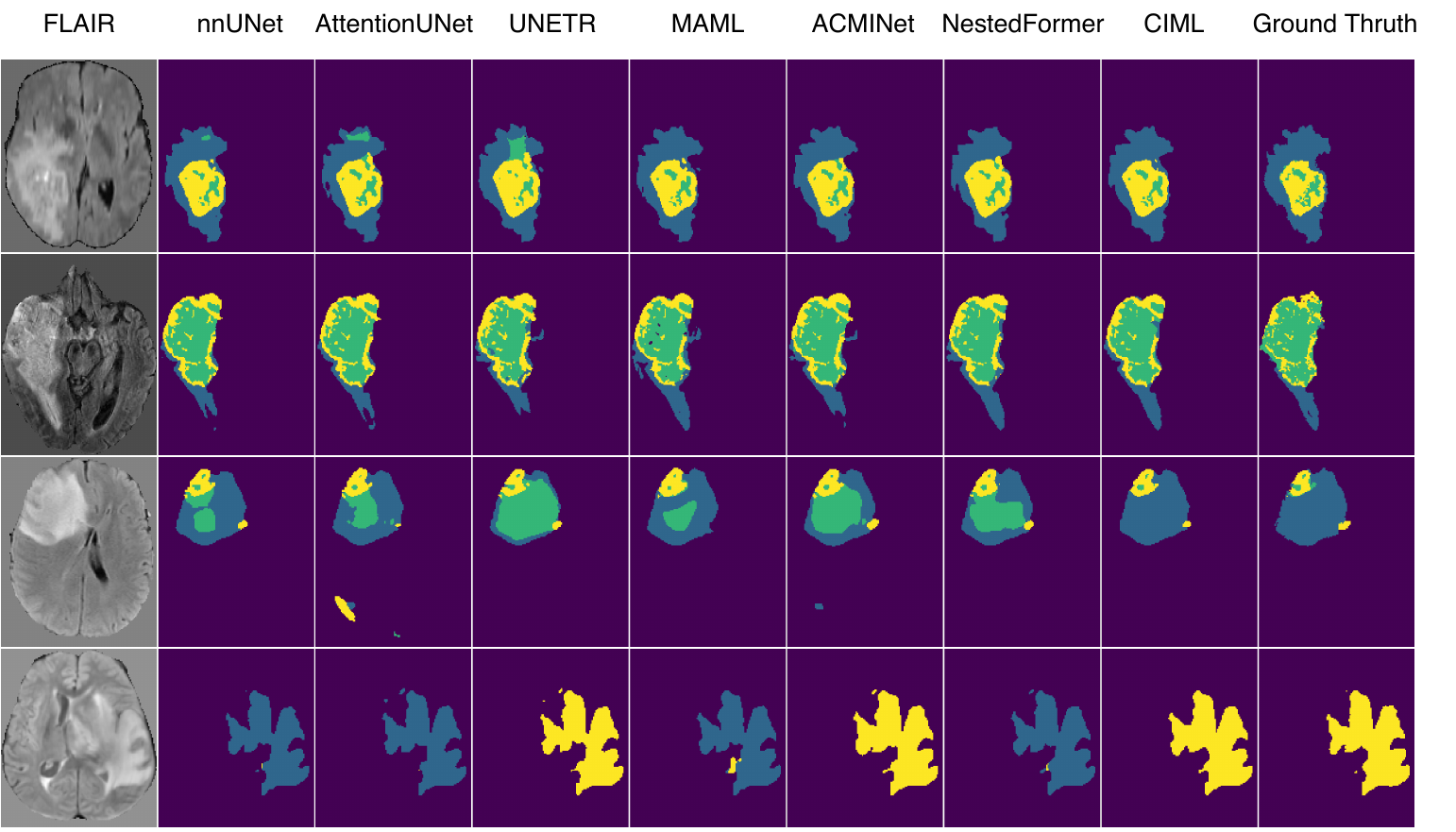}
    \caption{Visualization of the predicted segmentation maps compared with the SOTA methods.
    The first column shows FLAIR images, and the other column shows the prediction results and the corresponding ground truth. \blue{Note: the images are best viewed in color for optimal clarity.}}
    \label{fig:visresult}
\end{figure}

\subsection{Visualization with Interpretability}

\subsubsection{Segmentation Results}

As illustrated in Figure~\ref{fig:visresult}, we present the segmentation results of CIML alongside other state-of-the-art methods. Both nnUNet and our proposed method exhibit high accuracy across various cases. Notably, only UNETR and our method correctly classify the corresponding region as TC.

\subsubsection{Complementary Information}

In subsection~\ref{subsec:demo}, the demonstration experiment verifies that our proposed redundancy filtering is efficient.
In CIML, the extracted information representations are high-dimensional features, and to visualize the complementary information, we use the Grad-CAM algorithm to get class-discriminative localization maps (we will refer to it as a heatmap when there is no possibility of confusion) that highlight the voxels focused on by our proposed CIML algorithm.

In deep convolutional networks, the deeper layers typically extract semantic information but lose positional correspondence. 
Conversely, the shallower layers are able to extract detailed pattern information while preserving positional correspondence.
Therefore, in order to maintain a clear understanding of positional information, we choose to visualize the information representations from the shallowest layers of the network, which have the same resolutions as the images with $C$ channels.
In the \texttt{BraTS2020} dataset, we default decompose segmentation into four segmentors.
Additionally, there are six modality target region pairs.
For $i$-th pair, let $\{A_i\}_{c=1}^C$ be the information representations (with $C$ channels), and $y^k$ be the logit for a chosen pixel class $k$.
Grad-CAM averages the partial gradients of $y^k$ with respect to $N$ voxels of each information representation. 
The heatmap for class $k$ in pair $i$
\begin{equation}
    \zeta^k_i=\mathrm{ReLU}\left( \sum_{c} \alpha^{c}_k A_i^c \right),
\end{equation}
with
\begin{equation}
    \alpha^{c}_k = \frac{1}{N} \sum_{n} \frac{\partial y^k}{\partial A^c_{i,n}},
\end{equation}
where $\alpha^{c}_k$ is the neuron importance weights of the $c-th$ channel of information representations for pair $i$. 

In Figure~\ref{fig:visinf}, we visualize information representations extracted from the last messages by applying heatmaps as masks added to the original images.
The last message refers to the shallow embedding in the neural network, which is the message with the highest resolution.
\blue{We normalize the heatmaps across all rows so that the amount of information in each auxiliary modality can be easily compared.}
Yellow represents the largest value, dark blue represents the smallest value, and light blue represents the middle value.
As shown in the first row, FLAIR is the primary modality, and the T2 image contains the most complementary information compared to the other two modalities. 
In the second row, T2 is the primary modality, and the FLAIR image contains the most complementary information.
Additionally, the left-down region of the FLAIR image contains more information, which is consistent with medical domain knowledge.
This region, depicted in hyperintensity (lighter in source images), indicates the presence of edema, typically locates at the periphery of the WT.
The third and fourth rows illustrate TC is the target region and the hyperintensity regions in T1CE, which means ET regions, contain the most information.
\blue{In the final two rows concerning T1CE as the primary modality, it is observed that auxiliary modalities contribute less information overall. However, the T2 image still provides some valuable complementary insights, particularly in the hyperintense areas, which appear as low-intensity zones in the T1CE image indicating the necrotic and non-enhancing tumor core.}
These results demonstrate that the results predicted by our proposed CIML methods are consistent with the physician's domain knowledge and permit further verification that the algorithm can extract complementary information from high-dimensional data.

\subsubsection{Complementary Information Weights}

As shown in Figure~\ref{fig:visinf}, auxiliary modalities contain various complementary information for each \textit{modality-region} pair.
To further explore the contribution of different auxiliary modalities to segmentation, we propose using the complementary information weight to quantify the contribution.
We define the complementary information weight for pair $i$ as

\begin{equation}
    \varpi^k_i = \frac{\sum_n{\zeta^k_{i,n}}}{\sum_j{\sum_n{\zeta^k_{j,n}}}}.
\end{equation}

\begin{figure}[!htb]
    \centering
    \includegraphics[width=0.7\textwidth]{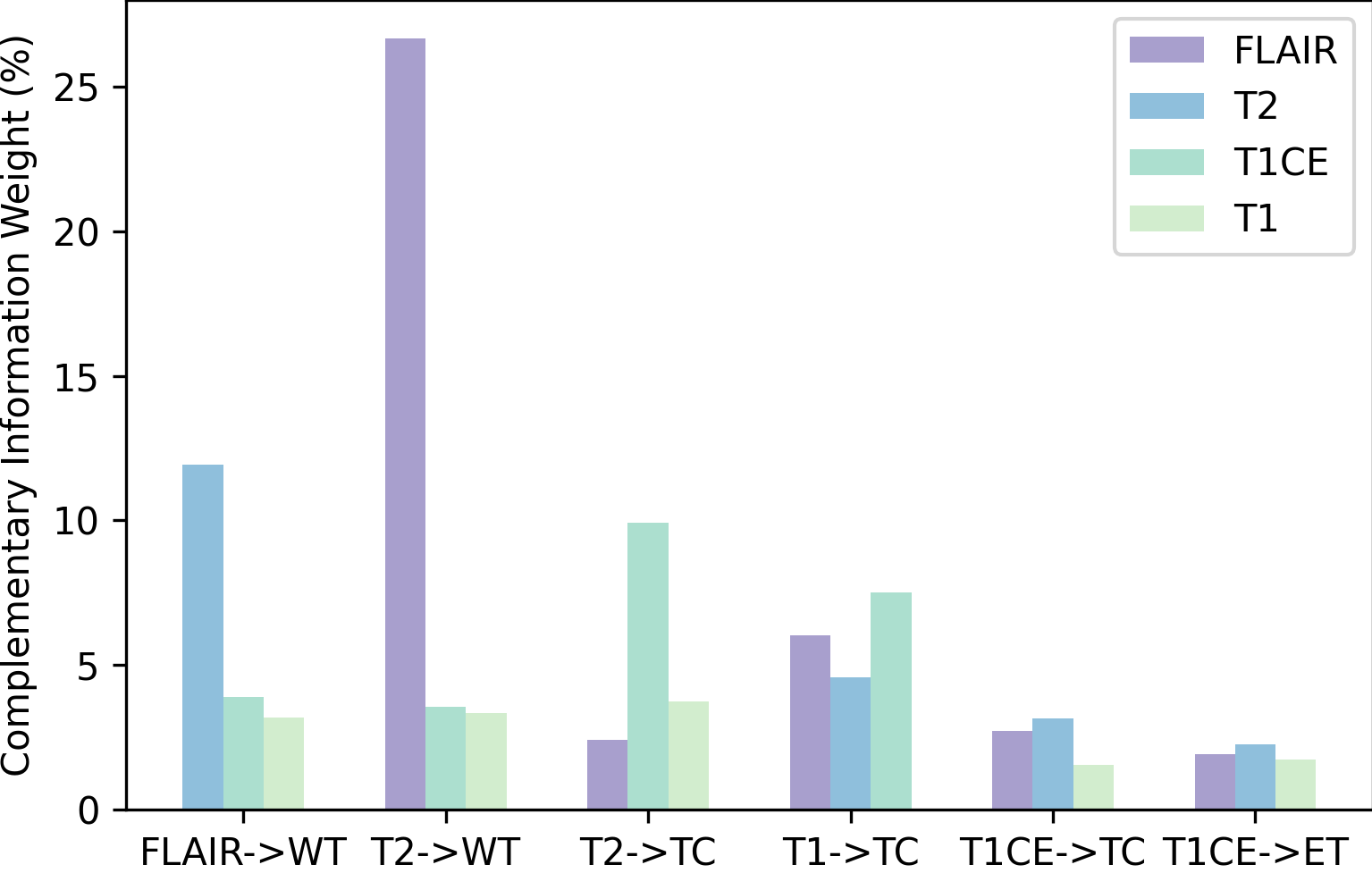}
    \caption{Visualization of the complementary information weights by our CIML on the \texttt{BraTS2020} dataset.}
    \label{fig:info_weight}
\end{figure}

Figure~\ref{fig:info_weight} reveals that FLAIR provides the most substantial complementary information weight for the T2 segmenting of the WT target region. 
% Additionally, when segmenting the TC and ET target regions, T1CE requires minimal complementary information. 
% Moreover, T1CE provides more complementary information for T1 and T2 segmentation of the TC target region than the other modalities.
\blue{Additionally, when segmenting the TC and ET target regions, T1CE requires minimal supplementary information. Furthermore, T1CE offers more complementary information than other modalities, especially for subtasks where the TC target region is segmented using T1 and T2.}
These findings are consistent with the observations in Figure~\ref{fig:brats2020}, where T1CE is sensitive to NCR/NET and ET regions, while FLAIR is sensitive to the WT target region. 
Thus, CIML can effectively prioritize sensitive modalities and extract discriminative features from auxiliary modalities.

\section{Acknowledgements }
\label{acknowledgements }

This work was supported by the STCSM (22511106000, 22511106004).
\section{Conclusion}
\label{conclusion}

In this study, we propose the \textbf{complementary information mutual learning (CIML)} framework, which provides a unique solution to the problem of \textit{inter-modal} redundant information in multimodal learning, an issue not addressed by previous state-of-the-art (SOTA) methods. 
Our framework, based on \textit{addition} operation, provides a systematic approach by employing inductive bias for task decomposition and message passing for redundancy filtering, thereby enhancing the effectiveness of multimodal medical image segmentation.
We extensively evaluate our approach and demonstrate its effectiveness, outperforming current SOTA methods.
Furthermore, the message passed through redundancy filtering enables the application of visualization techniques such as Grad-CAM, thus improving the interpretability of the algorithm.
In conclusion, our proposed CIML framework has the potential to significantly enhance the quality and reliability of multimodal medical image segmentation, ultimately leading to improved clinical diagnosis and treatment outcomes.

\bibliography{main}
\clearpage
\newpage
\appendix
\blue{
\section{Mutual Information and Venn Diagrams}
\label{sec:Mutual_Information_and_Venn_Diagrams}
\subsection{Entropy}

Entropy is a fundamental concept in information theory, quantifying the uncertainty or randomness in a random variable. 
Formally, the entropy $H(X)$ of a continuous random variable $X$ with probability density function $p(x)$ is defined as:
\begin{equation}
H(X) = -\int dx \, dy p(x) \log p(x) ,
\end{equation}
where  $p(x)$ is the probability density of $x$. Entropy measures the expected amount of information required to describe the outcome of $X$.

\subsection{Mutual Information}

Mutual information (MI) quantifies the amount of information one random variable contains about another. For two continuous random variables $X$ and $Y$ with a joint probability density function $p(x, y)$, the mutual information $\mathcal{I}(X; Y)$ is defined as:
\begin{equation}
\mathcal{I}(X; Y) = \int dx \, dy p(x, y) \log \frac{p(x, y)}{p(x)p(y)} ,
\end{equation}
where $p(x)$ and $p(y)$ are the marginal probability density functions of $X$ and $Y$, respectively. 
MI quantifies the reduction in uncertainty about $X$ given knowledge of $Y$.

\subsection{Conditional Mutual Information}

Conditional mutual information (CMI) extends MI to account for a third random variable $Z$. The conditional mutual information $\mathcal{I}(X; Y | Z)$ is defined as:
\begin{equation}
\mathcal{I}(X; Y | Z) = \int dx \, dy \, dz p(z) p(x, y | z) \log \frac{p(x, y | z)}{p(x | z)p(y | z)} ,
\end{equation}
where $p(x, y | z)$ is the conditional joint probability density function of $X$ and $Y$ given $Z$, and $p(x | z)$ and $p(y | z)$ are the conditional marginal densities. CMI measures the information shared between $X$ and $Y$ accounting for the effect of $Z$.

\subsection{Relationship with Venn Diagrams}

Venn diagrams provide a visual representation to illustrate relationships between entropy, MI, and CMI. In a Venn diagram, the entropy $H(X)$ of a random variable $X$ is depicted as the area of a circle, representing the total uncertainty of $X$. Similarly, the entropy $H(Y)$ of another random variable $Y$ is represented by another circle.

The overlap between the two circles represents the mutual information $\mathcal{I}(X; Y)$, which quantifies the shared information between $X$ and $Y$ and the reduction in uncertainty of one variable due to knowledge of the other.

Introducing a third random variable $Z$, the conditional mutual information $\mathcal{I}(X; Y | Z)$ can be visualized as the overlap between the areas of $X$ and $Y$, excluding the part influenced by $Z$. In a Venn diagram with three circles representing $X$, $Y$, and $Z$, CMI focuses on the shared area between $X$ and $Y$ while excluding the part of the circle $Z$.

Venn diagrams provide an intuitive method to understand interactions between different information-theoretic measures, highlighting dependencies and shared information among multiple random variables.

\subsection{Common Formulas and Properties for Mutual Information}

\textbf{Symmetry:} Mutual information is symmetric, which means that the mutual information between two random variables \(X\) and \(Y\) is the same regardless of the order in which the variables are considered. Mathematically, this can be expressed as:

\begin{equation}
\mathcal{I}(X; Y) = \mathcal{I}(Y; X).
\end{equation}

This property follows from the definition of mutual information, which is based on joint and marginal probabilities that are symmetric with respect to \(X\) and \(Y\).

\textbf{Non-negativity:} Mutual information is always non-negative. This means that the mutual information between any two random variables \(X\) and \(Y\) is greater than or equal to zero:

\begin{equation}
\mathcal{I}(X; Y) \geq 0.
\end{equation}

This property arises because mutual information is a measure of the reduction in uncertainty about one random variable given knowledge of another, and this reduction in uncertainty cannot be negative.

\textbf{Chain Rule:} The chain rule for mutual information allows us to break down the mutual information between multiple variables into simpler components. For three random variables \(A\), \(B\), and \(C\), the chain rule states:

\begin{equation}
\mathcal{I}(A, B; C) = \mathcal{I}(A; C) + \mathcal{I}(B; C \mid A).
\end{equation}

This rule can be extended to more variables. For example, for four random variables \(A\), \(B\), \(C\), and \(D\), the chain rule becomes:

\begin{equation}
\mathcal{I}(A, B, C; D) = \mathcal{I}(A; D) + \mathcal{I}(B; D \mid A) + \mathcal{I}(C; D \mid A, B).
\end{equation}

In general, for random variables \(X_1, X_2, \ldots, X_n\) and \(Y\), the chain rule is given by:

\begin{equation}
\mathcal{I}(X_1, X_2, \ldots, X_n; Y) = \sum_{i=1}^n \mathcal{I}(X_i; Y \mid X_1, X_2, \ldots, X_{i-1}).
\end{equation}

This powerful rule helps in decomposing mutual information into more manageable parts, which can be particularly useful in proofs and derivations.

\section{Bayesian Graph Representation}
\subsection{Introduction to Bayesian Graph Representation}
\label{sec:bayesian_graph_rep}
Bayesian networks, also known as Bayesian graph representations, are powerful tools for representing and reasoning about probabilistic dependencies and causal relationships among a set of variables. A Bayesian network is a directed acyclic graph (DAG) where each node represents a random variable, and the edges represent probabilistic dependencies between these variables.
Bayesian networks are particularly useful for causal inference because the directed edges can be interpreted as causal influences. If there is a directed edge from \(X_i\) to \(X_j\), it suggests that \(X_i\) is a direct cause of \(X_j\). This causal interpretation allows for the modeling of complex causal relationships and the prediction of the effects of interventions.

\subsection{Derivation for Equation (4)}
\label{sec:der_eq4}

We aim to derive the joint probability distribution:
\begin{equation}
\label{eq:prob}
p(X_1, X_2, Y_1, K_2) = p(K_2 \mid X_1, X_2) \cdot p(X_1, X_2, Y_1).
\end{equation}

Consider the Bayesian network depicted in Figure~\ref{fig:bayes-graph}, which illustrates the relationships among the variables $X$, $X_1$, $X_2$, $Y_1$, and $K_2$.

\begin{figure}[h!]
  \centering
  \begin{tikzpicture}
    % Nodes
    \node (X) at (6,2) [circle, draw] {$X$};
    \node (X1) at (4,0) [circle, draw] {$X_1$};
    \node (X2) at (4,4) [circle, draw] {$X_2$};
    \node (K2) at (2,2) [circle, draw] {$K_2$};
    \node (Y1) at (0,2) [circle, draw] {$Y_1$};
    
    % Arrows
    \draw[->] (X) -- (X1);
    \draw[->] (X) -- (X2);
    \draw[->] (X1) -- (K2);
    \draw[->] (X1) -- (Y1);
    \draw[->] (X2) -- (K2);
    \draw[->] (X2) -- (Y1);
  \end{tikzpicture}
  \caption{A Bayesian network representing the relationship of modalities $X_1$, $X_2$, target $Y_1$, and representation $K_2$.}
  \label{fig:bayes-graph}
\end{figure}
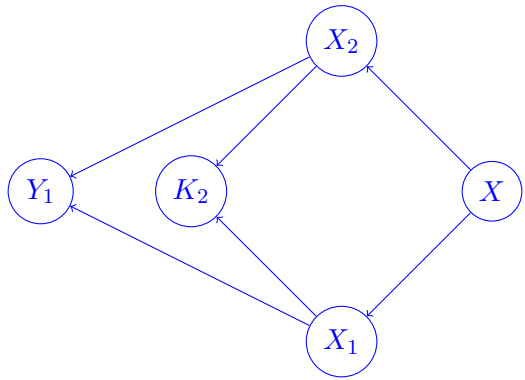

From the graph, it is clear that $Y_1$ and $K_2$ have $X_1$ and $X_2$ as their parent nodes, and $X_1$ and $X_2$ have $X$ as their parent node. The joint distribution of the variables can be expressed as:
\begin{equation}
\label{eq:joint_dist}
p(X_1, X_2, Y_1, K_2, X) = p(Y_1 \mid X_1, X_2) \cdot p(K_2 \mid X_1, X_2) \cdot p(X_1, X_2 \mid X) \cdot p(X).
\end{equation}

To derive the desired joint distribution, we integrate over the latent variable $X$:
\begin{align}
p(X_1, X_2, Y_1, K_2) &= \int p(X_1, X_2, Y_1, K_2, X) \, dX \nonumber \\
&= \int p(Y_1 \mid X_1, X_2) \cdot p(K_2 \mid X_1, X_2) \cdot p(X_1, X_2 \mid X) \cdot p(X) \, dX \nonumber \\
&= \int p(Y_1 \mid X_1, X_2) \cdot p(K_2 \mid X_1, X_2) \cdot p(X_1, X_2, X) \, dX \nonumber \\
&= p(Y_1 \mid X_1, X_2) \cdot p(K_2 \mid X_1, X_2) \cdot p(X_1, X_2) \nonumber \\
&= p(K_2 \mid X_1, X_2) \cdot p(X_1, X_2, Y_1).
\end{align}

Thus, we have derived that
\begin{equation}  
p(X_1, X_2, Y_1, K_2) = p(K_2 \mid X_1, X_2) \cdot p(X_1, X_2, Y_1),
\end{equation}

as required.

\section{Mutual Information Derivation}

\subsection{Derivation for Equation (2)}
\label{sec:der_eq2}

To prove the given equation,
\begin{equation}
\begin{split}
\mathcal{I}_1(X_1, X_2; K_2) &= \underbrace{\mathcal{I}_2(K_2; Y_1 \mid X_1)}_{\text{complementary predictive information}} \\
&+ \underbrace{\mathcal{I}_3(K_2; X_1)}_{\text{duplicated information}} \\
&+ \underbrace{\mathcal{I}_4(K_2; X_2 \mid X_1, Y_1)}_{\text{unique but irrelevant information}},
\end{split}
\end{equation}
we start by expressing each term and using the chain rule for mutual information.

First, using the chain rule for mutual information:

\begin{equation}
\mathcal{I}_1(X_1, X_2; K_2) = \mathcal{I}_3(K_2; X_1) + \mathcal{I}(X_2; K_2 \mid X_1).
\end{equation}

Next, we expand \(\mathcal{I}(X_2; K_2 \mid X_1)\) by the chain rule for mutual information:

\begin{equation}
\mathcal{I}(X_2; K_2 \mid X_1) = \mathcal{I}_2(K_2; Y_1 \mid X_1) + \mathcal{I}_4(K_2; X_2 \mid X_1, Y_1).
\end{equation}

Combining the above results, we get:

\begin{equation}
\mathcal{I}_1(X_1, X_2; K_2) = \underbrace{\mathcal{I}_2(K_2; Y_1 \mid X_1)}_{\text{complementary predictive information}} + \underbrace{\mathcal{I}_3(K_2; X_1)}_{\text{duplicated information}} + \underbrace{\mathcal{I}_4(K_2; X_2 \mid X_1, Y_1)}_{\text{unique but irrelevant information}}.
\end{equation}

Thus, the equation is proved.

\subsection{Derivation of Equation (5)}
\label{sec:der_eq5}

To prove the given equation:
\begin{equation}
    \begin{split}
        \mathcal{I}_1(X_1, X_2; K_2) &= \int p(x_1, x_2, \kappa_2) \log \left( \frac{p(\kappa_2 \mid x_1, x_2)}{p(\kappa_2)} \right) \, dx_1 \, dx_2 \, d\kappa_2 \\
        &\leq \int p(x_1, x_2, \kappa_2) \log \left( \frac{p(\kappa_2 \mid x_1, x_2)}{r(\kappa_2)} \right) \, dx_1 \, dx_2 \, d\kappa_2 \\
        &\approx \frac{1}{N} \sum_{i=1}^N \int p(\kappa_2 \mid x_1^i, x_2^i) \log \left( \frac{p(\kappa_2 \mid x_1^i, x_2^i)}{r(\kappa_2)} \right) \, d\kappa_2,
    \end{split}
\end{equation}

First, based on the definition of mutual information, we have:
\begin{equation}
\mathcal{I}_1(X_1, X_2; K_2) = \int p(x_1, x_2, \kappa_2) \log \left( \frac{p(\kappa_2 \mid x_1, x_2)}{p(\kappa_2)} \right) \, dx_1 \, dx_2 \, d\kappa_2,
\end{equation}

Next, by substituting \(p(x_1, x_2, \kappa_2)\), we obtain:
\begin{equation}
\mathcal{I}_1(X_1, X_2; K_2) = \int p(\kappa_2 \mid x_1, x_2) p(x_1, x_2) \log \left( \frac{p(\kappa_2 \mid x_1, x_2)}{p(\kappa_2)} \right) \, dx_1 \, dx_2 \, d\kappa_2.
\end{equation}

To derive the inequality, we apply the non-negativity of the Kullback-Leibler (KL) divergence:
\begin{equation}
D_{\text{KL}}(P \| Q) = \int p(x) \log \left( \frac{p(x)}{q(x)} \right) \, dx \geq 0,
\end{equation}
which implies:
\begin{equation}
\int p(\kappa_2 \mid x_1, x_2) \log \left( \frac{p(\kappa_2 \mid x_1, x_2)}{p(\kappa_2)} \right) \, d\kappa_2 \leq \int p(\kappa_2 \mid x_1, x_2) \log \left( \frac{p(\kappa_2 \mid x_1, x_2)}{r(\kappa_2)} \right) \, d\kappa_2.
\end{equation}

Thus, we have:
\begin{equation}
\begin{split}
\int p(\kappa_2 \mid x_1, x_2) p(x_1, x_2) \log \left( \frac{p(\kappa_2 \mid x_1, x_2)}{p(\kappa_2)} \right) \, dx_1 \, dx_2 \, d\kappa_2 \\
\leq \int p(\kappa_2 \mid x_1, x_2) p(x_1, x_2) \log \left( \frac{p(\kappa_2 \mid x_1, x_2)}{r(\kappa_2)} \right) \, dx_1 \, dx_2 \, d\kappa_2.
\end{split}
\end{equation}

Finally, using Monte Carlo sampling, we approximate:
\begin{equation}
\begin{split}
\int p(x_1, x_2, \kappa_2) \log \left( \frac{p(\kappa_2 \mid x_1, x_2)}{r(\kappa_2)} \right) \, dx_1 \, dx_2 \, d\kappa_2 \\
\approx \frac{1}{N} \sum_{i=1}^N \int p(\kappa_2 \mid x_1^i, x_2^i) \log \left( \frac{p(\kappa_2 \mid x_1^i, x_2^i)}{r(\kappa_2)} \right) \, d\kappa_2.
\end{split}
\end{equation}

\subsection{Derivation of Equation (6)}
\label{sec:der_eq6}
To prove the given equation:
\begin{equation}
    \begin{split}
       & \mathcal{I}_2(K_2; Y_1 \mid X_1) \\
       &= \int p(x_1, x_2, \kappa_2, y_1) \log \left( \frac{p(y_1 \mid \kappa_2, x_1)}{p(y_1 \mid x_1)} \right) \, dx_1 \, dx_2 \, d\kappa_2 \, dy_1 \\
        &\geq \int p(x_1, x_2, \kappa_2, y_1) \log \left( \frac{q(y_1 \mid \kappa_2, x_1)}{p(y_1 \mid x_1)} \right) \, dx_1 \, dx_2 \, d\kappa_2 \, dy_1 \\
        &\approx \frac{1}{N}  \sum_{i=1}^N \int p(\kappa_2 \mid x_1^i, x_2^i) \log q(y_1^i \mid \kappa_2, x_1^i) \, d\kappa_2 + H.
    \end{split}
\end{equation} 

First, based on the definition of mutual information, we get:
\begin{equation}
\mathcal{I}_2(K_2; Y_1 \mid X_1) = \int p(\kappa_2, x_1, y_1) \log \left( \frac{p(y_1 \mid \kappa_2, x_1)}{p(y_1 \mid x_1)} \right) \, dx_1 \, d\kappa_2 \, dy_1.
\end{equation}

Then, based on the integral, the joint distribution is equal to the marginal distribution:
\begin{equation}
p(\kappa_2, x_1, y_1) = \int p(\kappa_2, x_1, x_2, y_1) \, dx_2,
\end{equation}
we obtain:
\begin{equation}
\mathcal{I}_2(K_2; Y_1 \mid X_1) = \int p(\kappa_2, x_1, x_2, y_1) \log \left( \frac{p(y_1 \mid \kappa_2, x_1)}{p(y_1 \mid x_1)} \right) \, dx_1 \, dx_2 \, d\kappa_2 \, dy_1.
\end{equation}

To derive the inequality, we apply the non-negativity of the Kullback-Leibler (KL) divergence:
\begin{equation}
D_{\text{KL}}(P \| Q) = \int p(x) \log \left( \frac{p(x)}{q(x)} \right) \, dx \geq 0,
\end{equation}
which implies:
\begin{equation}
\int p(y_1 \mid \kappa_2, x_1) \log \left( \frac{p(y_1 \mid \kappa_2, x_1)}{p(y_1 \mid x_1)} \right) \, dy_1 \geq \int p(y_1 \mid \kappa_2, x_1) \log \left( \frac{q(y_1 \mid \kappa_2, x_1)}{p(y_1 \mid x_1)} \right) \, dy_1.
\end{equation}

Then:
\begin{equation}
\begin{split}
\int p(\kappa_2, x_1) \, d\kappa_2 \, dx_1 \int p(y_1 \mid \kappa_2, x_1) \log \left( \frac{p(y_1 \mid \kappa_2, x_1)}{p(y_1 \mid x_1)} \right) \, dy_1 \\
\geq \int p(\kappa_2, x_1) \, d\kappa_2 \, dx_1 \int p(y_1 \mid \kappa_2, x_1) \log \left( \frac{q(y_1 \mid \kappa_2, x_1)}{p(y_1 \mid x_1)} \right) \, dy_1.
\end{split}
\end{equation}

Thus:
\begin{equation}
\begin{split}
\int p(x_1, x_2, \kappa_2, y_1) \log \left( \frac{p(y_1 \mid \kappa_2, x_1)}{p(y_1 \mid x_1)} \right) \, dx_1 \, dx_2 \, d\kappa_2 \, dy_1 \\
\geq \int p(x_1, x_2, \kappa_2, y_1) \log \left( \frac{q(y_1 \mid \kappa_2, x_1)}{p(y_1 \mid x_1)} \right) \, dx_1 \, dx_2 \, d\kappa_2 \, dy_1.
\end{split}
\end{equation}

Further, we can split the upper bound into two terms:
\begin{equation}
\begin{split}
& \int p(x_1, x_2, \kappa_2, y_1) \log q(y_1 \mid \kappa_2, x_1) \, dx_1 \, dx_2 \, d\kappa_2 \, dy_1 \\ &
= \int p(x_1, x_2, \kappa_2, y_1) \log q(y_1 \mid \kappa_2, x_1) \, dx_1 \, dx_2 \, d\kappa_2 \, dy_1 - H,
\end{split}
\end{equation}
where 
\begin{equation}
\begin{split}
H &= \int p(x_1, x_2, \kappa_2, y_1) \log p(y_1 \mid x_1) \, dx_1 \, dx_2 \, d\kappa_2 \, dy_1 \\
&= \int p(x_1, y_1) \log p(y_1 \mid x_1) \, dx_1 \, dy_1,
\end{split}
\end{equation}
because \(H\) is independent of our optimization procedure and so can be ignored.
For the approximation using Monte Carlo sampling:
\begin{equation}
\begin{split}
& \int p(x_1, x_2, \kappa_2, y_1) \log \left( \frac{q(y_1 \mid \kappa_2, x_1)}{p(y_1 \mid x_1)} \right) \, dx_1 \, dx_2 \, d\kappa_2 \, dy_1 \\ &
\approx \frac{1}{N} \sum_{i=1}^N \int p(\kappa_2 \mid x_1^i, x_2^i) \log q(y_1^i \mid \kappa_2, x_1^i) \, d\kappa_2 + H.
\end{split}
\end{equation}
% \clearpage
% \newpage
% \bibliography{refa}

}
\end{document}